\documentclass{tlp}
\usepackage{aopmath}

\usepackage{latexsym}
\usepackage{amstext,amssymb}
\usepackage{fancybox}
\usepackage{url}

\newcommand{\namemap}{\ensuremath{n}}
\newcommand{\nameset}{\ensuremath{N}}

\newtheorem{definition}{Definition}

\newcommand{\oksym}{{\mathsf{ok}}}
\newcommand{\rdysym}{{\mathsf{rdy}}}
\newcommand{\kosym}{{\mathsf{ko}}}
\newcommand{\blockedsym}{{\mathsf{bl}}}
\newcommand{\appliedsym}{{\mathsf{ap}}}

\newcommand{\ok}[1]{\ensuremath{\oksym(#1)}}
\newcommand{\rdy}[2]{\ensuremath{\rdysym(#1,#2)}} 
\newcommand{\ko}[1]{\ensuremath{\kosym(#1)}}
\newcommand{\blocked}[1]{\ensuremath{\blockedsym(#1)}}
\newcommand{\blockedP}[1]{\ensuremath{\blockedsym^{+}(#1)}}
\newcommand{\blockedJ}[1]{\ensuremath{\blockedsym^{-}(#1)}}
\newcommand{\applied}[1]{\ensuremath{\appliedsym(#1)}}

\newcommand{\BEpreferred}{\textsc{be}-preferred} 

\newcommand{\LPif}{\leftarrow}
\newcommand{\RULE}[3]{#1:&#2&\LPif&#3}

\newcommand{\To}[1]{\ensuremath{T_{#1}}}
\newcommand{\T}[2]{\To{#1}#2}
\newcommand{\Tindo}[2]{\To{#2}^{#1}}
\newcommand{\Tind}[3]{\Tindo{#1}{#2}#3}
\newcommand{\head}[1]{\mathit{head}(#1)}
\newcommand{\pbody}[1]{\mathit{body}^{+}(#1)}
\newcommand{\nbody}[1]{\mathit{body}^{-}(#1)}
\newcommand{\body}[1]{\mathit{body}(#1)}
\newcommand{\nafo}[0]{\mathit{not}}
\newcommand{\naf}[0]{\nafo\;}

\newcommand{\reduct}[2]{\ensuremath{#1^{#2}}}
\newcommand{\reductr}[1]{\ensuremath{#1^{+}}}
\newcommand{\reductBE}[1]{\ensuremath{#1^{-}}}

\newcommand{\Tho}[0]{\ensuremath{Cn}}
\newcommand{\Th}[1]{\ensuremath{\Tho(#1)}}
\newcommand{\iec}[0]{i.e.,\ }

\newcommand{\PREC}[2]{\ensuremath{{#1}\prec{#2}}}
\newcommand{\PRECMo}[0]{\ensuremath{<}}
\newcommand{\PRECM}[2]{\ensuremath{{#1}\PRECMo{#2}}}
\newcommand{\notPRECM}[2]{\ensuremath{{#1}\not\PRECMo{#2}}}
\newcommand{\PRECMoi}[1]{\ensuremath{\PRECMo_{#1}}}
\newcommand{\PRECMi}[3]{\ensuremath{{#2}\PRECMoi{#1}{#3}}}

\newcommand{\TA}[1]{\ensuremath{{#1}^\star}}
\newcommand{\TAi}[2]{\ensuremath{#1_{\ref{#2}}^\star}}

\newcommand{\GR}[2]{\ensuremath{{\Gamma}_{#1}^{#2}}} 

\newcommand{\ap}[2]   {\ensuremath{{a_{#1}(#2   )}}}   
\newcommand{\bl}[3]   {\ensuremath{{b_{#1}(#2,#3)}}}   
\newcommand{\cokt}[2] {\ensuremath{{c_{#1}}(#2)}}       
\newcommand{\cok}[3]  {\ensuremath{{c_{#1}(#2,#3)}}}   
\newcommand{\de}[1]   {\ensuremath{{d(#1)}}}      

\newcommand{\Lit}{\ensuremath{\mathcal{L}}} 

\newcommand{\namef}[1]{\namemap(#1)}
\newcommand{\nameo}[0]{\mathit{n}}
\newcommand{\name}[1]{\nameo_{#1}}

\newcommand{\eg}[0]{e.g.\ }
\newcommand{\egc}[0]{e.g.,\ }

\newcommand{\WZL}{\textsc{wzl}}
\newcommand{\DST}{\textsc{dst}}
\newcommand{\BE}{\textsc{be}}

\newcommand{\nbcite}[1]{\cite{#1}} 

\title{A Framework for Compiling Preferences in Logic Programs}

\author[J.\ P.\ Delgrande, T.\ Schaub, and H.\ Tompits]{%
  JAMES P.\ DELGRANDE\\
  School of Computing Science,
  Simon Fraser University,\\
  Burnaby, B.C.,
  Canada  V5A 1S6\\
  \email{jim@cs.sfu.ca}
  \and
  TORSTEN SCHAUB%
  \thanks{Affiliated with the
    School of Computing Science at
    Simon Fraser University,
    Burnaby, Canada.}
  \\
  Institut f\"ur Informatik,
  Universit\"at Potsdam,      \\  
  Postfach 90 03 27,
  D--14439 Potsdam,
  Germany                    \\
  \email{torsten@cs.uni-potsdam.de}
  \and
  HANS TOMPITS\\
  Institut f\"ur Informationssysteme 184/3,
  Technische Universit\"at Wien,\\
  Favoritenstra{\ss}e~9--11,
  A--1040 Vienna, Austria\\
  \email{tompits@kr.tuwien.ac.at}
  }

\begin{document}

\maketitle

\begin{abstract}
We introduce a methodology and framework for expressing general preference
information in logic programming under the answer set semantics.
An \emph{ordered} logic program is an extended logic program in which
rules are named by unique terms, and in which preferences among rules
are given by a set of atoms of form $s \prec t$ where $s$ and $t$
are names.
An ordered logic program is transformed into a second, regular, extended
logic program wherein the preferences are respected, in that the answer
sets obtained in the transformed program correspond with the preferred
answer sets of the original program.
Our approach allows the specification of \emph{dynamic} orderings, in which
preferences can appear arbitrarily within a program.
\emph{Static} orderings (in which preferences are external to a logic
program) are a trivial restriction of the general dynamic case.
First, we develop a specific approach to reasoning with
preferences, wherein the preference ordering specifies the order in which
rules are to be applied.
We then demonstrate the wide range of applicability of our framework by
showing how other approaches, among them that of Brewka and Eiter, can be
captured within our framework.
Since the result of each of these transformations is an extended logic program,
we can make use of existing implementations, such as \textsf{dlv} and
\textsf{smodels}.
To this end, we have developed a publicly available compiler as a
front-end for these programming systems.
\end{abstract}

\section{Introduction}
\label{sec:introduction}

In commonsense reasoning in general, and in logic programming in particular,
one frequently prefers one conclusion over another, or the application of
one rule over another.
For example, in buying a car one may have various desiderata in mind
(inexpensive, safe, fast, etc.) where these preferences come with varying
degrees of importance.
In legal reasoning, laws may apply in different situations, but laws may also
conflict with each other.
Conflicts are resolved by appeal to higher-level principles such as authority
or recency.
So federal laws will have a higher priority than state laws, and newer laws
will take priority over old.
Further preferences, such as authority holding sway over recency, may also
be required.

In logic programming, in \emph{basic} logic programs, which do not employ
negation as failure, there is no issue with preferences, since a given basic
logic program is guaranteed to have a single, unique, set of conclusions.
However, basic logic programs are expressively weak.
Once negation as failure is introduced, as in \emph{extended} logic
programs~\cite{gellif91a}, we are no longer guaranteed a single set of
conclusions, but rather may have several \emph{answer sets}, each giving a
feasible set of conclusions.
There is no a priori reason to accept one answer set over another, yet,
as noted above, we may have reasons to prefer one over another.

In this situation, preferences are usually expressed by a strict partial
order on the set of rules.
For example, consider the following program (technical definitions and
notation are introduced in the next section):
\[
\begin{array}{rcrcl}
  r_1 & = & \neg a             &\LPif&
  \\
  r_2 & = & \phantom{\neg} b   &\LPif&\neg a, \naf c
  \\
  r_3 & = & \phantom{\neg} c   &\LPif& \naf b\mbox{.}
\end{array}
\]
Each $r_i$ identifies the respective rule.
This program has two regular answer sets, one given by
\(
\{ \neg a, b \}
\)
and the other given by
\(
\{ \neg a, c \}%
\).
For the first answer set, rules $r_1$ and $r_2$ are applied;
for the second, $r_1$ and $r_3$.
However, assume that we have reason to prefer $r_2$ to $r_3$, expressed
by $r_3 < r_2$.
In this case we would want to obtain just the first answer set.

There have been numerous proposals for expressing preferences in
extended logic programs, including
\cite{sakino96,brewka96a,gelson97a,zhafoo97a,breeit99a,wazhli00}.
The general approach in such work has been to employ meta-formalisms for
characterising ``preferred answer sets''.
For instance, a common approach is to generate all answer sets for a
program and then, in one fashion or other, select the most preferred set(s).
Consequently, non-preferred as well as preferred answer sets are first
generated, and the preferred sets next isolated by a filtering step.
Such approaches generally have a higher complexity than the underlying
logic programming semantics
(see Section~\ref{sec:related:work} for details).

Our goal in this paper is to (i) present a general methodology and
framework for expressing and implementing preferences where 
(ii) only preferred answer sets are generated.
We do this by describing a general methodology in which a logic program with
preferences is translated into a second ``regular'' logic program, such that
the answer sets of the second program in a precise sense correspond to and
express only the preferred answer sets of the first.
This makes it possible to encode preferences within the very same logic
programming framework.
As we argue, the framework is very general and admits the encoding of
different preference strategies.
Moreover, we are able to express \emph{dynamic} preferences within a logic
program, in contrast to most previous work, which adopts an external
\emph{static} preference order.
The complexity of our approach is in the same complexity class as the
underlying logic programming semantics.
This approach is suited to a \emph{prescriptive} interpretation of
preference, wherein the preference ordering specifies the order in which
the rules are to be applied.

We begin by developing and exploring our ``preferred'' interpretation
of preference, using a strongly prescriptive interpretation of preference.
However, we also show how it is possible to encode other interpretations.
To this end, we show how Brewka and Eiter's approach to 
preference~\cite{breeit99a} can be expressed in the framework.
This encoding shows that we can also handle more descriptive-oriented
approaches.
Encodings of other approaches, such as that given 
in Wang, Zhou,~\& Lin~\shortcite{wazhli00},
are briefly described as well.

The general framework then also provides a common basis in which different
approaches can be expressed and compared.
Equivalently, the framework provides a common setting in which various
different strategies can be encoded.
Thus it provides a uniform way of capturing the different strategies that
are originally given in rather heterogeneous ways.
The translations then in a sense \emph{axiomatise} how a preference
ordering is to be understood.
While the notion of semantics as such is not our major concern,
we do discuss
various encodings in connection with the notion of
\emph{order preservation}.
In fact,  the strategy discussed in Section~\ref{sec:order:preservation} was
developed from just this concept of order preservation.

Lastly, since we translate a program into an extended logic program,
it is straightforward implementing our approach.
To this end, we have developed a translator for ordered logic programs
that serves as a front-end for the logic programming systems
\textsf{dlv}~\cite{dlv97a} and \textsf{smodels}~\cite{niesim97a}.
The possibility to utilise existing logic programming systems for  
implementation purposes is a major advantage of our framework. 
In contrast, most other approaches are based on a change of semantics and thus
require dedicated algorithms to solve the respective reasoning tasks at hand.

The next section gives background terminology and notation, while
Section~\ref{sec:framework} describes the overall methodology.
We develop our central approach in Section~\ref{sec:order:preservation} and
explore its formal properties.
Section~\ref{sec:brewka:eiter} presents our encoding of Brewka and Eiter's
approach, while Section~\ref{sec:implementation} deals with our implementation.
Section~\ref{sec:related:work} considers other work, and
Section~\ref{sec:discussion} concludes with some further issues and 
a brief discussion.
This paper regroups and strongly extends the work found in
Delgrande, Schaub,~\& 
Tompits~\shortcite{descto00a,descto00b,descto00c,descto00d,descto01a}.

\section{Definitions and Notation}
\label{sec:background}

We deal with extended logic programs~\cite{lifschitz96a} that contain the
symbol $\neg$ for \emph{classical negation} in addition to $\nafo$ used for
\emph{negation as failure}.
This allows for distinguishing between goals that fail in the sense that they
\emph{do not succeed} and goals that fail in the stronger sense that their
\emph{negation succeeds}.
Classical negation is thus also referred to as \emph{strong negation}, whilst
negation as failure is termed \emph{weak negation}.

Our formal treatment is based on propositional languages.
Let $\mathcal{A}$ be a non-empty set of symbols, called \emph{atoms}.
The choice of $\mathcal{A}$ determines the language of the programs under
consideration.
As usual, a \emph{literal}, $L$, is an expression of form $A$ or $\neg A$,
where $A$ is an atom.
We assume a possibly infinite set of such atoms.
The set of all literals over $\mathcal{A}$ is denoted by $\Lit_\mathcal{A}$.
A literal preceded by the negation as failure sign $\nafo$ is said to be
a \emph{weakly negated literal}.
A \emph{rule}, $r$, is an expression of form
\begin{equation}\label{eqn:rule}
L_0\LPif L_1,\dots,L_m,\naf L_{m+1},\dots,\naf L_n,
\end{equation}
where $n\geq m\geq 0$, and each $L_i$ $(0\leq i\leq n)$ is a literal.
The literal $L_0$ is called the \emph{head} of $r$, and the set
\(
\{L_1,\dots,L_m,$ $\naf L_{m+1},\dots,\naf L_n\}
\)
is the \emph{body} of $r$.
If $n=m$, then $r$ is said to be a \emph{basic rule};
if $n=0$, then $r$ is called a \emph{fact}.
An (\emph{extended}\/) \emph{logic program}, or simply a \emph{program},
is a finite set of rules.
A program is \emph{basic} if all rules in it are basic.
For simplicity, we associate the language of a program with the set of
literals $\Lit_\mathcal{A}$, rather than the set of all rules 
induced by~$\Lit_\mathcal{A}$.

We use $\head{r}$ to denote the head of rule $r$, and $\body{r}$ to
denote the body of~$r$.
Furthermore, let
\(
\pbody{r}
=
\{L_1,\dots, L_m\}
\)
and
\(
\nbody{r}
=
\{L_{m+1},\dots, L_n\}
\)
for $r$ as in (\ref{eqn:rule}).
The elements of $\pbody{r}$ are referred to as the \emph{prerequisites}
of $r$.
Thus, if $\pbody{r}=\emptyset$ (or $m=0$), then $r$ is said to be
\emph{prerequisite free}. 
We say that rule $r$ is \emph{defeated} by a set $X$ of literals iff
$\nbody{r}\cap X\neq\emptyset$.
Given that $\body{r}\neq\emptyset$, we also allow the situation 
where $r$ has an empty head, in which case 
$r$ is called an \emph{integrity constraint}, or \emph{constraint}, for short.
A constraint with body as in~(\ref{eqn:rule}) appearing in some 
program $\Pi$ can be regarded as a rule of form
\[
p\LPif L_1,\dots,L_m,\naf L_{m+1},\dots,\naf L_n, \naf p,
\]
where $p$ is an atom not occurring elsewhere in $\Pi$.

A set of literals $X\subseteq \Lit_\mathcal{A}$ 
is \emph{consistent} iff it does not contain a
complementary pair $A$, $\neg A$ of literals.
We say that $X$ is \emph{logically closed} iff it is either consistent or
equals $\Lit_\mathcal{A}$.
Furthermore, $X$ is \emph{closed under} a basic program $\Pi$ iff, for any
$r\in \Pi$, $\head{r}\in X$ whenever $\body{r}\subseteq X$.
In particular, if $X$ is closed under the basic program $\Pi$, 
then $\body{r}\not\subseteq X$
for any constraint $r\in\Pi$.
The smallest set of literals which is both logically closed and closed under a
basic program $\Pi$ is denoted by $\Th{\Pi}$.

Let $\Pi$ be a basic program and $X\subseteq\Lit_\mathcal{A}$ a set of literals.
The operator $\To{\Pi}$ is defined as follows:
\[
\T{\Pi}{X} = \{\head{r}\mid r\in\Pi\text{ and }\body{r}\subseteq X\}
\]
if $X$ is consistent, and $\T{\Pi}{X} = \Lit_\mathcal{A}$ otherwise.
Iterated applications of $\To{\Pi}$ are written as $\Tindo{j}{\Pi}$ for
$j\geq 0$, where
\(
\Tind{0}{\Pi}{X}=X
\)
and
\(
\Tind{i}{\Pi}{X}=\T{\Pi}{\Tind{i-1}{\Pi}{X}}
\)
for $i\geq 1$.
It is well-known that
\(
\Th{\Pi}=\bigcup_{i\geq 0}\Tind{i}{\Pi}{\emptyset}
\),
for any basic program $\Pi$.

Let $r$ be a rule.
Then $\reductr{r}$ denotes the basic rule obtained from $r$ by deleting all
weakly negated literals in the body of $r$, that is,
$\reductr{r}=\head{r}\LPif\pbody{r}$.
Accordingly, we define for later usage
$\reductBE{r}=\head{r}\LPif\nbody{r}$.
The \emph{reduct}, $\reduct{\Pi}{X}$, of a program $\Pi$ relative to a
set $X$ of literals is defined by
\[
\reduct{\Pi}{X}
=
\{\reductr{r}\mid r\in\Pi\text{ and }\nbody{r}\cap X=\emptyset\}\mbox{.}
\]
In other words, $\reduct{\Pi}{X}$ is obtained from $\Pi$ by:
\begin{enumerate}
\item
deleting any $r\in\Pi$ which is defeated by $X$, and
\item
deleting each weakly negated literal occurring in the bodies of the
remaining rules.
\end{enumerate}
We say that a set $X$ of literals is an \emph{answer set} of a program $\Pi$
iff
$\Th{\reduct{\Pi}{X}}=X$.
Clearly, for each answer set $X$ of a program $\Pi$, it holds that
\(
X=\bigcup_{i\geq 0}\Tind{i}{\reduct{\Pi}{X}}{\emptyset}
\).

The set $\GR{\Pi}{X}$ of all \emph{generating rules} of an answer set $X$ from 
$\Pi$ is given by
\[
\GR{\Pi}{X}
=
\{r\in\Pi\mid\pbody{r}\subseteq X\text{ and }\nbody{r}\cap X=\emptyset\}\mbox{.}
\]
That is, $\GR{\Pi}{X}$ comprises all rules $r\in\Pi$ such that $r$ is not
defeated by $X$ and each prerequisite of $r$ is in $X$.
Finally, a sequence
\(
\langle r_i\rangle_{i\in I}
\)
of rules is \emph{grounded} iff, for all $i\in I$,
$\pbody{r_i}\subseteq\{\head{r_j}\mid j<i\}$ providing
$\{\head{r_j}\mid j<i\}$ is consistent.
We say that a rule $r$ is \emph{grounded}%
\footnote{Note that some authors (for example Niemel\"a~\& 
  Simons~\shortcite{niesim97a}) use \emph{grounded} to refer to the process 
  of eliminating variables from a rule by replacing it with its set of 
  ground instances.
  We use the term \emph{grounded} in reference to grounded enumerations only.
  When we discuss replacing variables with ground terms in
  Section~\ref{sec:implementation}, we will refer to an \emph{instantiation}
  of a rule.}
in a set $R$ of rules  iff
there is an grounded enumeration of $R$ and 
$\pbody{r}\subseteq\{\head{r}\mid r\in R\}$.

The answer set semantics for extended logic programs has been defined
in Gelfond~\& Lifschitz~\shortcite{gellif91a} as a generalisation of the stable model
semantics~\cite{gellif88b} for \emph{general logic programs} (\iec programs
not containing classical negation, $\neg$).
The reduct $\reduct{\Pi}{X}$ is often called the
\emph{Gelfond-Lifschitz reduction}. 

\section{From Ordered to Tagged Logic Programs}
\label{sec:framework}

A logic program $\Pi$ over a propositional language $\mathcal{L}$ is said to
be \emph{ordered} if $\mathcal{L}$ contains the following pairwise disjoint
categories:
\begin{itemize}
\item a set $\nameset$ of terms serving as \emph{names} for rules;
  
\item a set $\mathcal{A}$ of regular (propositional) atoms of a program; and
  
\item a set $\mathcal{A}_\prec$ of \emph{preference atoms} $\PREC{s}{t}$,
  where $s,t\in\nameset$ are names.
\end{itemize}
We assume furthermore a bijective%
\footnote{In practice, function $n$ is only required to be injective since
only rules participating in the preference relation require names.}
function $\namef{\cdot}$ assigning to each rule $r\in\Pi$ a name
$\namef{r}\in\nameset$.
To simplify our notation, we usually write $\name{r}$ instead of
$\namef{r}$
(and we sometimes abbreviate $n_{r_i}$ by $n_i$).
Also, the relation $t=\namef{r}$ is written as $t:r$, leaving the naming
function $\namef{\cdot}$ implicit.
The elements of $\mathcal{A}_\prec$ express preferences among rules. 
Intuitively,
$\PREC{\name{r}}{\name{r'}}$ asserts that $r'$ has ``higher priority'' than $r$.
Thus, $r'$ is viewed as having precedence over $r$.
That is,
$r'$ should, in some sense, always be considered ``before'' $r$.
(Note that some authors use $\prec$ or $\PRECMo$ in the opposite sense from us.)

Formally, given an alphabet $\mathcal{A}$,
an ordered logic program can be understood as a triple
\(
(\Pi,\nameset,\namemap)
\),
where $\Pi$ is an extended logic program over
\(
\mathcal{L}_{\mathcal{A}\cup\mathcal{A}_\prec}
\)
and $\namemap$ is a bijective function between $\Pi$ and the set of names
$\nameset$.%
\footnote{Note that $\mathcal{A}_\prec$ is determined by $\nameset$.}
In what follows,
we leave the set of names $\nameset$ and the naming function $\namemap$
implicit and rather associate the notion of an ordered logic program with the
underlying extended logic program $\Pi$.

It is important to note that
we impose no restrictions on the occurrences of preference atoms.
This allows for expressing preferences in a very flexible, dynamic way.
For instance, we may specify
\[
\PREC{\name{r}}{\name{r'}}\LPif p, \naf q
\]
where $p$ and $q$ may themselves be (or rely on) preference atoms.

A special case is given by programs containing preference atoms only among
their facts.
We say that a logic program $\Pi$ over
\(
\mathcal{L}
\)
is \emph{statically ordered}
if it is of form
\(
\Pi=\Pi'\cup\Pi''
\),
where $\Pi'$ is an extended logic program over $\mathcal{L}_{\mathcal{A}}$
and
\(
\Pi''
\subseteq
\{(\PREC{\name{r}}{\name{r'}})\LPif{}\mid r,r'\in\Pi'\}.
\)
The static case can be regarded as being induced from an external order,
$\PRECMo$, where the relation \PRECM{r}{r'} between two rules holds iff
the fact $(\PREC{\name{r}}{\name{r'}})\LPif$ is included in the ordered
program.
We make this explicit by denoting a statically ordered program $\Pi$ as a pair
\(
(\Pi',\PRECMo)
\),
representing the program
\(
\Pi'\cup\{(\PREC{\name{r}}{\name{r'}})\LPif{}\mid r,r'\in\Pi',\PRECM{r}{r'}\}
\).
We stipulate for each statically ordered program $(\Pi,\PRECMo)$ that
$\PRECMo$ is a strict partial order.
The static concept of preference corresponds to most previous approaches
to preference handling in logic programming and nonmonotonic reasoning,
where the preference information is specified as a fixed relation at the
meta-level (cf.\ \cite{baahol93a,brewka94a,zhafoo97a,breeit99a,wazhli00}).

The idea behind our methodology for compiling preferences is
straightforward.
Given a preference handling strategy $\sigma$,
our approach provides a mapping $\mathcal{T}_\sigma$ that transforms an
ordered logic program $\Pi$ into a standard logic program
$\mathcal{T}_\sigma(\Pi)$, such that the answer sets of $\Pi$ preferred by
$\sigma$ correspond to the (standard) answer sets of $\mathcal{T}_\sigma(\Pi)$.
Intuitively, the translated program $\mathcal{T}_\sigma(\Pi)$ is constructed
in such a way that the resulting answer sets comply with $\sigma$'s
interpretation of the preference information
(induced by the original program $\Pi$).
This is achieved by adding sufficient control elements to the rules of $\Pi$
that will guarantee that successive rule applications are in accord with the
intended order.
Such control elements, or \emph{tags} for short, are given through newly
introduced atoms that allow us to detect and control rule applications within
the object language.
Hence, a translation $\mathcal{T}_\sigma$ maps ordered logic programs onto
standard logic programs whose language is extended by appropriate tags.

Given the relation $\PRECM{r}{r'}$ (or the atom $\PREC{\name{r}}{\name{r'}}$,
respectively), we want to ensure that $r'$ is considered before $r$,
in the sense that rule $r'$ is known to be applied or blocked ahead of $r$
(with respect to the order of rule application).
We do this by first translating rules so that the order of rule application
can be explicitly controlled.
For this purpose, we need to be able to detect when a rule applies
or when a rule is defeated or ungroundable;
as well we need to be able to control the application of a rule based on
other antecedent conditions.

First, we introduce, for each rule $r$ in the given program $\Pi$,
a new special-purpose atom \applied{\name{r}} to detect the case where
a rule's applicability conditions are satisfied.
For instance, the rule
\[
r_{42} \quad = \quad p\LPif{}q,\naf w
\]
is mapped onto the rule
\begin{eqnarray}
  \label{eq:rule:applied:i}
  \applied{n_{42}}&\LPif&q,\naf w\ \mbox{.}
\end{eqnarray}
The consequent of the original rule, $p$, is replaced by the tag
$\applied{n_{42}}$, just recording the fact that $r_{42}$ is applicable.
The addition of
\begin{eqnarray}
  \label{eq:rule:applied:ii}
  p&\LPif&\applied{n_{42}}\ 
\end{eqnarray}
then ``restores'' the effect of the original rule.
This mapping separates the applicability of a rule from its actual application.

Second, for detecting when a rule's applicability conditions cannot be satisfied
(\iec the rule will be blocked), we introduce, for each rule $r$ in $\Pi$,
another new atom \blocked{\name{r}}.
Unlike the above, however, there are two cases for a rule $r$ not to be applied:
it may be that some literal in $\pbody{r}$ does not appear in the answer set,
or it may be that a literal in $\nbody{r}$ is in the answer set.
For rule $r_{42}$ we thus get two rules:
\begin{eqnarray}
  \label{eq:rules:blocked:i}
  \blocked{n_{42}}&\LPif&\naf q\ ,
  \\
  \label{eq:rules:blocked:ii}
  \blocked{n_{42}}&\LPif&     w\ \mbox{.}
\end{eqnarray}
As made precise in Section~\ref{sec:order:preservation},
this tagging technique provides us already with complete information about the
applicability status of each rule $r\in\Pi$ with respect to any answer set
$X$ of the ``tagged program'' $\Pi'$:
\begin{equation}
  \label{eq:complete:info}
  \applied{n_r} \in X
  \text{ iff }
  \blocked{n_r} \not\in X
  \ \mbox{.}
\end{equation}
(Informally, the ``tagged program'' $\Pi'$ is obtained from $\Pi$ by treating
each rule in $\Pi$ as described above for $r_{42}$, that is, by replacing
each rule by the appropriate rules corresponding to 
(\ref{eq:rule:applied:i})--(\ref{eq:rules:blocked:ii}).)
Although this is all relatively straightforward,
it can be seen that via these tags we can detect when a rule is or is not
applied.
We note that a more fine-grained approach is obtainable by distinguishing the
two causes of blockage by means of two different tags, like \blockedP{n} and
\blockedJ{n}.

Lastly, for controlling the application of rule $r$, we introduce an atom
\ok{\name{r}} among the body literals of $r$.
Then, clearly the transformed rule can (potentially) be applied only if
\ok{n_r} is asserted.
More generally, we can combine this with the preceding mapping and so have
\ok{n_r} appear in the prerequisite of rules 
(\ref{eq:rule:applied:i}), (\ref{eq:rules:blocked:i}), and (\ref{eq:rules:blocked:ii}).
In our example, we get
\begin{eqnarray*}
    \applied{n_{42}}&\LPif&\ok{n_{42}},q,\naf w\ ,
    \\
    \blocked{n_{42}}&\LPif&\ok{n_{42}},  \naf q\ ,
    \\
    \blocked{n_{42}}&\LPif&\ok{n_{42}},       w\ \mbox{.}
\end{eqnarray*}
With this tagging, (\ref{eq:complete:info}) can be refined so that for each
rule $r\in\Pi$ and any answer set $X$ of the ``tagged program'' $\Pi''$
we have
\[
\text{if }
\ok{n_r}\in X
\ ,
\text{ then }
\applied{n_r} \in X
\text{ iff }
\blocked{n_r} \not\in X
\ \mbox{.}
\]

For preference handling, informally,
we conclude \ok{n_r} for rule $r$ just if it is ``$\oksym$'' with
respect to every $\PRECMo$-greater rule $r'$.
The exact meaning of this process is fixed by the given preference handling
strategy.
For instance,
in the strategy of Section~\ref{sec:order:preservation},
\ok{n_r} is concluded just when each $\PRECMo$-greater rule $r'$ is known to be
blocked or applied.
Using tags, a single preference like
\(
\PRECM{r_2}{r_4} 
\),
saying that $r_4$ is preferred to $r_2$,
could thus be encoded directly in the following way
(together with the appropriately tagged rules):
\begin{eqnarray*}
    \ok{n_4}&\LPif&\ ,
    \\
    \ok{n_2}&\LPif&\applied{n_4}\ ,
    \\
    \ok{n_2}&\LPif&\blocked{n_4}\ \mbox{.}
\end{eqnarray*}

Similar to the addition of $\oksym$-literals to the rules' prerequisites,
we can introduce literals of form
\(
\naf\ko{n}
\)
among the weakly negated body literals of a rule.
Now the transformed rule behaves exactly as the original, except that we can
defeat (or ``knock out'') this rule by asserting \ko{n}.
For instance, adding $\naf\ko{n_{42}}$ to the body of
(\ref{eq:rule:applied:ii}) would allow us to assert the applicability of
$r_{42}$ without asserting its original consequent, whenever \ko{n_{42}} is
derivable.
This manner of blocking a rule's application has of course appeared earlier in
the literature, where $\kosym$ was most commonly called $ab$ (for
``abnormal'' \cite{mccarthy86}).

\section{Order Preserving Logic Programs}
\label{sec:order:preservation}

This section elaborates upon a fully prescriptive strategy for preference
handling, having its roots in Delgrande~\& Schaub~\shortcite{delsch97a}.
The idea of this approach is to select those answer sets of the program that
can be generated in an ``order preserving way''.
This allows us to enforce the ordering information during the construction of
answer sets.

In order to guarantee that rules are treated in an ``order preserving way''
our strategy stipulates that the question of applicability must be settled for
higher ranked rules before that of lower ranked rules.
In this way, a rule can never be defeated or grounded by lower ranked rules.
While a lower ranked rule may of course depend on the presence (or
absence) of a literal appearing in the head of a higher ranked rule, the
application of lower ranked rules is independent of \emph{the fact} that
a higher ranked rule has been applied or found to be blocked.
This is important to guarantee the selection among existing answer sets,
since otherwise not \emph{all} rules of the original program are considered
(cf.\ Proposition~\ref{thm:results:i} and~\ref{thm:be:results}).

Let us illustrate this by means of the following statically ordered program
taken from Baader~\& Hollunder~\shortcite{baahol93a}:%
\footnote{Letters $b,p,f,w$ stand, as usual, for 
  \emph{birds}, \emph{penguins}, \emph{flies}, and \emph{wings},
  respectively.
  Baader~\& Hollunder~\shortcite{baahol93a} formulate their 
  approach in default logic \cite{reiter80}.}
\begin{equation}
  \label{eq:baader:hollunder}
  \begin{array}[t]{rcrcl}
    r_1 & = & \neg f &\LPif& p, \naf      f
    \\
    r_2 & = &      w &\LPif& b, \naf \neg w
    \\
    r_3 & = &      f &\LPif& w, \naf \neg f
    \\
    r_4 & = &      b &\LPif& p
    \\
    r_5 & = &      p &\LPif& 
  \end{array}
  \qquad\qquad
  \PRECM{r_2}{r_1}
  \ \mbox{.}
\end{equation}
Program $\Pi_{\ref{eq:baader:hollunder}}=\{r_1,\dots,r_5\}$ has two answer sets:
$X_1=\{p,b,w,\neg f\}$ and $X_2=\{p,b,w,f\}$.
Consider
\[
\Pi^{X_1}_{\ref{eq:baader:hollunder}}=\{
  \begin{array}[t]{rcl}
    \neg f &\LPif& p
    \\
         w &\LPif& b
    \\
           &     &
    \\
         b &\LPif& p
    \\
         p &\LPif& \multicolumn{1}{r}{\quad\}}
  \end{array}
\qquad\text{ and }\qquad
\Pi^{X_2}_{\ref{eq:baader:hollunder}}=\{
  \begin{array}[t]{rcl}
           &     & 
    \\
         w &\LPif& b             
    \\
         f &\LPif& w             
    \\
         b &\LPif& p
    \\
         p &\LPif& \multicolumn{1}{r}{\quad\}}
    \ \mbox{.}
  \end{array}
\]
This gives rise to the following constructions:
\[
\begin{array}[t]{rcl}
\Tind{1}{\Pi^{X_1}_{\ref{eq:baader:hollunder}}}{\emptyset}&=&\{p\}
\\
\Tind{2}{\Pi^{X_1}_{\ref{eq:baader:hollunder}}}{\emptyset}&=&\{p,b,\neg f\}
\\
\Tind{3}{\Pi^{X_1}_{\ref{eq:baader:hollunder}}}{\emptyset}&=&\{p,b,\neg f,w\}
\end{array}
\qquad\text{ and }\qquad
\begin{array}[t]{rcl}
\Tind{1}{\Pi^{X_2}_{\ref{eq:baader:hollunder}}}{\emptyset}&=&\{p\}
\\
\Tind{2}{\Pi^{X_2}_{\ref{eq:baader:hollunder}}}{\emptyset}&=&\{p,b\}
\\
\Tind{3}{\Pi^{X_2}_{\ref{eq:baader:hollunder}}}{\emptyset}&=&\{p,b,w\}
\\
\Tind{4}{\Pi^{X_2}_{\ref{eq:baader:hollunder}}}{\emptyset}&=&\{p,b,w,f\}
\ \mbox{.}
\end{array}
\]
Let us now take a closer look at the grounded enumerations associated with
$X_1$ and $X_2$.
The construction of $X_1$ leaves room for three grounded enumerations:
\begin{equation}
  \label{eq:sequences:one}
  \langle
  r_5,r_4,r_2,r_1
  \rangle,
  \langle
  r_5,r_4,r_1,r_2
  \rangle,
  \text{ and }
  \langle
  r_5,r_1,r_4,r_2
  \rangle
  \ \mbox{.}
\end{equation}
According to our strategy only the last two enumerations are order preserving
since they reflect the fact that $r_1$ is treated before $r_2$.
The construction of $X_2$ induces a single grounded enumeration
\begin{equation}
  \label{eq:sequences:two}
  \langle
  r_5,r_4,r_2,r_3
  \rangle
  \ \mbox{.}
\end{equation}
Unlike the above, $r_1$ does not occur in this sequence.
So the question arises whether it was blocked in an order preserving way.
In fact, once $r_5$ and $r_4$ have been applied both $r_1$ and $r_2$ are
applicable,
that is,
their prerequisites have been derived and neither of them is defeated at this
point.
Clearly, the application of $r_2$ rather than $r_1$ violates the preference
\PRECM{r_2}{r_1} (in view of the given strategy).
Therefore, only $X_1$ can be generated in an order preserving way,
which makes our strategy select it as the only preferred
answer set of $(\Pi_{\ref{eq:baader:hollunder}},\PRECMo)$.

Let us make this intuition precise for statically ordered programs.
%
\begin{definition}\label{def:order:preserving}
  Let $(\Pi,\PRECMo)$ be a statically ordered program
  and
  let $X$ be a consistent answer set of~$\Pi$.

  Then, $X$ is called  \emph{$\PRECMo$-preserving},
  if there exists an enumeration
  \(
  \langle r_i\rangle_{i\in I}
  \)
  of\/ $\GR{\Pi}{X}$
  such that, for every $i,j\in I$, we have that:
  \begin{enumerate}
  \item\label{def:order:preserving:zero}
    $\pbody{r_i}\subseteq\{\head{r_k}\mid k<i\}$;
  \item\label{def:order:preserving:one}
    if $\PRECM{r_i}{r_j}$, then $j<i$;
    \quad and
  \item\label{def:order:preserving:two}
    if
    $\PRECM{r_i}{r'}$
    and 
    \(
    r'\in {\Pi\setminus\GR{\Pi}{X}},
    \)
    then
    \begin{enumerate}
    \item\label{def:order:preserving:two:a}
      $\pbody{r'}\not\subseteq X$
      \quad or
    \item\label{def:order:preserving:two:b}
      $\nbody{r'}\cap\{\head{r_k}\mid k<i\}\neq\emptyset$.
    \end{enumerate}
  \end{enumerate}
\end{definition}
%
While Condition~\ref{def:order:preserving:one} guarantees that all generating
rules are applied according to the given order,
Condition~\ref{def:order:preserving:two} assures that any preferred yet
inapplicable rule is either blocked due to the non-derivability of its
prerequisites or because it is defeated by higher ranked or unrelated rules.

Condition~\ref{def:order:preserving:zero} makes the property of
\emph{groundedness} explicit.
Although any standard answer set is generated by a grounded sequence of
rules, we will see in the following sections that this property is sometimes
weakened when preferences are at issue.

For simplicity, our motivation was given in a static setting.
The generalisation of order preservation to the dynamic case is discussed in
Section~\ref{sec:properties}.

\subsection{Encoding}
\label{sec:encoding}

We now show how the selection of $\PRECMo$-preserving answer sets can be
implemented via a translation of ordered logic programs back into standard
programs.
Unlike the above, this is accomplished in a fully dynamic setting, as put
forward in Section~\ref{sec:framework}.

Given an ordered program $\Pi$ over $\mathcal{L}$,
let $\mathcal{L}^{+}$ be the language obtained from $\mathcal{L}$ by adding,
for each $r,r'\in\Pi$, new pairwise distinct propositional atoms
\applied{\name{r}}, \blocked{\name{r}}, \ok{\name{r}}, and
\rdy{\name{r}}{\name{r'}}.%
\footnote{\rdy{\cdot}{\cdot}, for ``ready'', are auxiliary atoms for
  \ok{\cdot}.}
Then, our translation $\mathcal{T}$ maps an ordered program $\Pi$ over
$\mathcal{L}$ into a regular program $\mathcal{T}(\Pi)$ over $\mathcal{L}^{+}$
in the following way.
%
\begin{definition}\label{def:compilation}
Let $\Pi = \{r_1,\dots, r_k\}$ be an ordered logic program over $\mathcal{L}$.

Then,
the logic program $\mathcal{T}(\Pi)$ over $\mathcal{L}^{+}$ is defined as
\[
\mathcal{T}(\Pi)
=
\mbox{$\bigcup_{r\in\Pi}$}\tau(r)
\ ,
\]
where the set $\tau(r)$ consists of the following rules,
for
$L^{+}\in\pbody{r}$,
$L^{-}\in\nbody{r}$,
and
$r',r'' \in \Pi$~:
\[
\begin{array}{rrcl}
  \RULE{\ap{1}{r}}
       {\head{r}}
       {\applied{\name{r}} }
  \\
  \RULE{\ap{2}{r}}
       {\applied{\name{r}}}
       {\ok{\name{r}},\body{r}}
  \\
  \RULE{\bl{1}{r}{L^{+}}}
       {\blocked{\name{r}}}
       {\ok{\name{r}}, \naf L^{+}}
  \\
  \RULE{\bl{2}{r}{L^{-}}}
       {\blocked{\name{r}}}
       {\ok{\name{r}},L^{-}}
  \\[2ex]
  \RULE{\cokt{1}{r}}
       {\ok{\name{r}}}
       {\rdy{\name{r}}{\name{r_1}},\dots,\rdy{\name{r}}{\name{r_k}}}
  \\
  \RULE{\cok{2}{r}{r'}}
       {\rdy{\name{r}}{\name{r'}}}
       {\naf(\PREC{\name{r}}{\name{r'}})}
  \\
  \RULE{\cok{3}{r}{r'}}
       {\rdy{\name{r}}{\name{r'}}}
       {(\PREC{\name{r}}{\name{r'}}),\applied{\name{r'}}}
  \\
  \RULE{\cok{4}{r}{r'}}
       {\rdy{\name{r}}{\name{r'}}}
       {(\PREC{\name{r}}{\name{r'}}),\blocked{\name{r'}}}
  \\[2ex]
  \RULE{t(r,r',r'')}
       {\PREC{\name{r}}{\name{r''}}}
       {\PREC{\name{r}}{\name{r'}},\PREC{\name{r'}}{\name{r''}}}
  \\
  \RULE{as(r,r')}
       {{\neg(\PREC{\name{r'}}{\name{r}})}}
       {\PREC{\name{r}}{\name{r'}}} \ \mbox{.}
\end{array}
\]
\end{definition}
%
We write
\(
\mathcal{T}(\Pi,\PRECMo)
\)
instead of 
\(
\mathcal{T}(\Pi')
\)
whenever we deal with a statically ordered program
\(
\Pi'=\Pi\cup\{(\PREC{\name{r}}{\name{r'}})\LPif{}\mid r,r'\in\Pi,\PRECM{r}{r'}\}
\).

The first four rules of $\tau(r)$ express applicability 
and blocking conditions of the original rules:
For each rule $r\in\Pi$, we obtain two rules, \ap{1}{r} and \ap{2}{r},
along with $|\pbody{r}|$ rules of form \bl{1}{r}{L^{+}} and $|\nbody{r}|$
rules of form \bl{2}{r}{L^{-}}.
A rule $r$ is thus represented by $|\body{r}|+2$ rules in $\tau(r)$.

The second group of rules encodes the strategy for handling preferences.
The first of these rules, \cokt{1}{r}, ``quantifies'' over the rules in
$\Pi$,
in that we conclude $\ok{\name{r}}$ just when $r$ is ``ready'' to
be applied with respect to all the other rules.
This is necessary when dealing with dynamic preferences since preferences may
vary depending on the corresponding answer set.
The three rules \cok{2}{r}{r'}, \cok{3}{r}{r'}, and \cok{4}{r}{r'} specify
the pairwise dependency of rules in view of the given preference ordering:
For any pair of rules $r$, $r'$,
we derive $\rdy{\name{r}}{\name{r'}}$ whenever $\PREC{\name{r}}{\name{r'}}$ fails 
to hold, or else whenever either $\applied{\name{r'}}$ or
$\blocked{\name{r'}}$ is true.
This allows us to derive $\ok{\name{r}}$, indicating that $r$ may potentially
be applied whenever we have for all $r'$ with $\PREC{\name{r}}{\name{r'}}$ that
$r'$ has been applied or cannot be applied.
It is important to note that this is only one of many strategies for dealing
with preferences: different strategies are obtainable by changing the
specification of \ok{\cdot} and \rdy{\cdot}{\cdot},
as we will see in the subsequent sections.

The last group of rules renders the order information a strict partial order.

Given that occurrences of $\name{r}$ can be represented by variables,
the number of rules in $\mathcal{T}(\Pi)$ is limited by
\(
(|\Pi|\cdot(\mathit{max}_{r\in\Pi} |\body{r}|+2)) + 4 + 2
\).
A potential bottleneck in this translation is clearly rule \cokt{1}{r} whose
body may contain $|\Pi|$ literals.
In practice, however, this can always be reduced to the number of rules
involved in the preference handling (cf.\ Section~\ref{sec:implementation}).

As an illustration of our approach, consider the following program,
$\Pi_{\ref{eq:example}}$:
\begin{equation}
  \label{eq:example}
  \begin{array}{rcrcl}
    r_1 & = & \neg a             &\LPif&
    \\
    r_2 & = & \phantom{\neg} b   &\LPif&\neg a, \naf c
    \\
    r_3 & = & \phantom{\neg} c   &\LPif& \naf b
    \\
    r_4 & = & \PREC{n_3}{n_2}    &\LPif& \naf d \ ,
  \end{array}
\end{equation}
where $\name{i}$ denotes the name of rule $r_i$ $(i=1,\dots,4)$.
This program has two regular answer sets:
\(
X_1=\{\neg a,b,\PREC{n_3}{n_2}\}
\)
and
\(
X_2=\{\neg a,c,\PREC{n_3}{n_2}\}
\).
While \PREC{n_3}{n_2} is in both answer sets, it is only respected by $X_1$,
in which $r_2$ overrides $r_3$.

In fact, $X_1$ corresponds to the single answer set obtained from
${\mathcal{T}(\Pi_{\ref{eq:example}})}$.
To see this, observe that for any
\(
X\subseteq\{\head{r}\mid r\in\mathcal{T}(\Pi_{\ref{eq:example}})\}
\),
we have
\(
\PREC{n_i}{n_j}\not\in X
\)
whenever $(i,j)\neq (3,2)$.
We thus get for such $X$ and $i,j$ that
\(
\rdy{\name{i}}{\name{j}}\in\Tind{1}{\reduct{\mathcal{T}(\Pi_{\ref{eq:example}})}{X}}{\emptyset}
\)
by (reduced) rules $\cok{2}{r_i}{r_j}^{+}$, and so
\(
\ok{\name{i}}\in\Tind{2}{\reduct{\mathcal{T}(\Pi_{\ref{eq:example}})}{X}}{\emptyset}
\)
via rule $\cokt{1}{r_i}^{+}=\cokt{1}{r_i}$ for $i=1,2,4$.
Analogously, we get that
\(
\applied{\name{1}},\applied{\name{4}},
\neg a, \text{ and } \PREC{n_3}{n_2}
\)
belong to any answer set of ${\mathcal{T}(\Pi_{\ref{eq:example}})}$.

Now consider the following rules from $\mathcal{T}(\Pi_{\ref{eq:example}})$:
\[
\begin{array}{rcrcl}
  \ap{2}{r_2}&:&  \applied{\name{2}}&\LPif& \ok{\name{2}},\neg a, \naf c 
  \\
  \bl{1}{r_2}{\neg a}&:&  \blocked{\name{2}}&\LPif& \ok{\name{2}}, \naf \neg a
  \\
  \bl{2}{r_2}{c}&:&  \blocked{\name{2}}&\LPif& \ok{\name{2}},c 
  \\[1ex]
  \ap{2}{r_3}&:&  \applied{\name{3}}&\LPif& \ok{\name{3}}, \naf b 
  \\
  \bl{2}{r_3}{b}&:&  \blocked{\name{3}}&\LPif& \ok{\name{3}},b 
  \\[1ex]
  \cok{3}{r_3}{r_2}&:&  \rdy{\name{3}}{\name{2}}&\LPif& (\PREC{\name{3}}{\name{2}}), \applied{\name{2}}
  \\
  \cok{4}{r_3}{r_2}&:&  \rdy{\name{3}}{\name{2}}&\LPif& (\PREC{\name{3}}{\name{2}}), \blocked{\name{2}} \ \mbox{.}
\end{array}
\]
Given $\ok{\name{2}}$ and $\neg a$, rule \ap{2}{r_2} leaves us with the
choice between
\(
c\not\in X
\)
or
\(
c\in X
\).

First, assume $c\not\in X$.
We get \applied{\name{2}} from
\(
\ap{2}{r_2}^{+}\in\reduct{\mathcal{T}(\Pi_{\ref{eq:example}})}{X}
\).
Hence, we get $b$, \rdy{\name{3}}{\name{2}}, and finally \ok{\name{3}},
which results in \blocked{\name{3}} via \bl{2}{r_3}{b}.
Omitting further details, this yields an answer set containing $b$ while 
excluding $c$.

Second, assume $c\in X$.
This eliminates \ap{2}{r_2} when turning $\mathcal{T}(\Pi_{\ref{eq:example}})$ 
into $\reduct{\mathcal{T}(\Pi_{\ref{eq:example}})}{X}$.
Also, \bl{1}{r_2}{\neg a} is defeated since $\neg a$ is derivable.
\bl{2}{r_2}{c} is inapplicable, since $c$ is only derivable
(from \applied{\name{3}} via \ap{1}{r_3})
in the presence of \ok{\name{3}}.
But \ok{\name{3}} is not derivable since neither
\applied{\name{2}} nor \blocked{\name{2}} is derivable.
Since this circular situation is unresolvable, there is no answer set
containing $c$.

Whenever one deals exclusively with static preferences,
a simplified version of translation $\mathcal{T}$ can be used.
Given a statically ordered program $(\Pi,\PRECMo)$, a static translation
$\mathcal{T}'(\Pi,\PRECMo)$ is obtained from $\mathcal{T}(\Pi,\PRECMo)$
by 
\begin{enumerate}
\item 
replacing 
{\cokt{1}{r}},
{\cok{2}{r}{r'}},
{\cok{3}{r}{r'}}, and
{\cok{4}{r}{r'}}
in $\tau(r)$ by
\[
\begin{array}{rrcl}
  \RULE{\cokt{1}{r}}
       {\ok{\name{r}}}
       {\rdy{\name{r}}{\name{s_1}},\dots,\rdy{\name{r}}{\name{s_k}}}
  \\
  \RULE{\cok{3}{r}{s_i}}
       {\rdy{\name{r}}{\name{s_i}}}
       {\applied{\name{s_i}}}
  \\
  \RULE{\cok{4}{r}{s_i}}
       {\rdy{\name{r}}{\name{s_i}}}
       {\blocked{\name{s_i}}}
\end{array}
\]
for $i=1,\dots,k$ where $\{s_1,\dots,s_k\}=\{s\mid\PRECM{r}{s}\}$,
and
\item deleting ${t(r,r',r'')}$ and ${as(r,r')}$ in $\tau(r)$.
\end{enumerate} 
Note that the resulting program $\mathcal{T}'(\Pi,\PRECMo)$ does not contain
any preference atoms anymore.
Rather the preferences are directly ``woven'' into the resulting program in
order to enforce order preservation.

\subsection{Formal Elaboration}
\label{sec:properties}

Our first result ensures that the dynamically generated preference information
enjoys the usual properties of partial orderings.
To this end, we define the following relation:
for each set $X$ of literals and every $r,r'\in\Pi$, the relation
$\PRECMi{X}{r}{r'}$ holds iff $\PREC{\name{r}}{\name{r'}}\in X$.
%
\begin{proposition}\label{thm:strict:order}
  Let $\Pi$ be an ordered logic program and $X$ a consistent answer set
  of~$\mathcal{T}(\Pi)$.
  
  Then, we have:
  \begin{enumerate}
  \item\label{thm:strict:order:one}
    $\PRECMoi{X}$ is a strict partial order; and
  \item\label{thm:strict:order:two}
    if\/ $\Pi$ has only static preferences,
    then ${\PRECMoi{X}} = {\PRECMoi{Y}}$,
    \ for any answer set $Y$ of~$\mathcal{T}(\Pi)$.
  \end{enumerate}
\end{proposition}
%
For statically ordered programs, one can show that the addition of preferences
can never increase the number of $<$-preserving answer sets.

The next result ensures that we consider \emph{all} rules and that we gather
complete knowledge on their applicability status.
%
\begin{proposition}\label{thm:results:i}
  Let $\Pi$ be an ordered logic program 
  and
  $X$ a consistent answer set of~$\mathcal{T}(\Pi)$.

  Then,
  we have 
  for any 
  \(
  r\in \Pi
  \):
  \begin{enumerate}
  \item  \label{l:results:2}
    $\ok{\name{r}}\in X$; and
  \item \label{l:results:3}
    $\applied{\name{r}}\in X$
    iff\/
    $\blocked{\name{r}}\not\in X$.
  \end{enumerate}
\end{proposition}
%
The following properties shed light on the program induced by translation
$\mathcal{T}$; they elaborate upon the logic programming operator
\To{\reduct{\mathcal{T}(\Pi)}{X}} of a reduct $\reduct{\mathcal{T}(\Pi)}{X}$~:
%
\begin{proposition}\label{thm:results:ii}
  Let $\Pi$ be an ordered logic program
  and
  $X$ a consistent answer set of~$\mathcal{T}(\Pi)$.

  Let
  \(
  \Omega=
  \reduct{\mathcal{T}(\Pi)}{X}
  \). 
  Then,
  we have 
  for any 
  \(
  r\in \Pi
  \):
  \begin{enumerate}
  \item   \label{l:results:4}
    if $r$ is not defeated by $X$, $\ok{\name{r}}\in \Tind{i}{\Omega}{\emptyset}$, 
    and
    $\pbody{r}\subseteq\Tind{j}{\Omega}{\emptyset}$, then 
    $\applied{\name{r}}\in\Tind{\max(i,j)+1}{\Omega}{\emptyset}$;
  \item   \label{l:results:5}
    $\ok{\name{r}}\in \Tind{i}{\Omega}{\emptyset}$
    and
    $\pbody{r}\not\subseteq X$
    implies 
    $\blocked{\name{r}}\in \Tind{i+1}{\Omega}{\emptyset}$;
  \item   \label{l:results:6}
    if $r$ is defeated by $X$ and
    $\ok{\name{r}}\in \Tind{i}{\Omega}{\emptyset}$, then
    $\blocked{\name{r}}\in \Tind{j}{\Omega}{\emptyset}$
    for some $j>i$;
  \item   \label{l:results:7}
    $\ok{\name{r}}\not\in \Tind{i}{\Omega}{\emptyset}$
    implies 
    $\applied{\name{r}}\not\in \Tind{j}{\Omega}{\emptyset}$
    and
    $\blocked{\name{r}}\not\in \Tind{k}{\Omega}{\emptyset}$
    for all $j,k< i+2$.
  \end{enumerate}
\end{proposition}

The next result captures the prescriptive principle underlying the use of
$\oksym$-literals and their way of guiding the inference process along
the preference order.
In fact,
it shows that their derivation strictly follows the partial order induced by
the given preference relation.
Clearly, this carries over to the images of the original program rules,
as made precise in the subsequent corollary.
%
\begin{theorem}\label{thm:order:implementing}
  Let $\Pi$ be an ordered logic program, 
  $X$ a consistent answer set of\/ $\mathcal{T}(\Pi)$,
  and
  \(
  \langle r_i\rangle_{i\in I}
  \)
  a grounded enumeration of the set \GR{\mathcal{T}(\Pi)}{X}
  of generating rules of $X$ from $\mathcal{T}(\Pi)$.

  Then, we have for all
  \(
  r,r'\in \Pi
  \)\/:
  \[
  \text{If }
  \PRECMi{X}{r}{r'},
  \text{ then }
  j<i
  \text{ for }
  r_i=\cokt{1}{r}
  \text{ and }
  r_j=\cokt{1}{r'}
  \ \mbox{.}
  \]
\end{theorem}
%
This result obviously extends to the respective images of rules $r$ and $r'$.
%
\begin{corollary}\label{cor:order:implementing}
  Given the same prerequisites as in Theorem~\ref{thm:order:implementing},
  we have for all
  \(
  r,r'\in \Pi
  \):
  
  \[
  \text{If } \PRECMi{X}{r}{r'}, \text{ then } j<i
  \]
  for all $r_i$
  equaling 
  \ap{2}{r}
  or
  \bl{k}{r}{L}
  for some $k=1,2$
  and some $L\in\body{r}$, and
  some
  $r_j$
  equaling 
  \ap{2}{r'}
  or
  \bl{k'}{r'}{L'}
  for some
  $k'=1,2$
  and some $L'\in\body{r'}$.
\end{corollary}
%
The last results reflect nicely how the prescriptive strategy is enforced.
Whenever a rule $r'$ is preferred to another rule $r$, one of its images, say
\bl{2}{r'}{\neg d}, occurs necessarily before all images of the less preferred
rule, \egc \ap{2}{r}.
(Note that the existential quantification of $r_j$ is necessary since $r'$
could be blocked in several ways.)

For static preferences, our translation $\mathcal{T}$ amounts to selecting
$\PRECMo$-preserving answer sets of the underlying (unordered) program,
as given by Definition~\ref{def:order:preserving}.
%
\begin{theorem}\label{thm:order:preserving}
  Let $(\Pi,\PRECMo)$ be a statically ordered logic program
  and
  let $X$ be a consistent set of literals.

  Then,
  $X$ is a $\PRECMo$-preserving answer set of\/ $\Pi$
  iff
  $X=Y\cap\mathcal{L_\mathcal{A}}$
  for some answer set $Y$ of\/ $\mathcal{T}(\Pi,\PRECMo)$.
\end{theorem}
%
This result provides semantics for statically ordered programs;
it provides an exact correspondence between answer sets issued by our
translation and regular answer sets of the original program.

Note that inconsistent answer sets are not necessarily reproduced by
$\mathcal{T}$.
To see this,
consider
\(
\Pi=\{p\LPif,\neg p\LPif\}
\).
While $\Pi$ has an inconsistent answer set,
$\mathcal{T}(\Pi)$ has no answer set.
This is due to the fact that rules like \cok{2}{r}{r'} are removed from
$\reduct{\mathcal{T}(\Pi)}{\mathcal{L}}$.
In such a case, there is then no way to derive ``$\oksym$''-literals via
\cokt{1}{r}.
On the other hand, one can show that $\mathcal{T}(\Pi,\PRECMo)$ has 
an inconsistent answer set only if $\Pi$ has an
inconsistent answer set,
for any statically ordered program $(\Pi,\PRECMo)$.%
\footnote{Observe that the static translation $\mathcal{T}'(\Pi,\PRECMo)$
preserves inconsistent answer sets of $\Pi$, however.}

We obtain the following corollary to Theorem~\ref{thm:order:preserving},
demonstrating that our strategy implements a selection function among the
standard answer sets of the underlying program.
%
\begin{corollary}\label{thm:standard}
  Let $(\Pi,\PRECMo)$ be a statically ordered logic program
  and
  $X$ a set of literals.

  If
  \(
  X=Y\cap\mathcal{L_\mathcal{A}}
  \)
  for some answer set $Y$ of\/ $\mathcal{T}(\Pi,\PRECMo)$,
  then
  $X$ is an answer set of\/ $\Pi$.
\end{corollary}
%
Note that the last two results do not directly carry over to the general
(dynamic) case,
since a dynamic setting admits no way of selecting answer sets in an
undifferentiated way due to the lack of a uniform preference relation.

The key difference between static and dynamic preference handling boils down
to the availability of preferences.
While the full preference information is available right from the start in the
static case,
it develops with the formation of answer sets in the dynamic case.
Moreover,
from the viewpoint of rule application,
a dynamic setting not only necessitates that the question of applicability has
been settled for all higher ranked rules before a rule is considered for
application but also that one knows about all higher ranked rules at this
point.
That is, before considering a rule $r$ for application, all preferences of
form \PREC{n_r}{n_{r'}} must have been derived.
This adds a new requirement to the concept of order preservation in the
dynamic case, expressed by Condition~\ref{def:order:preserving:x:tri} below.
A major consequence of this additional requirement is that we cannot restrict
our attention to the generating rules of an answer set anymore, as in the
static case, but that we have to take all rules of the program into account,
no matter whether they eventually apply or not.

Preference information is now drawn from the considered answer sets.
We only have to make sure that the resulting preferences amount to a strict
partial order.
For this purpose,
let \TA{\Pi} denote the program obtained from $\Pi$ by adding transitivity
and antisymmetry rules, that is
\(
\TA{\Pi}
=
\Pi\cup\{{t(r,r',r'')},{as(r,r')}\mid r,r',r'' \in \Pi\}
\).

Taking all this into account, we arrive at the following concept of order
preservation for dynamic preferences.
%
\begin{definition}\label{def:order:preserving:x}
  Let $\Pi$ be an ordered program
  and
  let $X$ be a consistent answer set of \TA{\Pi}.

  Then, $X$ is called  \emph{\PRECMoi{X}-preserving},
  if there exists an enumeration
  \(
  \langle r_i\rangle_{i\in I}
  \)
  of\/ \TA{\Pi}
  such that, for every $i,j\in I$, we have that:
  \begin{enumerate}
  \item\label{def:order:preserving:x:one}
    if \PRECMi{X}{r_i}{r_j}, then $j<i$;
  \item\label{def:order:preserving:x:tri}
    if
    \PRECMi{X}{r_i}{r_j},
    then
    there is some $r_k\in\GR{\TA{\Pi}}{X}$
    such that
    \begin{enumerate}
    \item\label{def:order:preserving:x:tri:a}
      $k<i$
      \quad  and
    \item\label{def:order:preserving:x:tri:b}
      \(
      \head{r_k}= (\PREC{\name{r_i}}{\name{r_j}})
      \);
    \end{enumerate}
  \item\label{def:order:preserving:x:zero}
    if
    \(
    r_i\in\GR{\TA{\Pi}}{X}
    \),
    then
    \(
    \pbody{r_i}\subseteq\{\head{r_k}\mid r_k\in\GR{\TA{\Pi}}{X}, k<i\}
    \);
    \quad and
  \item\label{def:order:preserving:x:two}
    if
    \(
    r_i\in{\TA{\Pi}\setminus\GR{\TA{\Pi}}{X}},
    \)
    then
    \begin{enumerate}
    \item\label{def:order:preserving:x:two:a}
      \(
      \pbody{r_i}\not\subseteq X
      \)
      \quad or
    \item\label{def:order:preserving:x:two:b}
      \(
      \nbody{r_i}\cap\{\head{r_k}\mid r_k\in\GR{\TA{\Pi}}{X}, k<i\}\neq\emptyset
      \).
    \end{enumerate}
  \end{enumerate}
\end{definition}
%
Despite the fact that this conception of order preservation takes all rules
into account, as opposed to generating rules only,
it remains a generalisation of its static counterpart given in
Definition~\ref{def:order:preserving}.
In fact, above 
Conditions~\ref{def:order:preserving:x:zero},
           \ref{def:order:preserving:x:one}, and
           \ref{def:order:preserving:x:two}
can be seen as extension of 
Conditions~\ref{def:order:preserving:zero},
           \ref{def:order:preserving:one}, and
           \ref{def:order:preserving:two} in
Definition~\ref{def:order:preserving} from sets of generating rules to entire
programs.

For illustration, consider answer set
\(
X=\{\neg a,b,\PREC{n_3}{n_2},\neg(\PREC{n_3}{n_2})\}
\)
of $\TAi{\Pi}{eq:example}$;
this answer set is \PRECMoi{X}-preserving.
To see this, consider the following enumeration of ${\Pi_{\ref{eq:example}}}$
(omitting rules in
$\TAi{\Pi}{eq:example}\setminus{\Pi_{\ref{eq:example}}}$ for
simplicity;
we write $\overline{r}$ to indicate that
$r\not\in\GR{\TAi{\Pi}{eq:example}}{X}$):
\begin{equation}
  \label{eq:example:enumeration}
  \langle
  {r_1},
  {r_4},
  {r_2},
  \overline{r}_3
  \rangle \ \mbox{.}
\end{equation}
The only preference \PRECMi{X}{\overline{r}_3}{{r_2}} induces, 
through Condition~\ref{def:order:preserving:x:one}, that
${r_2}$ occurs before $\overline{r}_3$
and that
${r_4}$ occurs before $\overline{r}_3$
via Condition~\ref{def:order:preserving:x:tri}.
Furthermore, due to Condition~\ref{def:order:preserving:x:zero},
${r_1}$ occurs before ${r_2}$,
while Condition~\ref{def:order:preserving:x:two:b} necessitates (also) that
${r_2}$ occurs before $\overline{r}_3$.
Intuitively, the enumeration in~(\ref{eq:example:enumeration}) corresponds to
that of the images of $\Pi_{\ref{eq:example}}$ generating the standard answer
set of $\mathcal{T}(\TAi{\Pi}{eq:example})$, viz.\
\[
\langle
\dots,
\ap{2}{r_1},
\dots,
\ap{2}{r_4},
\dots,
\ap{2}{r_2},
\dots,
\bl{2}{r_3}{b},
\dots
\rangle
\ \mbox{.}
\]

As discussed above, the difference to the static case manifests itself
in Condition~\ref{def:order:preserving:x:tri}, which reflects the fact that
all relevant preference information must be derived before a rule is considered for
application.
This also mirrors the strong commitment of the strategy to a successive
development of preferred answer sets in accord with groundedness.
This is different from the approach discussed in
Section~\ref{sec:brewka:eiter} that drops the groundedness requirement.

For illustrating Condition~\ref{def:order:preserving:x:tri},
consider programs
\(
\Pi_{\ref{eq:dynamic:two}a}=\{r_1,r_2,r_{3a}\}
\)
and
\(
\Pi_{\ref{eq:dynamic:two}b}=\{r_1,r_2,r_{3b}\}
\),
where
\begin{equation}
  \label{eq:dynamic:two}
  \begin{array}{lcrcl}
    r_1 & = & a & \LPif & \naf\neg a
    \\
    r_2 & = & b & \LPif & \naf\neg b
    \\ 
    r_{3a} & = & \PREC{\name{1}}{\name{2}} & \LPif& a
  \end{array}
  \qquad
  \begin{array}{lcrcl}
    r_1 & = & a & \LPif & \naf\neg a
    \\
    r_2 & = & b & \LPif & \naf\neg b
    \\ 
    r_{3b} & = & \PREC{\name{1}}{\name{2}} & \LPif& b \ \mbox{.}
  \end{array}  
\end{equation}
Both programs have the same standard answer set
$X_{\ref{eq:dynamic:two}}=\{a,b,\PREC{\name{1}}{\name{2}}\}$.
However, this answer set is only \PRECMoi{X_{\ref{eq:dynamic:two}}}-preserving
for $\Pi_{\ref{eq:dynamic:two}b}$ but not for $\Pi_{\ref{eq:dynamic:two}a}$.
This is because $\Pi_{\ref{eq:dynamic:two}b}$ allows to derive the preference
atom \PREC{\name{1}}{\name{2}} before the lower ranked rule, $r_1$, is
considered for application,
while this is impossible with $\Pi_{\ref{eq:dynamic:two}a}$.
That is, while there is an enumeration of $\Pi_{\ref{eq:dynamic:two}b}$
satisfying both Condition~\ref{def:order:preserving:x:tri}
and~\ref{def:order:preserving:x:zero} there is no such enumeration of
$\Pi_{\ref{eq:dynamic:two}a}$.
A dynamic extension of the approach dropping
Condition~\ref{def:order:preserving:x:tri} is described in
Section~\ref{sec:be:dynamic}.

We have the following result demonstrating soundness and completeness of
translation $\mathcal{T}$ with respect to \PRECMoi{X}-preserving answer sets.
%
\begin{theorem}\label{thm:order:preserving:x}
  Let $\Pi$ be an ordered logic program
  and
  let $X$ be a consistent set of literals.

  Then,
  $X$ is a \PRECMoi{X}-preserving answer set of\/ \TA{\Pi}
  iff\/
  $X=Y\cap\mathcal{L}$ for some answer set $Y$ of\/ $\mathcal{T}(\Pi)$.
\end{theorem}
%
As above, we obtain the following corollary.
%
\begin{corollary}\label{thm:dynamic:standard}
  Let $\Pi$ be an ordered logic program
  and
  let $X$ be a set of literals.

  If
  $X=Y\cap\mathcal{L}$ for some answer set $Y$ of\/ $\mathcal{T}(\Pi)$,
  then
  $X$ is an answer set of\/ \TA{\Pi}.
\end{corollary}

Also, if no preference information is present, 
transformation $\mathcal{T}$ amounts to standard answer set semantics.
Moreover, the notions of statically ordered and (dynamically) ordered programs
coincide in this case.
The following result expresses this property in terms of order preserving
answer sets.
%
\begin{theorem}\label{thm:conservative}
  Let $\Pi$ be a logic program over $\mathcal{L}_\mathcal{A}$
  and
  let $X$ be a consistent set of literals.

  Then,
  the following statements are equivalent:
  \begin{enumerate}
  \item\label{thm:conservative:two}
    $X$ is a \PRECMoi{X}-preserving answer set of the dynamically ordered 
    program\/ $\Pi$ and ${\PRECMoi{X}}=\emptyset$.
  \item\label{thm:conservative:one}
    $X$ is a \PRECMo-preserving answer set of the statically ordered program 
    $(\Pi,\PRECMo)$ and ${\PRECMo}=\emptyset$.
  \item\label{thm:conservative:three}
    $X$ is a regular answer set of logic program $\Pi$.
\end{enumerate}
\end{theorem}

From these properties, we immediately obtain the following result for 
our encoding $\mathcal{T}$, showing that preference-free programs yield
  regular answer sets.
%
\begin{corollary}\label{thm:conservative:transformation}
  Let\/ $\Pi$ be a logic program over $\mathcal{L}_\mathcal{A}$
  and
  let $X$ be a consistent set of literals.
  
  Then, $X$ is an answer set of\/ $\Pi$ iff $X=Y\cap\mathcal{L}$
  for some answer set $Y$ of\/ $\mathcal{T}(\Pi)$.
\end{corollary}

Brewka~\& Eiter~\shortcite{breeit99a} have suggested two properties,
termed \emph{Principle~I} and \emph{Principle~II}, which
they argue any defeasible rule system handling preferences should satisfy.
The next result shows that order preserving answer sets%
\footnote{In the sense of Definitions~\ref{def:order:preserving} and
  \ref{def:order:preserving:x}.}
enjoy these properties. 
However, since the original formulation of Principle~I and II is
rather generic---motivated by the aim to cover as many different 
approaches as possible---we must instantiate them 
in terms of the current framework.
It turns out that Principle~I is only suitable for statically ordered
programs, whilst Principle~II admits two guises, one for statically
ordered programs, and another one for (dynamically) ordered programs.

Principles I and II, formulated for our setting, are as follows:

\begin{description}
  \item[Principle I.] 
    Let $(\Pi,\PRECMo)$ be a statically ordered logic program
    and
    let $X_1$ and $X_2$ be two (regular) answer sets of $\Pi$
    generated by 
    $\GR{\Pi}{X_1}=R\cup\{r_1\}$
    and
    $\GR{\Pi}{X_2}=R\cup\{r_2\}$,
    respectively,
    where $r_1,r_2\not\in R$.

    If $\PRECM{r_1}{r_2}$,
    then $X_1$ is not a $\PRECMo$-preserving answer set of $\Pi$.

  \item[Principle II-S (Static Version).] 
    Let $(\Pi,\PRECMo)$ be a statically ordered logic program
    and
    let $X$ be a $\PRECMo$-preserving answer set of $\Pi$.
    Let $r$ be a rule where $\pbody{r}\not\subseteq X$
    and
    let $(\Pi\cup\{r\},\PRECMo')$ be a statically ordered logic program
    where 
    $< \; = \; {<'\cap{(\Pi\times\Pi)}}$.

    Then, 
    $X$ 
    is a $\PRECMo'$-preserving answer set of
    $\Pi\cup\{r\}$.

  \item[Principle II-D (Dynamic Version).]
    Let $\Pi$ be a (dynamically) ordered logic program
    and
    let $X$ be a \PRECMoi{X}-preserving answer set of $\TA{\Pi}$.
    Let $r$ be a rule such that $\pbody{r}\not\subseteq X$.

    Then, 
    $X$ 
    is a \PRECMoi{X}-preserving answer set of $\TA{\Pi}\cup\{r\}$.

\end{description}
%
\begin{theorem}\label{thm:principles}
  We have the following properties.
  \begin{enumerate}
  \item Order preserving answer sets in the sense of
    Definition~\ref{def:order:preserving} satisfy Principles I and II-S.
  \item Order preserving answer sets in the sense of
    Definition~\ref{def:order:preserving:x} satisfy Principle II-D.
  \end{enumerate}
\end{theorem}

Finally, we mention some properties concerning the computational complexity of
order preserving answer sets. 
Since transformation $\mathcal{T}$ is clearly polynomial in the size of
ordered logic programs, in virtue of Theorems~%
\ref{thm:order:preserving},
\ref{thm:order:preserving:x}, and
\ref{thm:conservative:transformation},
it follows in a straightforward way that the complexity of answer set
semantics under order preservation coincides with the complexity of
standard answer set semantics.
We note the following results.
%
\begin{theorem} Order preserving answer sets enjoy the following properties:
\begin{enumerate}
\item  Given an ordered program $\Pi$,
  checking whether $\Pi$ has an answer set $X$ which is \PRECMoi{X}-preserving
  is NP-complete.
\item  Given an ordered program $\Pi$ and some literal $L$,
  checking whether $\Pi$ has an answer set $X$ which is \PRECMoi{X}-preserving
  and which contains $L$ is NP-complete.
\item  Given an ordered program $\Pi$ and some literal $L$,
  checking whether $L$ is contained in any answer set $X$ which is
  \PRECMoi{X}-preserving is  coNP-complete. 
\end{enumerate}
\end{theorem}

\subsection{Variations}
\label{sec:order:variations}

A strategy similar to the one described above has been advocated
in Wang, Zhou,~\& Lin~\shortcite{wazhli00}.
There, fixed-point definitions are used to characterise preferred answer sets
(of statically ordered programs).
It turns out that this preferred answer set semantics can be
implemented by a slight modification of the translation given in
Definition~\ref{def:compilation}.
\pagebreak
\removebrackets
\begin{definition}[\nbcite{schwan01b}]
Given the same prerequisites as in Definition~\ref{def:compilation},
the logic program $\mathcal{W}(\Pi)$ over $\mathcal{L}^{+}$ is defined as
\[
\mathcal{W}(\Pi)
=
\mbox{$\bigcup_{r\in\Pi}$}\tau(r)\cup\{\cok{5}{r}{r'}\mid r,r'\in\Pi\},
\]
where
\[
\begin{array}{rrcl}
  \RULE{\cok{5}{r}{r'}}
       {\rdy{\name{r}}{\name{r'}}}
       {(\PREC{\name{r}}{\name{r'}}),\head{r'}} \ \mbox{.}
\end{array}
\]
\end{definition}
%
The purpose of \cok{5}{r}{r'} is to eliminate rules from the preference
handling process once their head has been derived.
A corresponding soundness and completeness result can be found in 
Schaub~\& Wang~\shortcite{schwan01b}.

In terms of order preservation, this amounts to weakening the integration of
preferences and groundedness:
%
\removebrackets
\begin{definition}[\nbcite{schwan01b}]\label{def:weak:order:preserving}
  Let $(\Pi,<)$ be a statically ordered program
  and
  let $X$ be a consistent answer set of~$\Pi$.
  
  Then, $X$ is called  \emph{$<^{\WZL}$-preserving},
  if there exists an enumeration
  \(
  \langle r_i\rangle_{i\in I}
  \)
  of\/ $\GR{\Pi}{X}$
  such that for every $i,j\in I$ we have that:
  \begin{enumerate}
  \item 
    \begin{enumerate}
    \item $\pbody{r_i}\subseteq\{\head{r_j}\mid j<i\}$
      \quad or
    \item $\head{r_i} \in \{\head{r_j}\mid j<i\}$;
    \end{enumerate}
  \item if $r_i<r_j$, then $j<i$;
    \quad and
  \item if
    $\PRECM{r_i}{r'}$
    and 
    \(
    r'\in {\Pi\setminus\GR{\Pi}{X}},
    \)
    then
    \begin{enumerate}
    \item 
      $\pbody{r'}\not\subseteq X$
      \quad or
    \item 
      $\nbody{r'}\cap\{\head{r_j}\mid j<i\}\neq\emptyset$
      \quad or
    \item
      $\head{r'}\in \{\head{r_j}\mid j<i\}$.
    \end{enumerate}
  \end{enumerate}
\end{definition}
%
The primary difference of this concept of order preservation to the original
one is clearly the weaker notion of groundedness.
This involves the rules in \GR{\Pi}{X} (via Condition~1b) as well as those
in ${\Pi\setminus\GR{\Pi}{X}}$ (via Condition~3c).
The rest of the definition is the same as in
Definition~\ref{def:order:preserving}.
We refer to Schaub~\& Wang~\shortcite{schwan01b} for formal details.

Informally, the difference between both strategies can be explained by means
of the following program
\(
\Pi_{\ref{eq:wang}}=\{r_1,r_2,r_3\}
\),
where
\begin{equation}
  \label{eq:wang}
  \begin{array}[t]{lcrcl}
    r_1 & = & a & \LPif &     \naf\neg a
    \\
    r_2 & = & b & \LPif & a , \naf\neg b
    \\ 
    r_3 & = & b & \LPif &
  \end{array}
  \qquad\text{ and }\qquad  
  \PRECM{r_1}{r_2}
  \ \mbox{.}
\end{equation}
Clearly, $\Pi_{\ref{eq:wang}}$ has a single standard answer set,
\(
\{a,b\}
\).
This answer set cannot be generated by an enumeration of $\Pi_{\ref{eq:wang}}$
that preserves \PRECM{r_1}{r_2} in the sense of
Definition~\ref{def:order:preserving}.
This is different after adding Condition~1b.
In accord with the approach in Wang, Zhou,~\& Lin~\shortcite{wazhli00},
the modified concept of order preservation accepts the enumeration
\(
\langle r_3,r_2,r_1\rangle
\).

All of the above strategies enforce a selection among the standard answer
sets of the underlying program.
That is, no new answer sets appear; neither are existing ones modified.
While the former seems uncontroversial,
the latter should not be taken for granted,
as exemplified by a ``winner-takes-all'' strategy.
In such an approach, one wants to apply the highest ranked default, if
possible, and only the highest ranked default.
This can be accomplished by removing rule
\begin{equation}
  \label{eq:inheritance}
  \begin{array}{rrcl}
  \RULE{\cok{3}{r}{r'}}
       {\rdy{\name{r}}{\name{r'}}}
       {(\PREC{\name{r}}{\name{r'}}),\applied{\name{r'}}}
\end{array}
\end{equation}
from Definition~\ref{def:compilation}.
Then, once a higher ranked rule, such as $r'$, is applied,
none of the lower ranked rules, like $r$, are considered for application
anymore
because there is no way to derive the indispensable ``ok-literal'',
\ok{\name{r}}.
A similar strategy is developed in Gelfond~\& Son~\shortcite{gelson97a}
(cf.\ Section~\ref{sec:related:work}).

A particular instance of a ``winner-takes-all'' strategy is inheritance of
default properties.
In this case,
the ordering on rules reflects a relation of \emph{specificity} among
the (default) rules prerequisites.
Informally, for adjudicating among conflicting defaults, one determines
the most specific (with respect to rule antecedents) defaults as candidates
for application.
Consider for example defaults concerning primary means of locomotion:
``animals normally walk'',
``birds normally fly'',
``penguins normally swim''.
This can be expressed using (default) rules as follows:
\begin{equation}
  \label{eq:animals}
  \begin{array}[t]{lcrcl}
    r_1 & = & w & \LPif & a ,\naf\neg w
    \\
    r_2 & = & f & \LPif & b ,\naf\neg f
    \\ 
    r_3 & = & s & \LPif & p ,\naf\neg s
  \end{array}
  \qquad\qquad
  \begin{array}[t]{lcrcl}
    r_a & = & a & \LPif & b
    \\
    r_b & = & b & \LPif & p
    \\
    \multicolumn{5}{l}{r_1\PRECMo r_2\PRECMo r_3 \ \mbox{.}}
  \end{array}  
\end{equation}
If we learn that some object is a penguin, viz.\ $p\LPif$, (and so a bird and
animal via $r_a$ and $r_b$), then we would want to apply the highest ranked
default, if possible, and only the highest ranked default.
Significantly, if the penguins-swim default $r_3$ is blocked (say, the penguin in
question has a fear of water, viz.\ $\neg s\LPif$) we \emph{don't} try to apply
the next default to see if it might fly.

So, given a chain of rules expressing default properties
\(
r_1\PRECMo r_2\PRECMo \dots\PRECMo r_m,
\)
the elimination of (\ref{eq:inheritance}) in Definition~\ref{def:compilation}
portrays the following comportment:
We apply $r_i$, if possible, where $r_i$
is the $<$-maximum default such that for every default
$r_j$, $j = i+1,\dots ,m$, the prerequisite of 
$r_j$ is not known to be true.
Otherwise, no default in the chain is applicable.
As mentioned above, the elimination of (\ref{eq:inheritance}) allows lower
ranked default rules to be applied only in case higher ranked rules are
blocked because their prerequisite is not derivable.
Otherwise, the propagation of \ok{\cdot}-atoms is interrupted so that no
defaults below $r_i$ are considered.

\section{Brewka and Eiter's Approach to Preference}
\label{sec:brewka:eiter}

We now turn to a strategy for preference handling proposed
in Brewka~\& Eiter~\shortcite{breeit97a,breeit99a}.
Unlike above, this approach is not fully prescriptive in the sense that
it does not enforce the ordering information during the construction of an
answer set.
Rather it relies on the existence of a regular answer set of the underlying
non-ordered program,
whose order preservation is then verified in a separate test
(see Definition~\ref{def:be:extension:general}).

\subsection{Original Definition}
\label{sec:be}

To begin with, we describe the approach to dealing with ordered logic programs,
as introduced in Brewka~\& Eiter~\shortcite{breeit99a}.
First, Brewka and Eiter deal with statically ordered logic programs only.
Also, partially ordered programs are reduced to totally ordered ones:%
\footnote{While Brewka~\& Eiter~\shortcite{breeit99a} deal with potentially infinite
  well-orderings, we limit ourselves here to finite programs.}
A \emph{fully ordered logic program} is a statically ordered logic program
$(\Pi,\ll)$ where $\ll$ is a total ordering.  
The case of arbitrarily ordered programs is reduced to this restricted case in
the following way.
%
\begin{definition}\label{def:be:total:order}
Let $(\Pi,\PRECMo)$ be a statically ordered logic program 
and let
$X$ be a set of literals. 

Then,
$X$ is a \emph{\BEpreferred{} answer set} of $(\Pi,\PRECMo)$ 
iff
$X$ is a \BEpreferred{} answer set of some fully ordered logic program
$(\Pi,\ll)$ such that ${\PRECMo}\subseteq{\ll}$.
\end{definition}

The construction of \BEpreferred{} answer sets relies on an operator,
defined for prerequisite-free programs.
%
\begin{definition}\label{def:be:operator}
  Let $(\Pi,\ll)$ be a fully ordered prerequisite-free logic program,
  let
  \(
  \langle r_i\rangle_{i\in I}
  \)
  be the enumeration of $\Pi$ according to the ordering $\ll$, and let $X$ be a set
  of literals.

  Then,\footnote{Note that $C$ is implicitly parameterised with $(\Pi,\ll)$.}
  $C(X)$ is the smallest logically closed set of literals containing
  \(
  \bigcup_{i\in I} X_i
  \),
  where $X_j=\emptyset$ for $j\not\in I$ and
  \[
  X_i = 
  \left\{
    \begin{array}{ll}
      X_{i-1}
      &
      \text{ if }
      \nbody{r_i}\cap X_{i-1}\neq\emptyset;
      \\[1ex]
      X_{i-1}
      &
      \text{ if }
      \head{r_i}\in X
      \text{ and }
      \nbody{r_i}\cap X\neq\emptyset;
      \\[1ex]
      X_{i-1}\cup\{\head{r_i}\}
      &
      \text{ otherwise.}
  \end{array}
  \right.
  \]
\end{definition}
%
This construction is unique insofar that for any fully ordered
prerequisite-free program $(\Pi,\nolinebreak{\ll)}$,
there is at most one (standard) answer set $X$ of $\Pi$ such that $C(X)=X$
(cf.\ Lemma~4.1 in Brewka~\& Eiter~\shortcite{breeit99a}).
Accordingly, this set is used to define a \BEpreferred{} answer set
for \emph{prerequisite-free logic programs}, if it exists:
%
\begin{definition}\label{def:be:extension:supernormal}
  Let $(\Pi,\ll)$ be a fully ordered prerequisite-free logic program
  and let $X$ be a set of literals.

  Then,
  $X$ is the \BEpreferred{} answer set of $(\Pi,\ll)$
  iff\/
  \(
  C(X)=X
  \).
\end{definition}

The second condition in Definition~\ref{def:be:operator} amounts to
eliminating from the above construction all rules whose heads are in $X$ but
which are defeated by $X$.
This can be illustrated by the following logic program,
adapted from Brewka~\& Eiter~\shortcite{breeit98a}:
\begin{equation}\label{eq:three}
  \begin{array}[t]{rcrcl}
    r_1 & = & a      &\LPif& \naf b
    \\
    r_2 & = & \neg a &\LPif& \naf a
    \\
    r_3 & = & a      &\LPif& \naf\neg a
    \\
    r_4 & = & b      &\LPif& \naf\neg b
  \end{array}
  \qquad\text{ with }\qquad
  \{\PRECM{r_j}{r_i}\mid i<j\}
  \ \mbox{.}
\end{equation}
Program $\Pi_{\ref{eq:three}}=\{r_1,\dots,r_4\}$ has two answer sets,
$\{a,b\}$ and $\{\neg a,b\}$.
The application of operator $C$ relies on sequence
\(
\langle r_1,r_2,r_3,r_4 \rangle
\).
For illustration, consider the process induced by
\(
C(\{a,b\})
\),
with and without the second condition:
\[
\begin{array}{llll}
 X_1=\{ \} & X_2=\{\neg a\} & X_3=\{\neg a\} & X_4=\{\neg a,b\} 
\\
X'_1=\{a\} &X'_2=\{     a\} &X'_3=\{     a\} &X'_4=\{     a,b\} \ \mbox{.}
\end{array}
\]
Thus, without the second condition, $\{a,b\}$ would be a preferred answer set.
However, Brewka~\& Eiter~\shortcite{breeit99a,breeit98a} argue that 
such an answer set
does not preserve priorities because $r_2$ is defeated in $\{a,b\}$ by applying a
rule which is less preferred than $r_2$, namely $r_3$.
The above program has therefore no \BEpreferred{} answer set.

The next definition accounts for the general case by reducing it to the
prerequisite-free one.
For checking whether a given regular answer set $X$ is \BEpreferred, 
Brewka and Eiter evaluate the prerequisites of the rules with respect to
the answer set $X$.
%
\begin{definition}\label{def:be:answer:set}
  Let $(\Pi,\ll)$ be a fully ordered logic program
  and
  $X$ a set of literals.

  The logic program $(\Pi_X,\ll_X)$ is obtained from $(\Pi,\ll)$ as follows:
  \begin{enumerate}
  \item\label{l:one:b}
    $\Pi_X=\{\reductBE{r}\mid r\in\Pi \text{ and } \pbody{r}\subseteq X\}$;
    \quad and
  \item\label{l:two}
  for any 
  $r'_1, r'_2 \in \Pi_X$, 
  $r'_1 \ll_X r'_2$
  iff $r_1 \ll r_2$ where
  \(
  r_i=\max_\ll\{r\in \Pi\mid\reductBE{r}=r'_i\}
  \).
\end{enumerate}
\end{definition}
%
In other words, $\Pi_X$ is obtained from $\Pi$ by
first eliminating every rule $r\in\Pi$ such that $\pbody{r}\not\subseteq X$,
and
then substituting all remaining rules $r$ by 
$\reductBE{r}=\head{r}\LPif\nbody{r}$.
This results in a prerequisite-free logic program.

\BEpreferred{} answer sets are then defined as follows.
%
\begin{definition}\label{def:be:extension:general}
  Let $(\Pi,\ll)$ be a fully ordered logic program
  and
  $X$ a set of literals.

  Then, $X$ is a \BEpreferred{} answer set of $(\Pi,\ll)$, 
  if
  \begin{enumerate}
  \item $X$ is a (standard) answer set of\/ $\Pi$, and
  \item $X$ is a \BEpreferred{} answer set of $(\Pi_X,\ll_X)$.
  \end{enumerate}
\end{definition}

For illustration, consider Example 5.1 of Brewka~\& Eiter~\shortcite{breeit99a}:
\begin{equation}\label{eq:five:one}
  \begin{array}[t]{rcrcl}
    r_1 & = &      b & \LPif & a, \naf\neg b
    \\
    r_2 & = & \neg b & \LPif & \naf     b
    \\ 
    r_3 & = &      a & \LPif & \naf\neg a
  \end{array}
  \qquad\text{ with }\qquad
  \{\PRECM{r_j}{r_i}\mid i<j\}
  \ \mbox{.}
\end{equation} 
Program $\Pi_{\ref{eq:five:one}}=\{r_1,r_2,r_3\}$ has two standard answer sets: 
\(
X_1=\{a,b\}
\)
and
\(
X_2=\{a,\neg b\}
\).
$(\Pi_{\ref{eq:five:one}})_{X_1}$ turns $r_1$ into
\(
b \LPif \naf\neg b
\)
while leaving $r_2$ and $r_3$ unaffected.
Also, we obtain that
\(
C(X_1) = X_1
\), 
that is, $X_1$ is a \BEpreferred{} answer set.
In contrast to this, $X_2$ is not \BEpreferred.
While $(\Pi_{\ref{eq:five:one}})_{X_2}=(\Pi_{\ref{eq:five:one}})_{X_1}$,
we get
\(
C(X_2)=X_1\neq X_2
\).
That is, $C(X_2)$ reproduces $X_1$ rather than~$X_2$.

\subsection{Order Preservation}
\label{sec:be:order}

Before giving an encoding for the approach of Brewka and Eiter, we would
like to provide some insight into its semantics and its
relation to the previous strategy in terms of the notion of order
preservation.
First, the approach is equivalent to the strategy developed in
Section~\ref{sec:order:preservation} on normal prerequisite-free default
theories \cite{delsch00a}.
Interestingly, this result does not extend to either normal or to
prerequisite-free theories.
While the former difference is caused by the second condition in
Definition~\ref{def:be:operator},
the latter is due to the different attitude towards groundedness 
(see below).
Despite these differences it turns out that every order preserving extension
in the sense of Definition~\ref{def:order:preserving} is also obtained in
Brewka and Eiter's approach but not vice versa
(cf.\ \cite{delsch00a}).

To see the latter, consider the example in~(\ref{eq:baader:hollunder}).
Among the two answer sets $X_1=\{p,b,w,\neg f\}$ and $X_2=\{p,b,w,f\}$ of
$\Pi_{\ref{eq:baader:hollunder}}$, only $X_1$ is order preserving in the
sense of Definition~\ref{def:order:preserving}.
In contrast to this, both answer sets are \BEpreferred.
This is because groundedness is not at issue when verifying order
preservation in the approach of Brewka and Eiter due to the removal of
prerequisites in Definition~\ref{def:be:answer:set}.
This can be made precise by means of a corresponding notion of order
preservation, taken from Schaub~\& Wang~\shortcite{schwan01b}:\footnote{%
  An alternative characterisation of \BEpreferred{} answer sets for fully
  ordered logic programs, originally given in Proposition~5.1 in
  Brewka~\& Eiter~\shortcite{breeit99a}, is repeated in
  Theorem~\ref{thm:be:characterisation}.}
%
\removebrackets
\begin{definition}
\label{def:be:order:preserving}
  Let $(\Pi,\PRECMo)$ be a statically ordered program
  and
  let $X$ be a consistent answer set of~$\Pi$.

  Then, $X$ is called  \emph{$\PRECMo^{\BE}$-preserving},
  if there exists an enumeration
  \(
  \langle r_i\rangle_{i\in I}
  \)
  of\/ $\GR{\Pi}{X}$
  such that, for every $i,j\in I$, we have that:
  \begin{enumerate}
    \item if $r_i<r_j$, then $j<i$;
          \quad and
    \item if
          $\PRECM{r_i}{r'}$
          and 
          \(
          r'\in {\Pi\setminus\GR{\Pi}{X}},
          \)
          then
          \begin{enumerate}
          \item 
            $\pbody{r'}\not\subseteq X$
          \quad or
          \item 
            $\nbody{r'}\cap\{\head{r_j}\mid j<i\}\neq\emptyset$
          \quad or
          \item
            $\head{r'}\in X$.
          \end{enumerate}
        \end{enumerate}
\end{definition}
%
This criterion differs from the one in Definition~\ref{def:order:preserving}
in the two (aforementioned) respects:
First, it drops the requirement of a \emph{grounded} enumeration
and,
second, it adds the second condition from Definition~\ref{def:be:operator}.
Note that any $\PRECMo^{\BE}$-preserving answer set is still generated by some
grounded enumeration of rules;
it is just that this property is now separated from the preference handling process.

Although the only grounded sequence given in~(\ref{eq:sequences:two}) is still
not $\PRECMo^{\BE}$-preserving,
there are now other $\PRECMo^{\BE}$-preserving enumerations supporting the
second answer set $X_2$,
\eg
\(
\langle
r_5,r_4,r_3,r_2
\rangle
\).
None of these enumerations enjoys groundedness.
Rather, all of them share the appearance of $r_3$ before $r_2$ although
the application of $r_3$ relies on that of $r_2$.
The reversal of the groundedness relation between $r_2$ and $r_3$ is however
essential for defeating $r_1$ (before $r_2$ is considered for application).
Otherwise, there is no way to satisfy Condition~2b.

The discussion of Example~(\ref{eq:baader:hollunder}) has already shed light
on the difference between the strictly prescriptive strategy discussed in the
previous section and Brewka and Eiter's more descriptive approach.
The more descriptive attitude of their approach is nicely illustrated by the
example in~(\ref{eq:five:one}).
Without dropping the requirement of deriving the prerequisite of the most
preferred rule $r_1$, there is no way of justifying an answer set containing
$b$.
Such an approach becomes reasonable if we abandon the strictly prescriptive
view of Section~\ref{sec:order:preservation} and adopt a more descriptive one
by decomposing the construction of a preferred answer set into a respective
guess and 
check step (as actually reflected by Definition~\ref{def:be:answer:set}).
That is, once we know that there is an answer set containing certain formulas,
we may rely on it when dealing with preferences.
The descriptive flavour of this approach stems from the availability of an
existing answer set of the underlying program when dealing with preference
information.
For instance, given the regular answer set $X_1=\{a,b\}$ we know that $a$ and
$b$ are derivable in a grounded fashion.
We may then rely on this information when taking preferences into account,
although
the order underlying the construction of the regular answer set may be
incompatible with the one imposed by the preference information.

Observe that program $\Pi_{\ref{eq:five:one}}$ has no $\PRECMo$-preserving
answer set.
Intuitively, this can be explained by the observation that for 
the highest ranked default $r_1$, neither applicability nor blockage can be
asserted:
Either of these properties relies on the applicability of less ranked
defaults, effectively resulting in a circular situation destroying any
possible $<$-preserving answer set.

\subsection{Encoding}
\label{sec:be:encoding}

Although  Brewka and Eiter's approach was originally defined only for
the static case, we provide in this section an encoding for the more
general dynamic case.

We have seen above that the underlying preference handling strategy drops
groundedness, an indispensable property of standard answer sets.
As a consequence, our encoding cannot keep the strictly prescriptive nature,
as the one given in Section~\ref{sec:order:preservation}.
Rather, we have to decouple the generation of standard answer sets from the
actual preference handling process.
Given a statically ordered program $(\Pi,\PRECMo)$,
the idea is thus to take the original program $\Pi$ and to extend it by a set
of rules $\mathcal{T}'(\Pi,\PRECMo)$ that ensures that the preferences in
\PRECMo{} are preserved (without necessitating groundedness).
In fact, $\mathcal{T}'(\Pi,\PRECMo)$ provides an image of $\Pi$, similar to that
of Section~\ref{sec:order:preservation}, except that the rules in
$\mathcal{T}'(\Pi,\PRECMo)$ are partially disconnected from those in $\Pi$ by
means of an extended language.
All this is made precise in the sequel.

Given a program $\Pi$ over language $\mathcal{L}$,
we assume a disjoint language $\mathcal{L'}$ containing literals $L'$ for each
$L$ in $\mathcal{L}$.
Likewise, rule $r'$ results from $r$ by replacing each literal $L$ 
in $r$ by $L'$. 
Then, in analogy to Section~\ref{sec:encoding},
we map ordered programs over some language $\mathcal{L}$ onto standard
programs in the language $\mathcal{L}^\circ$ obtained by extending 
$\mathcal{L}\cup\mathcal{L}'$ by new atoms
$(\PREC{\name{r}}{\name{s}})$,
\ok{\name{r}},
\rdy{\name{r}}{\name{s}},
\blocked{\name{r}},
and
\applied{\name{r}},
for each $r,s$ in $\Pi$.

\begin{definition}\label{def:be:compilation}
Let $\Pi = \{r_1,\dots, r_k\}$ be an ordered logic program over $\mathcal{L}$.

Then,
the logic program $\mathcal{U}(\Pi)$ over $\mathcal{L}^\circ$ is defined as
\[
\mathcal{U}(\Pi)
=
\Pi\cup\mbox{$\bigcup_{r\in\Pi}$}\tau(r) \ ,
\]
where the set $\tau(r)$ consists of the following rules,
for
$L\in\pbody{r}$,
$K\in\nbody{r}$,
$s,t \in\Pi$,
and
$J\in\nbody{s}$:
\[
\begin{array}{rrcl}
  \RULE{\ap{1}{r}}
       {\head{r'}}
       {\applied{\name{r}} }
  \\
  \RULE{\ap{2}{r}}
       {\applied{\name{r}}}
       {\ok{\name{r}},\body{r},\naf\nbody{r'}}
  \\
  \RULE{\bl{1}{r}{L}}
       {\blocked{\name{r}}}
       {\ok{\name{r}}, \naf L, \naf L'}
  \\
  \RULE{\bl{2}{r}{K}}
       {\blocked{\name{r}}}
       {\ok{\name{r}},K,K'}
  \\[2ex]
  \RULE{\cokt{1}{r}}
       {\ok{\name{r}}}
       {\rdy{\name{r}}{\name{r_1}},\dots,\rdy{\name{r}}{\name{r_k}}}
  \\
  \RULE{\cok{2}{r}{s}}
       {\rdy{\name{r}}{\name{s}}}
       {\naf(\PREC{\name{r}}{\name{s}})}
  \\
  \RULE{\cok{3}{r}{s}}
       {\rdy{\name{r}}{\name{s}}}
       {(\PREC{\name{r}}{\name{s}}),\applied{\name{s}}}
  \\
  \RULE{\cok{4}{r}{s}}
       {\rdy{\name{r}}{\name{s}}}
       {(\PREC{\name{r}}{\name{s}}),\blocked{\name{s}}}
  \\
  \RULE{\cok{5}{r}{s,J}}
       {\rdy{\name{r}}{\name{s}}}
       {\head{s},J}
  \\[2ex]
  \RULE{\de{r}}
       {}
       {\naf\ok{\name{r}}}
  \\[2ex]
  \RULE{t(r,s,t)}
       {\PREC{\name{r}}{\name{t}}}
       {\PREC{\name{r}}{\name{s}},\PREC{\name{s}}{\name{t}}}
  \\
  \RULE{as(r,s)}
       {{\neg(\PREC{\name{s}}{\name{r}})}}
       {\PREC{\name{r}}{\name{s}}} \ \mbox{.}
\end{array}
\]
\end{definition}
%
First of all,
it is important to note that the original program $\Pi$ is contained in the
image $\mathcal{U}(\Pi)$ of the translation.
As we show in Proposition~\ref{thm:translation:results},
this allows us to construct (standard) answer sets of~$\Pi$ within answer sets
of~$\mathcal{U}(\Pi)$.
Such an answer set can be seen as the \emph{guess} in a guess-and-check approach;
it corresponds to Condition~1 in Definition~\ref{def:be:extension:general}.

The corresponding \emph{check},
viz.\ Condition~2 in Definition~\ref{def:be:extension:general},
is accomplished by the remaining rules in $\tau(r)$.
Before entering into details, let us stress the principal differences among
these rules and those used in translation $\mathcal{T}$ in
Section~\ref{sec:order:preservation}:
\begin{enumerate}
\item
      The  first group of rules is ``synchronised'' with the original rules in
      $\Pi$.
      That is, except for the literals in $\pbody{\ap{2}{r}}$, all primed body
      literals are doubled in the original language.
\item
      As with the encoding in Section~\ref{sec:order:variations}, 
      the second group of rules accommodates the modified strategy by adding a
      fifth rule.
\item
      Integrity constraint \de{r} is added for ruling out unsuccessful
      candidate answer sets of $\Pi$.
\end{enumerate}

As before, 
the first group of rules of $\tau(r)$ expresses applicability 
and blocking conditions.
The rules of form $\ap{i}{r}, i=1,2,$ aim at rebuilding the guessed answer set
within $\mathcal{L}'$.
They form in $\mathcal{L}'$ the prerequisite-free counterpart of the original
program.
In fact, the prerequisite of $\ap{2}{r}$ refers via
\(
\pbody{r}\subseteq\body{r}
\)
to the guessed extension in $\mathcal{L}$; no formula in $\mathcal{L}'$ must
be derived for applying $\ap{2}{r}$.
This accounts for the elimination of prerequisites in
Condition~\ref{l:one:b} of Definition~\ref{def:be:answer:set}.
The elimination of rules whose prerequisites are not derivable
is accomplished by rules of form~$\bl{1}{r}{L}$.
Rules of form~$\bl{2}{r}{L}$ guarantee that defaults are only defeatable
by rules with higher priority.
In fact, it is $K'$ that must be derivable in such a way only.

The application of rules according to the given preference information is
enforced by the second group of rules.
In addition to the four rules used in Definition~\ref{def:compilation},
an atom like \rdy{\name{r}}{\name{s}} is now also derivable if the
head of $s$ is true although $s$ itself is defeated
(both relative to the candidate answer set of $\Pi$ in $\mathcal{L}$).
This treatment of rules like $s$ through {\cok{5}{r}{s,J}} amounts to their
elimination from the preference handling process, as originated by the
second condition in Definition~\ref{def:be:operator}.
As before, the whole group of ``$\rdysym$''-rules allows us to derive
$\ok{\name{r}}$, indicating that $r$ may potentially be applied, whenever we
have for all $s$ with $\PRECM{r}{s}$ that $s$ has been applied or cannot
be applied,
or $s$ has already been eliminated from the preference handling process.

Lastly, $\de{r}$ rules out unsuccessful attempts in rebuilding the answer set
from $\mathcal{L}$ within $\mathcal{L}'$ according to the given preference
information.
In this way, we eliminate all answer sets that do not respect preferences.

For illustration, reconsider program $\Pi_{\ref{eq:five:one}}$:
\[
\begin{array}{rcrcl}
r_1 & = &      b & \LPif & a, \naf\neg b
\\
r_2 & = & \neg b & \LPif & \naf     b
\\ 
r_3 & = &      a & \LPif & \naf\neg a
\end{array}
\]
with $r_1$ being most preferred, then $r_2$, and finally $r_3$.
We get, among others:
\[
\begin{array}{lllp{5pt}lllp{5pt}lll}
  b'     & \LPif & \applied{\name{1}}
  &&                               
  \neg b'& \LPif & \applied{\name{2}}
  &&                               
  a'     & \LPif & \applied{\name{3}}
  \\
  \applied{\name{1}}& \LPif & \ok{\name{1}},
  &&                                        
  \applied{\name{2}}& \LPif & \ok{\name{2}},
  &&                                        
  \applied{\name{3}}& \LPif & \ok{\name{3}},
  \\                           
  && a,
  &&                                                           
  &&   
  &&                                                           
  &&   
  \\                           
  &&    \naf\neg b,  
  &&                                                           
  &&    \naf     b,
  &&                                                           
  &&    \naf\neg a,
  \\                           
  &&    \naf\neg b'
  &&                                                         
  &&    \naf     b'
  &&                                                         
  &&    \naf\neg a'
  \\
  \blocked{\name{1}}& \LPif & \ok{\name{1}}, 
  &&                                        
  &&                                        
  &&                                        
  &&                                        
  \\
  && \naf a,
  &&                                        
  &&                                        
  &&                                        
  &&                                        
  \\
  && \naf a'
  &&                                        
  &&                                        
  &&                                        
  &&                                        
  \\
  \blocked{\name{1}}& \LPif & \ok{\name{1}},
  &&                                        
  \blocked{\name{2}}& \LPif & \ok{\name{2}},
  &&                                        
  \blocked{\name{3}}& \LPif & \ok{\name{3}},
  \\
  && \neg b,  \neg b'
  &&
  &&      b,   b'
  &&
  && \neg a,  \neg a'  \mbox{.}                                        
\end{array}
\]

First, suppose there is an answer set of $\mathcal{U}(\Pi_{\ref{eq:five:one}})$ 
containing the standard answer set $X_2=\{a,\neg b\}$ of $\Pi_{\ref{eq:five:one}}$.
Clearly, $r_2$ and $r_3$ would contribute to this answer set, since they also
belong to $\mathcal{U}(\Pi_{\ref{eq:five:one}})$.
Having $\neg b$ denies the derivation of
\(
\applied{n_1}
\).
Also, we do not get
\(
\blocked{n_1}
\)
since we can neither derive
\(
\neg b'
\)
nor
\(
\naf a
\).
Therefore, we do not obtain \ok{n_2}.
This satisfies integrity constraint \de{r_2}, 
which destroys the putative answer set at hand.
Hence, as desired, $\mathcal{U}(\Pi_{\ref{eq:five:one}})$ has no answer set
including $X_2$.

For a complement,
consider the $\PRECMo^{\BE}$-preserving answer set $X_1=\{a, b\}$ of
$\Pi_{\ref{eq:five:one}}$.
In this case, $r_3$ and $r_1$ apply.
Given \ok{n_1} and $a$, we derive $b'$ and \applied{n_1}.
This gives \ok{n_2}, $b$,  and $b'$, from which we get \blocked{n_2}.
Finally, we derive \ok{n_3} and $a'$ and \applied{n_3}.
Unlike above, no integrity constraint is invoked and we obtain an answer set
containing $a$ and $b$.

For another example, consider the program given in~(\ref{eq:three}).
While $\Pi_{\ref{eq:three}}$ has two standard answer sets, it has no
\BEpreferred{} ones under the ordering imposed in~(\ref{eq:three}).
Suppose $\mathcal{U}(\Pi_{\ref{eq:three}})$ had an answer set containing $a$ and $b$.
This yields \rdy{n_2}{n_1} (via {\cok{5}{n_2}{n_1,b}) from which we get \ok{n_2}.
Having $a$ excludes 
\[
\ap{2}{r_2}  =  \applied{\name{2}}\LPif\ok{n_2},\naf a,\naf a' \ \mbox{.}
\]
Moreover, 
\[
\bl{2}{r_2}{a} = \blocked{\name{2}}\LPif\ok{n_2},a'
\]
is inapplicable
since $a'$ is not derivable (by higher ranked rules).
We thus cannot derive \ok{n_3}, which makes \de{r_3} destroy the putative
answer set.

\subsection{Formal Elaboration}
\label{sec:be:properties}

The next theorem gives the major result of this section.
%
\begin{theorem}\label{thm:brewka:eiter}
  Let $(\Pi,\PRECMo)$ be a statically ordered logic program
  and
  let $X$ be a consistent set of literals.
  
  Then,
  $X$ is a \BEpreferred{} answer set of $(\Pi,\PRECMo)$
  iff
  $X=Y\cap\mathcal{L}_{\mathcal{A}}$
  for some answer set $Y$ of
  \(
  \mathcal{U}(\Pi)\cup\{(\PREC{n_1}{n_2})\LPif\,\mid (r_1, r_2) \in\;\PRECMo \}
  \).
\end{theorem}

In what follows, we elaborate upon the structure of the encoded logic programs:
%
\begin{proposition}\label{thm:translation:results}
  Let $\Pi$ be an ordered logic program over $\mathcal{L}$
  and
  let $X$ be a consistent answer set of
  \(
  \mathcal{U}(\Pi)
  \).

  Then, we have the following results:
  \begin{enumerate}
  \item\label{thm:translation:results:1}
    $X\cap\mathcal{L}$ is a consistent (standard) answer set of $\Pi$;
  \item\label{thm:translation:results:2}
    $(X\cap\mathcal{L})'=X\cap\mathcal{L}'$
    \quad
    $($or $L\in X$ iff $L'\in X$ for $L\in\mathcal{L})$;
  \item\label{thm:translation:results:3}
    $r\in\Pi\cap\GR{\mathcal{U}(\Pi)}{X}$
    iff
    $\ap{2}{r}\in\GR{\mathcal{U}(\Pi)}{X}$;
  \item\label{thm:translation:results:4}
    $r\in\Pi\setminus\GR{\mathcal{U}(\Pi)}{X}$
    iff
    $\bl{1}{r}{L}\in\GR{\mathcal{U}(\Pi)}{X}$ 
    or
    $\bl{2}{r}{L}\in\GR{\mathcal{U}(\Pi)}{X}$
    for some $L\in\pbody{r}\cup\nbody{r}$;
  \item\label{thm:translation:results:5}
    if $\cok{5}{r}{s,L}\in\GR{\mathcal{U}(\Pi)}{X}$,
    then $\bl{2}{s}{L}\in\GR{\mathcal{U}(\Pi)}{X}$
    for $L\in\nbody{s}$.
\end{enumerate}
\end{proposition}
%
The last property shows that eliminated rules are eventually found to be
inapplicable.
Thus, it is sufficient to remove $r'$ from the preference handling process in
$\mathcal{L}'$; the rule is found to be blocked anyway.

In analogy to Proposition~\ref{thm:results:i},
the following result shows that our encoding treats all rules and it gathers
complete knowledge on their applicability status.
%
\begin{proposition}\label{thm:be:results}
  Let $X$ be a consistent answer set of
  \(
  \mathcal{U}(\Pi)
  \)
  for ordered logic program $\Pi$.

  We have for all
  \(
  r\in\Pi
  \)
  that
  \begin{enumerate}
   \item\label{l:be2:results:2}
    $\ok{\name{r}}\in X$; and
  \item\label{l:be2:results:3}
    $\applied{\name{r}}\in X$
    iff\/
    $\blocked{\name{r}}\not\in X$.
  \end{enumerate}
\end{proposition}
%
The last result reveals an alternative choice for integrity constraint \de{r},
namely
\[
\LPif\naf\applied{\name{r}},\naf\blocked{\name{r}} \ \mbox{.}
\]

One may wonder how our translation avoids the explicit use of total extensions
of the given partial order.
The next theorem shows that these total extensions are reflected by the
subsequent derivation of $\oksym$-literals within the grounded enumerations of
the generating default rules.
%
\begin{theorem}\label{thm:be:ordering}
Let $X$ be a consistent answer set of\/
\[
\mathcal{U}'(\Pi)
=
\mathcal{U}(\Pi)\cup\{(\PREC{n_1}{n_2})\LPif\,\mid (r_1, r_2) \in\;\PRECMo \}
\]
for some statically ordered logic program $(\Pi,\PRECMo)$.
Let
\(
\langle s_i\rangle_{i\in I}
\)
be a grounded enumeration of \GR{\mathcal{U}'(\Pi)}{X}.

Define
\[
\hat{\Pi}
=
\Pi\setminus\{r\in\Pi\mid\head{r}\in X,\nbody{r}\cap X\neq\emptyset\}
\]
and, for all $r_1,r_2\in\hat{\Pi}$, define
\(
r_1\ll r_2
\)
iff $k_2<k_1$ where 
\(
s_{k_j}=\cokt{1}{r_j}
\)
for $j=1,2$.

Then,
$\ll$ is a total ordering on $\hat{\Pi}$ such
that
\(
(\PRECMo\cap\;({\hat{\Pi}\times \hat{\Pi}}))\subseteq\;\ll
\).
\end{theorem}
%
That is, whenever $\Pi=\hat{\Pi}$,
we have that $\ll$ is a total ordering on $\Pi$ such that
\(
\PRECMo\subseteq\ll
\).

Finally, one may ask why we do not need to account for the ``inherited''
ordering in Condition~\ref{l:two} of Definition~\ref{def:be:answer:set}.
In fact,
this is taken care of through the tags \applied{\name{r}} in the
consequents of rules \ap{2}{r} that guarantee an isomorphism between
$\Pi$ and $\{\ap{2}{r}\mid r\in\Pi\}$ in Definition~\ref{def:be:answer:set}.
More generally, such a tagging of consequents provides an effective
correspondence between the applicability of default rules and the presence of
their consequents in an answer set at hand.

\subsection{Extensions to the Dynamic Case}
\label{sec:be:dynamic}

The original elaboration of \BEpreferred{} answer sets in
Brewka~\& Eiter~\shortcite{breeit97a,breeit99a} deals with the static case only.%
\footnote{An extension to the dynamic case is briefly described for default
  theories in Brewka~\& Eiter~\shortcite{breeit98a}.}
In fact, our framework leaves room for two alternative extensions to the
dynamic case, depending on how (or: ``where'') preference information is
formed.
The more prescriptive option is to form preference information ``on the
fly'' within language $\mathcal{L}'$.
This is close to the dynamic setting explored in
Section~\ref{sec:order:preservation}.
The more descriptive approach is to gather all preference information inside
the standard answer set formed in language $\mathcal{L}$ and then to rely on
this when verifying preferences within $\mathcal{L}'$.

While the latter approach is already realised through $\mathcal{U}$ in
Definition~\ref{def:be:compilation},%
\footnote{A closer look at the encoding of static preferences in
  Theorem~\ref{thm:brewka:eiter} reveals that there are no ``primed''
  occurrences of preference atoms.}
the former is obtained by replacing all occurrences of $\prec$ in
{\cok{2}{r}{s}}, {\cok{3}{r}{s}}, and {\cok{4}{r}{s}} by $\prec'$
and by adding $(t(r,s,t))'$ and $(as(r,s))'$ so that $\prec'$ becomes a strict
partial order.
Let us refer to the variant of $\mathcal{U}$ obtained in this way by
$\mathcal{V}$.
The difference between both strategies can be explained by the following
example.
Consider ordered program
\(
\Pi_{\ref{eq:be:dynamic}}=\{r_1,\dots,r_4\}
\),
where
\begin{equation}
  \label{eq:be:dynamic}
  \begin{array}{rcrcl}
    r_1 & = & a & \LPif & \naf\neg a \ ,
    \\
    r_2 & = & b & \LPif & \naf\neg b \ ,
    \\ 
    r_3 & = & \PREC{\name{1}}{\name{2}} & \LPif & , 
    \\ 
    r_4 & = & \PREC{\name{3}}{\name{1}} & \LPif & \mbox{.}
  \end{array}
\end{equation}
Clearly, $\Pi_{\ref{eq:be:dynamic}}$ has a single answer set,
\(
\{a,b,\PREC{\name{3}}{\name{1}},\PREC{\name{1}}{\name{2}}\}
\).
While this answer set is accepted by the strategy underlying
$\mathcal{U}(\Pi_{\ref{eq:be:dynamic}})$,
it is denied by that of $\mathcal{V}(\Pi_{\ref{eq:be:dynamic}})$.
To see this,
observe that $\mathcal{V}$ necessitates that all preference information
concerning higher ranked rules has been derived before
a rule can be considered for application
(as formalised in Condition~\ref{def:order:preserving:x:tri} in 
Definition~\ref{def:order:preserving:x}).
This is however impossible for rule $r_1$ because it dominates $r_3$ which
contains preference information about $r_1$.
Unlike this, $\mathcal{U}$ accepts each rule application order tolerating
\PREC{\name{3}}{\name{1}} and \PREC{\name{1}}{\name{2}}.

In fact,
we can characterise the strategy underlying transformation $\mathcal{U}$ in
the general (dynamic) case by adapting its static concept of order
preservation, given in Definition~\ref{def:be:order:preserving}.
For this,
it is sufficient to substitute $\Pi$ and \PRECMo{} by \TA{\Pi} and \PRECMoi{X}
in Definition~\ref{def:be:order:preserving}.
In contrast to Definition~\ref{def:order:preserving:x}, 
this extension to the dynamic case avoids an additional condition assuring
that all preference information about a rule is available before it is
considered for application.
This works because $\mathcal{U}$ gathers all preference information in
language $\mathcal{L}$ and then relies on it when verifying preferences within
$\mathcal{L}'$.
Similar to the static setting, this dynamic strategy thus
separates preference formation from preference verification.
This is different from the prescriptive strategy of $\mathcal{V}$ that
integrates both processes and thus necessitates a concept of order
preservation including Condition~\ref{def:order:preserving:x:tri} in
Definition~\ref{def:order:preserving:x}. 

Both strategies $\mathcal{U}$ as well as $\mathcal{V}$ differ from the
strictly prescriptive strategies discussed in
Section~\ref{sec:order:preservation}.
To see this,
consider the program
\(
\Pi_{\ref{eq:be:dynamic:two}}=\{r_1,r_2,r_3\}
\)
with
\begin{equation}
  \label{eq:be:dynamic:two}
  \begin{array}{rcrcl}
    r_1 & = & a & \LPif & \naf\neg a\ ,
    \\
    r_2 & = & b & \LPif & \naf\neg b\ ,
    \\ 
    r_3 & = & \PREC{\name{1}}{\name{2}} & \LPif& a, b\ \mbox{.}
  \end{array}
\end{equation}
$\Pi_{\ref{eq:be:dynamic:two}}$ has a single answer set,
\(
\{a,b,\PREC{\name{1}}{\name{2}}\}
\).
Both $\mathcal{U}$ and $\mathcal{V}$ accept this set as a preferred one.
This is due to the above discussed elimination of groundedness from the
preference test.
Unlike this, $\mathcal{T}$ yields no answer set on $\Pi_{\ref{eq:be:dynamic:two}}$
because there is no way to apply $r_3$ before $r_1$, as stipulated by
Condition~\ref{def:order:preserving:x:tri} in Definition~\ref{def:order:preserving:x}. 

While it is easy to drop Condition~\ref{def:order:preserving:x:tri} 
in Definition~\ref{def:order:preserving:x},
it takes more effort to realise such a concept of order preservation.
First of all, one has to resort to a generate and test construction similar to
the one used above.
That is, apart from the original program in some language $\mathcal{L}$,
we need a constrained reconstruction process in a mirror language
$\mathcal{L'}$.
A corresponding translation can be obtained from the one in
Definition~\ref{def:be:compilation} as follows.

Given an ordered logic program $\Pi$,
let $\tau'(r)$ be the collection of rules obtained from $\tau(r)$ in
Definition~\ref{def:be:compilation} by
\begin{enumerate}
\item replacing {\ap{2}{r}} by
  \[
  \begin{array}{rrcl}
    \RULE{\ap{2}{r}}
         {\applied{\name{r}}}
         {\ok{\name{r}},\body{r},\body{r'}}; \quad \mbox{and}
  \end{array}
  \]

\item deleting {\cok{5}{r}{s,J}}.
  
\end{enumerate}
(Condition~1 leaves the prerequisites of rules intact, and Condition~2 
eliminates the filter expressed by Condition 2c in
Definition~\ref{def:be:order:preserving}.)

Then, define the logic program
$\mathcal{S}(\Pi)$ over $\mathcal{L}^\circ$ as
\(
\mathcal{S}(\Pi)
=
\Pi \; \cup \; \mbox{$\bigcup_{r\in\Pi}$}\tau'(r)
\).
One can then show that the answer sets of $\mathcal{S}(\Pi)$ are order
preserving in the sense of Definition~\ref{def:order:preserving},
once $\Pi$ and \PRECMo{} are replaced by \TA{\Pi} and \PRECMoi{X}.
This leaves us with an enumeration of the generating rules, as opposed to the
entire program and there is no need for stipulating
Condition~\ref{def:order:preserving:x:tri} in
Definition~\ref{def:order:preserving:x} anymore.

Note that the strategy underlying $\mathcal{S}$ amounts to a synergy between
that of $\mathcal{T}$ and $\mathcal{U}$:
While borrowing the guessing of preferences from $\mathcal{U}$,
it sticks to the integration of groundedness into the associated concept of
order preservation.
The above translations along with their properties are summarised in
Table~\ref{tab:summary}.

\begin{table}[t!]
\caption{Summary of dynamic translations.}\label{tab:summary}
\begin{minipage}{\textwidth}
    \begin{tabular}{lcccc}
\hline\hline
{Integrated features}\hspace{5cm}
      & $\mathcal{T}$ & $\mathcal{S}$ & $\mathcal{U}$ & $\mathcal{V}$
      \\\hline
preference verification and groundedness 
      &$\times$&$\times$&&
      \\
preference verification and  preference formation
      &$\times$&&&$\times$\\
\hline\hline
    \end{tabular}
\end{minipage}
\end{table}

%

\section{Implementation}
\label{sec:implementation}

Our approach has been implemented and serves as a front-end to the logic
programming systems \textsf{dlv}~\cite{dlv97a} and
\textsf{smodels}~\cite{niesim97a}.
These systems represent state-of-the-art implementations for logic programming
within the family of stable model semantics~\cite{gellif88b}.
For instance, \textsf{dlv} also admits strong negation and disjunctive rules.
The resulting compiler, called \textsf{plp}
(for \textsf{p}reference \textsf{l}ogic \textsf{p}rogramming),
is implemented in Prolog and is publicly available at
\begin{center}
  \texttt{http://www.cs.uni-potsdam.de/\~{}torsten/plp/}\ .
\end{center}
This URL contains also diverse examples taken from the literature,
including those given in the preceding sections.
The current implementation runs under the Prolog systems ECLiPSe and SICStus
and comprises roughly 1400 lines of code.
It currently implements the three major strategies elaborated upon in this paper,
namely  $\mathcal{T}, \mathcal{U}$, and $\mathcal{W}$, given in
Sections~\ref{sec:order:preservation}, \ref{sec:brewka:eiter}, and
\ref{sec:order:variations}, respectively.

The \textsf{plp} compiler differs from the approach described above in two
minor respects:
first, the translation applies to named rules only, that is, it leaves unnamed
rules unaffected;
and second, it admits the specification of rules containing variables, whereby
rules of this form are processed by applying an additional instantiation
step.

The syntax of \textsf{plp} is summarised in Table~\ref{tab:syntax}.
%
%
\begin{table}[t!]
    \caption{The syntax of {\rm \textsf{plp}} input files.}
    \label{tab:syntax}
\begin{minipage}{\textwidth}
    \begin{tabular}{lll}
\hline\hline
      Meaning & Symbols & Internal\\
      \hline
      $\bot,\top$& \texttt{false/0}, \texttt{true/0}
      \\
      $\neg$     & \texttt{neg/1}, \texttt{-/1} (prefix) & \texttt{neg\_}$L$, $L\in\Lit$
      \\
      $\naf$     & \texttt{not/1}, \texttt{$\sim$/1} (prefix)
      \\
      $\wedge$   & \texttt{,/1} (infix; in body)
      \\
      $\vee$     & \texttt{;/1}, \texttt{v/2}, \texttt{|/2} (infix; in head)
      \\
      $\LPif$    & \texttt{:-/1} (infix; in rule)
      \\\hline
      $\prec$    & \texttt{</2} (infix) & \texttt{prec/2}
      \\\hline
        $n_r:{\langle\head{r}\rangle\LPif\langle\body{r}\rangle}$ \hspace{2mm}
      & $\langle\head{r}\rangle\text{\texttt{ :- }}\mathtt{name(}n_r\mathtt{)},\langle\body{r}\rangle$ \hspace{2mm}
      \\
      & $\langle\head{r}\rangle\text{\texttt{ :- }}\mathtt{[}n_r\mathtt{]},\langle\body{r}\rangle$ 
      \\\hline
      $\oksym,\rdysym$ && \texttt{ok/1}, \texttt{rdy/2}
      \\
      $\appliedsym,\blockedsym$ && \texttt{ap/1}, \texttt{bl/1}\\
\hline\hline
    \end{tabular}
\end{minipage}
\end{table}
For illustration, we give in Figure~\ref{fig:example} the file comprising
Example~(\ref{eq:example}).
Once this file is read into \textsf{plp} it is subject to multiple
transformations.
Most of these transformations are rule-centered in the sense that they
apply in turn to each single rule.
The first phase of the compilation is system independent and corresponds more
or less to the transformations given in the preceding sections.
While the original file is supposed to have the extension \texttt{lp},
the result of the system-independent compilation phase is kept in an
intermediate file with extension \texttt{pl}.
Applying the implementation of transformation $\mathcal{T}$ to the source file
in Figure~\ref{fig:example} yields at first the file in
Figure~\ref{fig:example:pl}.
%
%
\begin{figure}[t!]
\figrule
  \begin{center}
\begin{verbatim}
    neg a .
        b :- name(n2), neg a, not c.
        c :- name(n3), not b.
(n3 < n2) :- not d.
\end{verbatim}
  \end{center}
      \caption{The source code of Example~(\ref{eq:example}),
        given in file \texttt{example.lp}.}
    \label{fig:example}
\figrule
\begin{center}
\begin{verbatim}
neg_a.
b :- ap(n2).
ap(n2) :- ok(n2), neg_a, not c.
bl(n2) :- ok(n2), not neg_a.
bl(n2) :- ok(n2), c.
c :- ap(n3).
ap(n3) :- ok(n3), not b.
bl(n3) :- ok(n3), b.
prec(n3, n2) :- not d.
ok(X) :- name(X), rdy(X, n2), rdy(X, n3).
rdy(X,Y) :- name(X), name(Y), not prec(X,Y).
rdy(X,Y) :- name(X), name(Y), prec(X,Y), ap(Y).
rdy(X,Y) :- name(X), name(Y), prec(X,Y), bl(Y).
neg_prec(Y, X) :- name(X), name(Y), prec(X,Y).
prec(X,Z) :- name(X), name(Z), name(Y), prec(X,Y), prec(Y,Z).
name(n3).                   name(n2).
false :- a, neg_a.          false :- b, neg_b.
false :- c, neg_c.          false :- d, neg_d.
false :- name(X), name(Y), prec(X,Y), neg_prec(X,Y).
\end{verbatim}
  \end{center}
    \caption{The (pretty-printed) intermediate code resulting from
      Example~(\ref{eq:example}), given in file \texttt{example.pl}.}
    \label{fig:example:pl}
\figrule
\end{figure}
%
%
We see that merely rules $r_2$ and $r_3$ are translated, while $r_1$ and $r_4$
are unaffected.
This is reasonable since only $r_2$ and $r_3$ are subject to preference
information.
This is similar to the Prolog rule representing \cokt{1}{r}, insofar as its
body refers to $r_2$ and $r_3$ only.
Classical negation is implemented in the usual way by appeal to integrity
constraints, given at the end of Figure~\ref{fig:example:pl}.

While this compilation phase can be engaged explicitly by the command
\texttt{lp2pl/1}, for instance in SICStus Prolog by typing
\begin{quote}
  \texttt{| ?- lp2pl('Examples/example').} \ ,
\end{quote}
one is usually interested in producing system-specific code that is directly
usable by either \textsf{dlv} or \textsf{smodels}.
This can be done by means of the commands \texttt{lp2dlv/2} and
\texttt{lp2sm/2},\footnote{These files are themselves obtainable from the
  intermediate \texttt{pl}-files via commands \texttt{pl2dlv/1} and
  \texttt{pl2sm/1}, respectively.}
which then produce system-specific code resulting in files having extensions
\texttt{dlv} and \texttt{sm}, respectively.
These files can then be fed into the respective system by a standard command
interpreter, such as a UNIX shell, or from within the Prolog system through
commands \texttt{dlv/1} or \texttt{smodels/1}.
For example, after compiling our example by \texttt{lp2dlv}, we may proceed as
follows.

{\small%
\begin{verbatim}
| ?- dlv('Examples/example').
Calling :dlv  Examples/example.dlv
dlv [build BEN/Jun 11 2001   gcc 2.95.2 19991024 (release)]

{true, name(n2), name(n3), neg_a, ok(n2), rdy(n2,n2), rdy(n2,n3),
 rdy(n3,n3),prec(n3,n2), neg_prec(n2,n3), ap(n2), b, rdy(n3,n2), 
 ok(n3), bl(n3)}
\end{verbatim}}
\noindent
Both commands can be furnished with the option \texttt{nice} (as an
additional argument) in order to strip off auxiliary predicates.

{\small%
\begin{verbatim}
| ?- dlv('Examples/example',nice).
Calling :dlv  -filter=a [...] -filter=neg_d Examples/example.dlv
dlv [build BEN/Jun 11 2001   gcc 2.95.2 19991024 (release)]

{neg_a, b}
\end{verbatim}}
The above series of commands can be engaged within a single one by means of
\texttt{lp4dlv/2} and \texttt{lp4sm/2}, respectively.
The overall (external) comportment of \textsf{plp} is illustrated in
Figure~\ref{fig:functioning}.%
%
%
\begin{figure}[t!]
\figrule
  \begin{center}
    \newcommand{\SOFTWARE}[1]{\makebox(0,0){\Ovalbox{\makebox(18,6)[c]{\textsf{#1}}}}}
    \newcommand{\FILE}[1]{\makebox(0,0){\framebox(22,8)[c]{$\langle$\textit{file}$\rangle$\texttt{.#1}}}}
    \newcommand{\STDIO}{\makebox(0,0){\footnotesize\shortstack{\texttt{Standard}\\\texttt{output}}}}
    \newcommand{\TEXTTT}[1]{\makebox(0,0){\texttt{#1}}}
    \setlength{\unitlength}{.9mm}
    \begin{picture}(140,80)(-72,-40)

      \put(-60,  0){\FILE{lp}}
      \put(-30,  0){\SOFTWARE{plp}}
      \put(  0,+10){\FILE{dlv}}
      \put(  0,-10){\FILE{sm}}
      \put( 30,+10){\SOFTWARE{dlv}}
      \put( 30,-10){\SOFTWARE{smodels}}
      \put( 58,+10){\STDIO}
      \put( 58,-10){\STDIO}

      \put(-47,  0){\vector(1, 0){5}}
      \put(-18,  0){\vector(1, 2){5}}
      \put(-18,  0){\vector(1,-2){5}}
      \put( 13,+10){\vector(1, 0){5}}
      \put( 13,-10){\vector(1, 0){5}}
      \put( 42,+10){\vector(1, 0){5}}
      \put( 42,-10){\vector(1, 0){5}}

      \put(-60,+5){\qbezier(0,0)(30, 30)(60,10)}
      \put(-60,-5){\qbezier(0,0)(30,-30)(60,-10)}
      \put(-60,+5){\qbezier(0,0)(30, 60)(120,10)}
      \put(-60,-5){\qbezier(0,0)(30,-60)(120,-10)}

      \put(-30,+25){\TEXTTT{lp2dlv/1}}
      \put(-40,+35){\TEXTTT{lp4dlv/1}}
      \put(-30,-25){\TEXTTT{lp2sm/1}}
      \put(-40,-35){\TEXTTT{lp4sm/1}}
    \end{picture}
  \end{center}
    \caption{Compilation with \textsf{plp}: external view.}
    \label{fig:functioning}
\figrule
\end{figure}
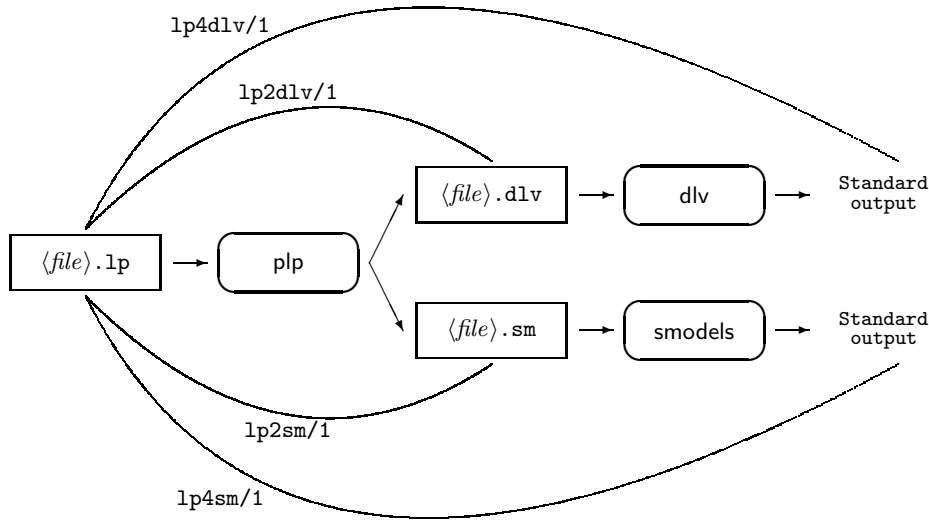

For treating variables, some more preprocessing is necessary for
instantiating the rules prior to their compilation.
The presence of variables is indicated by file extension \texttt{vlp}.
The contents of such a file is first instantiated by systematically replacing
variables by constants and then freed from function symbols by replacing
terms by constants, \egc \texttt{f(a)} is replaced by \texttt{f\_a}.
This is clearly a rather pragmatic approach.
A more elaborated compilation would be obtained by proceeding right from the
start in a system-specific way.

For illustration, we give in Figure~\ref{fig:legal} a formalisation of the
popular legal reasoning example due to~\cite{gordon93a}:
\begin{quote}\em
``%
A person wants to find out if her security interest in a certain ship
is `perfected', or legally valid. 
This person has possession of the ship, but has not filed a financing
statement.
According to the code UCC, a security interest
can be perfected by taking possession of the ship.
However, the federal Ship Mortgage Act (SMA) states that a security interest
in a ship may only be perfected by filing a financing statement.
Both UCC and SMA are applicable; the question is which takes precedence
here.
There are two legal principles for resolving such conflicts.
\emph{Lex Posterior} gives precedence to newer laws;
here we have that UCC is more recent than SMA.
But \emph{Lex Superior} gives precedence to laws supported by the higher
authority;
here SMA has higher authority since it is federal law.%
''
\end{quote}
%
\begin{figure}[t!]
\figrule
  \begin{center}
\begin{verbatim}
    perfected :- [ucc], possession,             not neg perfected.
neg perfected :- [sma], ship, neg finstatement, not     perfected.

(Y < X) :- [lex_posterior(X,Y)], 
           newer(X,Y), not neg (Y < X).
(X < Y) :- [lex_superior(X,Y)], 
           state_law(X), federal_law(Y), not neg (X < Y).

possession.              newer(ucc,sma).   
ship.                    federal_law(sma). 
neg finstatement.        state_law(ucc).   

(lex_posterior(X,Y) < lex_superior(X,Y)).
\end{verbatim}
  \end{center}
  \caption{A formalisation of Gordon's legal reasoning example, given in
      file \texttt{legal.vlp}.}
    \label{fig:legal}
\figrule
\end{figure}
%

Compiling file \texttt{legal.vlp} yields the following result.

{\small%
\begin{verbatim}
| ?- vlp4dlv('Examples/Variables/legal',nice).
Calling :dlv  -filter=[...] Examples/Variables/legal.dlv
dlv [build BEN/Jun 11 2001   gcc 2.95.2 19991024 (release)]

{possession, ship, neg_finstatement, newer(ucc,sma),
 state_law(ucc), federal_law(sma), neg_perfected}
\end{verbatim}}

More details on the \textsf{plp} compiler can be found at the above cited URL
or in~Delgrande, Schaub,~\& Tompits~\shortcite{descto00b}.

\section{Other Approaches to Preference}
\label{sec:related:work}

The present preference framework has its roots in an approach proposed in 
Delgrande~\& Schaub~\shortcite{delsch97a,delsch00a} for handling preference 
information in default logic. 
There, preference information is expressed using an
ordered \emph{default theory}, consisting of default rules, world knowledge,
and a preference relation between default rules. 
Such ordered default theories are then translated into standard default 
theories, determining extensions of the given theory in which the preferences 
are respected. 
Both static and dynamic orderings are discussed, albeit in a less uniform 
manner than realised in the present case.
More specifically, Delgrande~\& Schaub~\shortcite{delsch97a,delsch00a}
define the notion of an order 
preserving extension for static preference orderings only.
As well, the encoding of dynamic preferences relies on an additional 
predicate $\nprec$, expressing non-preference between two (names of) rules.
In contrast,
the current framework is specified right from the beginning for the dynamic 
case, and static preferences are just a special instance of dynamic ones.
Furthermore, Delgrande~\& Schaub~\shortcite{delsch97a,delsch00a} are primarily 
concerned with a \emph{specific} preference strategy, similar to the one 
expressed by Definition~\ref{def:order:preserving}, whilst here the emphasis is to 
demonstrate the flexibility of the framework by providing encodings for 
\emph{differing preference strategies}. 
Lastly, in contradistinction to other preference approaches requiring 
dedicated algorithms, the existence of readily available solvers for the 
answer set 
semantics, like \textsf{dlv} and \textsf{smodels}, makes a realisation of 
the current approach straightforward, as documented by the discussion in
the previous section.

The problem of dealing with preferences in the context of nonmonotonic rule
systems has attracted extensive interest in the past decades.
In fact, for almost every nonmonotonic approach there exist prioritised 
versions designed to handle preference information of some form (cf., 
\egc 
\cite{geprpr89,konolige88b,rintanen94a,bcdlp93a,nebel98a,eitgot95,brewka89e,brewka96a}). 
Prioritised versions of default logic and logic programming under the
answer set semantics
includes~\cite{baahol93a,brewka94a,rintanen98b,sakino96,gelson97a,zhafoo97a}
as well as those approaches discussed earlier.
As argued in Delgrande~\& Schaub~\shortcite{delsch00a}, these approaches can 
be divided into \emph{descriptive} and \emph{prescriptive} approaches. 
In the former case, one has a ``wish list'' where the intent is that one 
way or another the highest-ranked rules be applied.
In the latter case the ordering reflects the order in which rules should
be applied.
The relationship of Doyle and Wellman's work \cite{doywel91a} to such
preferences is also discussed in Delgrande~\& Schaub~\shortcite{delsch00a}.

The approach of Rintanen~\shortcite{rintanen98b} addresses descriptive 
preference orders in default logic.
This notion of preference differs conceptually from the preference
orderings dealt with here.
In particular, Rintanen's method is based on a meta-level filtering 
process eliminating extensions which are not considered to be preferred 
according to certain criteria.  
This method yields a higher worst-case complexity than standard default 
logic (providing the polynomial hierarchy does not collapse), and a similar 
method applied to logic programs under the answer set semantics results 
likewise in an increase of complexity. 
Consequently, Rintanen's approach is not polynomial-time translatable into 
standard logic programs (given the same proviso as before) and can thus not 
be efficiently represented within our framework. 
Furthermore, as argued in Brewka~\& Eiter~\shortcite{breeit99a}, 
Rintanen's approach does not obey Principle II, and 
exhibits counter-intuitive results in some cases.

For prescriptive approaches, Baader~\& Hollunder~\shortcite{baahol93a} and 
Brewka~\shortcite{brewka94a} present
prioritised variants of default logic in which the iterative specification 
of an extension is modified.
Preference information is given at the meta-level (thus, only static 
preferences are considered), and a default is only applicable at an 
iteration step if no $<$-greater default is applicable.%
\footnote{These authors use $<$ in the reverse order from us.}
The primary difference between these approaches rests on the number of
defaults applicable at each step: while Brewka allows only for applying a 
single default that is maximal with respect to a total extension of $<$, 
Baader and Hollunder allow for applying all $<$-maximal defaults at each step.
Both approaches violate Principle~I, but, as shown by 
Rintanen~\shortcite{rintanen98a}, neither 
of them result in an increase of worst-complexity compared to standard default
logic.
Thus, the relevant reasoning tasks associated with these preference approaches 
can be translated in polynomial time into equivalent problems of standard 
default logic, which makes the two preference methods amenable for the
framework of Delgrande~\& Schaub~\shortcite{delsch97a,delsch00a}.
However, neither Brewka nor Baader and Hollunder deal with context-sensitive
preferences.

The previously discussed methods rely on a two-level approach, treating
preferences at the meta-level by defining an alternative semantics.
A similar methodology is inherent to most approaches to preference handling 
in extended logic programming.
One of the few exceptions is given by the approach in
Gelfond~\& Son~\shortcite{gelson97a}, which avoids defining a new
logic programming formalism in order to cope with preference information,
although it still utilises a two-layered approach in reifying rules and
preferences.
For example, a rule such as
\(
p\leftarrow r, \neg s, \mathit{not}\ q
\)
is expressed by the formula
\(
\textit{default}(n, p, [r, \neg s], [q])
\)
(or, after reification, by the corresponding \emph{term} inside a
\emph{holds}-predicate, respectively)
where $n$ is the name of the rule.
The semantics of a domain description is given in terms of a set of
domain-independent rules for predicates like \textit{holds}.
These rules can be regarded as a meta-interpreter for the domain description.
Interestingly, the approach is based on the notion of ``defeat'' (of
justifications) in contrast to an order-preserving consideration of rules, 
as found in our approach.
Also, the specific strategy elaborated upon in Gelfond~\& 
Son~\shortcite{gelson97a} differs from
the ones considered here in that the preference $d_1<d_2$
\emph{``stops the application of default $d_2$ if defaults $d_1$ and $d_2$ are
  in conflict with each other and the default $d_1$ is applicable''}
\cite{gelson97a}.
(Such strategies are also studied in Delgrande~\& Schaub~\shortcite{delsch00b}.)
For detecting such conflicts, however, the approach necessitates an extra
conflict-indicating predicate.
That is, one must state explicitly $\textit{conflict}(d_1,d_2)$ to indicate
that $d_1$ and $d_2$ are conflicting.
In principle, as with our framework, the approach of Gelfond and Son offers a 
variety of different preference handling instances.

Zhang and Foo~\shortcite{zhafoo97a} present an operational semantics for ordered
logic programs based on an iterative reduction to standard programs.
The approach admits both static and dynamic preferences, in which the dynamic
case is reduced to the static one.
Interestingly, the semantics has a nondeterministic flavour in the sense that 
an ordered program may be reduced  in more than one way to a standard 
program. 
As shown by Zhang~\shortcite{zhangXX}, the somewhat elaborate 
semantics results in a worst-case complexity which lies at the second 
level of the polynomial hierarchy. 
Thus, it is intrinsically harder than standard answer set semantics, 
providing the polynomial hierarchy does not collapse. 
Consequently, under this proviso, it is not possible to express  
Zhang and Foo's method within our framework in a polynomial way, 
unless our language is extended.
Their approach leads also to new answer sets, once preferences are added,
as illustrated by the following program:
\[
\begin{array}[t]{rcl}
r_1
&=&
\phantom{\neg} p \LPif{} \naf q_1
\\
r_2
&=& 
\neg p \LPif{} \naf q_2
\\
\end{array}
\qquad\text{ and }\qquad  
r_1 < r_2
\ \mbox{.}
\]
Zhang and Foo's approach yields two answer sets  (one with $p$
and one with $\neg p$).
In contrast, no preferred answer set is obtained under \DST-, \WZL-, or
\BE-preference.
Finally, their approach violates Principle~II of Brewka 
and Eiter~\shortcite{breeit99a}.

The method of Sakama and Inoue~\shortcite{sakino96} has the interesting feature that the 
given preference order is not a relation between (names of) rules,  
as in the previous frameworks, but represents a relation between 
\emph{literals}.%
\footnote{Cf.~\cite{gefpea92} for another approach to preferences on literals.}
This relation is used to determine a preference relation on the answer 
sets of a disjunctive logic program.\footnote{Disjunctive logic programs 
are characterised by permitting disjunctions in rule heads.}
Intuitively, an answer set $X_1$ is at least as preferable as an answer 
set $X_2$ iff there are literals $L_1\in X_1\setminus X_2$ and 
$L_2\in X_2\setminus X_1$ such that $L_1$ has at least the priority of 
$L_2$, and there is no literal $L_2'\in X_2 \setminus X_1$ which has 
strictly higher priority than $L_1$. 
An answer set $X$ is preferred iff there is no other answer set which is 
strictly preferred over $X$.
The minimality criterion on answer sets makes this approach (presumably) 
harder than standard answer set semantics for disjunctive logic programs, 
yielding a worst-case complexity which lies at the third level of the 
polynomial hierarchy.
Thus, even by restricting rules to the non-disjunctive case, Sakama and 
Inoue's approach does not admit a polynomial representation within our 
framework (providing the polynomial hierarchy does not collapse). 
As well, their approach does not satisfy Principle~II of 
Brewka~\& Eiter~\shortcite{breeit99a}.

Finally, Wang, Zhou, \& Lin~\shortcite{wazhli00} present an approach 
intermediate between the
prescriptive approach described in Section~\ref{sec:order:preservation}
and the more descriptive approach of Brewka and Eiter~\shortcite{breeit99a},
expressed in our framework in
Section~\ref{sec:brewka:eiter}.
An encoding of Wang, Zhou, and Lin's approach in our framework was given in
Section~\ref{sec:order:variations}.
We obtain the following relations among these three approaches.
For differentiating the answer sets obtained in these approaches, let us
refer to the answer sets obtained in the approach of 
Wang \emph{et al.}~\shortcite{wazhli00} as \emph{\WZL-preferred}, 
in analogy to the term
\emph{\BE-preferred answer set} introduced earlier,  referring to an answer
set obtained in the approach of Brewka~\& Eiter~\shortcite{breeit99a}.

Then,  for a given statically ordered logic program $(\Pi,<)$, 
the following results can be shown \cite{schwan01b}:
\begin{enumerate}[(iii).]
\renewcommand{\theenumi}{(\roman{enumi})}
\item
Every $<$-preserving answer set (in the sense of 
Section~\ref{sec:order:preservation}) is \WZL-preferred.
\item
Every \WZL-preferred answer set is \BE-preferred.
\item
Every \BE-preferred answer set is an answer set.
\end{enumerate}
In no instance do we obtain the converse.
For example,
(\ref{eq:wang}) is a counterexample in the first case, and
(\ref{eq:be:dynamic:two}) in the second.
Thus, the framework illustrates that the preference approaches of 
Section~\ref{sec:order:preservation}, Wang \emph{et al}.~\shortcite{wazhli00}, and 
Brewka~\& Eiter~\shortcite{breeit99a} form a hierarchy of
successively-weaker notions of preference.
In fact, this hierarchy is of practical value since it allows us to apply the
concept of a ``back-up''-semantics~\cite{hoewit00a} for validating strategies on
specific ordered logic programs.
The idea is to first interpret a program with respect to the strongest
semantics, and then to gradually weaken the strategy until a desirable set of
conclusions is forthcoming.

A feature common to the latter three approaches is that, for statically
ordered programs, the addition of preferences never increases the number
of preferred answer sets.%
\footnote{Cf.\ Proposition~5.4 in Brewka~\& Eiter~\shortcite{breeit99a}.}
In fact, we have seen above that preferences may sometimes be too strong
and deny the existence of preferred answer sets although standard answer sets
exist.
This is because preferences impose additional dependencies among rules that
must be respected by the standard answer sets.
In a related paper \cite{schwan01c}, it is shown that programs
whose order amounts to a stratification guarantee a unique answer set.
Furthermore, Delgrande~\& Schaub~\shortcite{delsch00a} show how 
graph structures, as introduced in Papadimitriou~\&  Sideri~\shortcite{papsid94},
give sufficient conditions for guaranteeing preferred extensions of ordered
default theories;
these results obviously carry over to extended logic programs due to the
results in Gelfond~\& Lifschitz~\shortcite{gellif91a}.

\section{Discussion}
\label{sec:discussion}

\subsection{Further Issues and Refinements}
\label{sec:extensions}

We briefly sketch the range of further applicability and point out
distinguishing features of our framework here.

First, we draw the reader's attention to the expressive power offered by
dynamic preferences in connection with variables in the input language.
Consider the rule
\begin{equation}
\label{eqn:genpref}
\PREC{n_1(x)}{n_2(y)}\LPif p(y), \naf (x=c),
\end{equation}
where $n_1(x),n_2(y)$ are names of rules containing the variables $x$ and $y$,
respectively.
Although such a rule represents only its set of ground instances,
it nonetheless constitutes a much more concise specification.
Since most other approaches employ static preferences of the form
\(
\PREC{n_1(x)}{n_2(y)}\LPif,
\)
these approaches would necessarily have to express (\ref{eqn:genpref})
as an enumeration of static ground preferences rather than a single rule.

Second, we note that our methodology is also applicable to disjunctive logic
programs (where rule heads are disjunctions of literals).
To see this, observe that the transformed rules unfold the conditions
expressed in the body of the rules, while a rule's head remain untouched,
as manifested by rule
\(
\ap{1}{r} 
\).
In addition to handling preferences about disjunctive rules,
our approach allows us to express disjunctive preferences,
such as
\[
(\PREC{\name{3}}{\name{2}})\vee(\PREC{\name{3}}{\name{4}})\LPif \neg a\ ,
\]
saying that if $\neg a$ holds, then either $r_{2}$ or $r_{4}$ are preferred to
$r_{3}$.
Thus, informally, given $\neg a$ along with three mutually exclusive rules
$r_1,r_2,r_3$, 
the above preference results in two rather than three answer sets.

Finally, there is a straightforward way of accommodating preferences on
literals, as for instance put forward in Sakama~\& Inoue~\shortcite{sakino00a}.
A preference ${p}\lhd{q}$ among two literals
\(
p,q\in\mathcal{L}
\)
can then be realised by stipulating that
\PRECM{{r_1}}{{r_2}}
whenever
\(
\head{r_1}=p
\)
and
\(
\head{r_2}=q
\).

\subsection{Concluding Remarks}

We have described a general {\em methodology} in which logic programs
containing preferences on the set of rules can be compiled into logic
programs under the answer set semantics.
An ordered logic program, in which preference atoms appear in the program
rules, is transformed into a second, extended logic program.
In this second program, no explicit preference information appears, yet
preferences are respected, in that the answer sets obtained in the
transformed theory correspond to the preferred answer sets of the original
theory.

In our main approach, we provide an encoding for capturing a strongly
prescriptive notion of preference, in which rules are to be applied in
accordance with the prescribed order.
Since this approach admits a dynamic (as well as, trivially, static) notion
of preference, one is able to specify preferences holding in a given
context, by ``default'', or where one preference depends on another.
We also show how the approach of Brewka~\& Eiter~\shortcite{breeit99a}
can be expressed in our framework for preferences.
Further, we extend their approach to handle the dynamic case as well.

These translations illustrate the generality of our framework.
As well, they shed light specifically on Brewka and Eiter's approaches,
since we can provide a translation and encoding of their approaches
into extended logic programs.
In a sense then, our overall methodology provides a means of
\emph{axiomatising} different preference orderings, in that the image of
a translation shows how an approach can be encoded in an extended logic
program.
The contrast between
Definitions~\ref{def:compilation}~and~\ref{def:be:compilation} for
example illustrate succinctly how our main approach to preferences and that of
Brewka~\& Eiter~\shortcite{breeit99a} relate.
In general then our approach provides a uniform basis within which diverse
approaches and encodings can be compared.
This is illustrated by the hierarchy of approaches discussed at the end of
Section~\ref{sec:related:work}.
Similarly, the framework allows us to potentially formalise the interaction
of several orderings inside the same framework, and to specify in the
theory how they interact.

In all of these approaches we can prove that the resulting transformations
accomplish what is claimed:
that the appropriate rules in the image of the transformation are treated
in the correct order, that all rules are considered, that we do indeed obtain
preferred answer sets, and so on.
In addition to the formal results, we illustrate the generality of the
approach by formalising an example due to Gordon~\shortcite{gordon93a}, 
as well as giving examples
of context-based preference, canceling preferences,
preferences among preferences, and preferences by default.

Our approach of translating preferences into extended logic programs has
several other advantages over previous work.
First, we avoid the two-level structure of such work.
While the previous ``meta-level'' approaches must commit themselves to a
semantics and a fixed strategy, our approach (as well as that of 
Gelfond~\& Son~\shortcite{gelson97a}) is very flexible with 
respect to changing strategies, and
is open for adaptation to different semantics and different concepts of
preference handling (as illustrated by our encoding and extension of
the method of Brewka~\& Eiter~\shortcite{breeit99a}).
Second, for a translated program in our approach, any answer set is
``preferred'', in the sense that {\em only} ``preferred'' answer sets (as
specified by the ordering on rules) are produced.
In contrast, many previous approaches, in one fashion or another, select
among answer sets for the most preferred.
Hence one could expect the present approach to be (pragmatically) more
efficient, since it avoids the generation of unnecessary answer sets.
As well, as Section~\ref{sec:related:work} notes, some of these other
approaches are at a higher level of the complexity hierarchy than standard
extended logic programs.
Lastly, in ``compiling'' preferences into extended logic programs, some light
is shed on the foundations of extended logic programs.
In particular, we show implicitly that for several notions of explicit,
dynamic, preference expressed among rules, such preference information
does not increase the expressibility of an extended logic program.

Finally, this paper demonstrates that these approaches are easily
implementable.
We describe here a compiler (described more fully in 
Delgrande, Schaub,~\& Tompits~\shortcite{descto00b})
as a front-end for \texttt{dlv} and \texttt{smodels}.
The current prototype is available at
\begin{center}
\texttt{http://www.cs.uni-potsdam.de/\~{}torsten/plp/} \ .
\end{center}
This implementation is used as the core reasoning engine in a knowledge-based
information system for querying Internet movie databases, currently 
developed at the Vienna University of Technology~\cite{eifisato01a}.
Also, it serves for modeling linguistic phenomena occurring in phonology and
syntax~\cite{bemesc01a}, which are treated in linguistics within Optimality
Theory~\cite{prismo93,kager99}.

\section*{Acknowledgements}

The authors would like to thank the anonymous referees for their
constructive comments which helped to improve the paper.
The first author was partially supported by a Canadian NSERC Research
Grant;
the second author was partially supported by the German Science Foundation
(DFG) under grant FOR~375/1-1,~TP~C; and
the third author was partially supported by the Austrian Science Fund (FWF) 
under grants N~Z29-INF and P13871-INF.

\appendix
\section{Proofs}
\label{sec:proofs}

For later usage, we provide the following lemmata without proof.
%
\begin{lemma}\label{lem:one}
  Let $X$ be an answer set of logic program $\Pi$.
  Then, we have for $r\in\Pi$
  \begin{enumerate}
  \item if $\pbody{r}\subseteq X$ and $\nbody{r}\cap X =\emptyset$,
    then $\head{r}\in X$;
  \item if $r\in\GR{\Pi}{X}$, then $\head{r}\in X$.
  \end{enumerate}
\end{lemma}
The inverse of each of these propositions holds, whenever
\(
\head{r}=\head{r'}
\)
implies
\(
r=r'
\).
%
%
\begin{lemma}\label{lem:two}
  Let $X$ be an answer set of logic program $\Pi$.

  Then, there is a grounded enumeration 
  \(
  \langle r_i\rangle_{i\in I}
  \)
  of $\GR{\Pi}{X}$.  
\end{lemma}
%
That is, we have for consistent $X$ that
\(
\pbody{r_i}\subseteq\{\head{r_j}\mid j<i\}
\).

The following result is taken from Proposition 5.1 in 
Brewka~\& Eiter~\shortcite{breeit99a}.
%
\begin{theorem}\label{thm:be:characterisation}
  Let $(\Pi,\ll)$ be a fully ordered program
  and
  let $X$ be an answer set of $\Pi$.

  Then, $X$ is a \BEpreferred{} answer set of $(\Pi,\ll)$
  iff
  for each rule $r\in\Pi$ with
  \begin{enumerate}
  \item $\pbody{r}\subseteq X$ and
  \item $\head{r}\not\in X$
  \end{enumerate}
  there is some rule $r'\in\GR{\Pi}{X}$ such that
  \begin{enumerate}
  \item $r\ll r'$ and
  \item $\head{r'}\in\nbody{r}$.
  \end{enumerate}
\end{theorem}

In the sequel, we give the proofs for the theorems in the order the latter
appear in the body of the paper.

\subsection{Proofs of Section~\ref{sec:order:preservation}}

\begin{proof*}[Proof of Proposition~\ref{thm:strict:order}]
Let $X$ be a consistent answer set of
\(
\mathcal{T}(\Pi)
\) 
for an ordered program $\Pi$.
\begin{enumerate}
\item 
Since rules $t(r,r',r'')$ and $as(r,r')$ are members of $\mathcal{T}(\Pi)$, 
for any $r,r',r''\in\Pi$, the relation $\PRECMoi{X}$ is clearly transitive
and antisymmetric. 
Thus, $\PRECMoi{X}$ is a strict partial order.

\item 
Suppose $\Pi$ contains static preferences only, \iec $\Pi=\Pi'\cup\Pi''$,
where $\Pi''\subseteq\{(\PREC{\name{r}}{\name{r'}})\LPif \mid r,r'\in\Pi'\}$ 
and $\Pi'$ contains no occurrences of preference atoms 
of form $\PREC{n}{m}$ 
for some names $n,m$.
Hence, for each consistent answer set $Z,Z'$ of $\Pi$ and any 
$(\PREC{\name{r}}{\name{r'}})\LPif \ \in\Pi''$, we have that 
$(\PREC{\name{r}}{\name{r'}})\in Z$ iff $(\PREC{\name{r}}{\name{r'}})\in Z'$.
Since $X$ is assumed to be consistent, well-known properties of answer sets
imply that any other answer set $Y$ of $\Pi$ is also consistent. 
It follows that $(\PREC{\name{r}}{\name{r'}})\in X$ iff 
$(\PREC{\name{r}}{\name{r'}})\in Y$, for any answer set $Y$ of $\Pi$.
From this, $\PRECMoi{X}=\PRECMoi{Y}$ for any answer set $Y$ is an immediate
consequence. $\mathproofbox$
\end{enumerate}
\end{proof*}


\begin{proof*}[Proof of Proposition~\ref{thm:results:i}]
Let $X$ be a consistent answer set of
\(
\mathcal{T}(\Pi)
\) 
for an ordered program $\Pi$.
\begin{itemize}
\item [\ref{l:results:2}+\ref{l:results:3}.]
We prove both propositions by parallel induction on $<_X$.
\begin{description}
\item[Base.]
  Let $r$ be a maximal element of $<_X$.
  \begin{itemize}
  \item [\ref{l:results:2}.]
    By assumption,
    \(
    \PREC{\name{r}}{\name{r'}}\not\in X
    \)
    for all $r'\in\Pi$.
    This implies that
    \[
    \rdy{\name{r}}{\name{r'}}\LPif\ \in{\reduct{\mathcal{T}(\Pi)}{X}}
    \qquad
    \text{ for all }
    r'\in\Pi
    \ ,
    \]
    which results in
    \(
    \ok{\name{r}}\in\Th{\reduct{\mathcal{T}(\Pi)}{X}}
    \)
    because \Th{\reduct{\mathcal{T}(\Pi)}{X}} is closed under ${\reduct{\mathcal{T}(\Pi)}{X}}$.
    Hence,
    \(
    \ok{\name{r}}\in X,
    \)
    since $X=\Th{\reduct{\mathcal{T}(\Pi)}{X}}$.
  \item [\ref{l:results:3}.] 
    We distinguish the following two cases:
    \begin{itemize}
    \item If $\ap{2}{r}\in\GR{\mathcal{T}(\Pi)}{X}$,
      then $\applied{\name{r}}\in X$ by Lemma~\ref{lem:one}.
    \item If $\ap{2}{r}\not\in\GR{\mathcal{T}(\Pi)}{X}$,
      then we have in view of Property~\ref{l:results:2},
      namely
      \(
      \ok{\name{r}}\in X
      \),
      that one of the following two cases is true:
      \begin{itemize}
      \item If $\pbody{r}\not\subseteq X$,
        then there is some $L^{+}\in\pbody{r}$
        with $L^{+}\not\in X$.
        Hence,
        \(
        (\bl{1}{r}{L^{+}})^{+}\in\reduct{\mathcal{T}(\Pi)}{X}
        \),
        yielding $\blocked{\name{r}}\in X$
        in analogy to~\ref{l:results:2}.
      \item If $\nbody{r}\cap X\neq\emptyset$,
        then there is some $L^{-}\in\nbody{r}$
        with $L^{-}\in X$.
        Hence,
        \(
        (\bl{2}{r}{L^{-}})^{+}\in\reduct{\mathcal{T}(\Pi)}{X}
        \),
        yielding $\blocked{\name{r}}\in X$
        in analogy to~\ref{l:results:2}.
      \end{itemize}
      This shows that either $\applied{\name{r}}\in X$ or $\blocked{\name{r}}\in X$.
      In other words, $\applied{\name{r}}\in X$ iff $\blocked{\name{r}}\not\in X$.
    \end{itemize}
  \end{itemize}
\item[Step.]
  Consider $r\in\Pi$.
  Assume that $\ok{\name{r'}}\in X$ 
  and
  either $\applied{\name{r'}}\in X$ or $\blocked{\name{r'}}\in X$
  for all $r'$ with $r<_X r'$.

  \begin{itemize}
  \item [\ref{l:results:2}.]
    In analogy to the base case, we have
    \(
    \rdy{\name{r}}{\name{r''}}\in X
    \)
    for all $r''\in\Pi$ with $r\not <_X r''$.

    Consider $r'$ with $r<_X r'$.
    We thus have $\PREC{\name{r}}{\name{r'}}\in X$.
    By the induction hypothesis, we have either
    $\applied{\name{r'}}\in X$ or $\blocked{\name{r'}}\in X$.
    Hence we have either
    \[
    \body{(\cok{3}{r}{r'})^{+}}\subseteq X
    \text{ or }
    \body{(\cok{4}{r}{r'})^{+}}\subseteq X
    \ \mbox{.}
    \]
    This implies 
    \(
    \rdy{\name{r}}{\name{r'}}\in X
    \)
    for all $r'\in\Pi$ with $r <_X r'$.

    We have thus shown that
    \(
    \rdy{\name{r}}{\name{r'''}}\in X
    \)
    for all $r'''\in\Pi$.

    By the fact that $X$ is closed,
    we get
    \(
    \ok{\name{r}}\in X
    \).    

  \item [\ref{l:results:3}.]
    Analogous to the base case. $\mathproofbox$
  \end{itemize}
\end{description}
\end{itemize}
\end{proof*}


\begin{proof*}[Proof of Proposition~\ref{thm:results:ii}]
Let $X$ be a consistent answer set of
\(
\mathcal{T}(\Pi)
\) 
for an ordered program $\Pi$, let
  \(
  \Omega=
  \reduct{\mathcal{T}(\Pi)}{X}
  \), 
and consider some 
  \(
  r\in \Pi
  \).

\begin{itemize}
\item [\ref{l:results:4}.]
  Given
  $\ok{\name{r}}\in \Tind{i}{\Omega}{\emptyset}$
  and
  $\pbody{r}\subseteq\Tind{j}{\Omega}{\emptyset}$,
  we get
  $\pbody{\ap{2}{r}}\subseteq \Tind{\max(i,j)}{\Omega}{\emptyset}$.
  With
  $\nbody{\ap{2}{r}}\cap X=\emptyset$,
  this implies 
  $\applied{\name{r}}\in\Tind{\max(i,j)+1}{\Omega}{\emptyset}$.


\item [\ref{l:results:5}.]
  Given
$\ok{\name{r}}\in \Tind{i}{\Omega}{\emptyset}$
    and
    $\pbody{r}\not\subseteq X$,
  we get
  $\pbody{\bl{1}{r}{L^{+}}}\subseteq \Tind{i}{\Omega}{\emptyset}$ and
  $\nbody{\bl{1}{r}{L^{+}}}\cap X=\emptyset$,
  for some $L^{+}\in\pbody{r}$.
  Thus,
  $\blocked{\name{r}}\in \Tind{i+1}{\Omega}{\emptyset}$.

\item [\ref{l:results:6}.]
  We have
  $\ok{\name{r}}\in \Tind{i}{\Omega}{\emptyset}$
  and
  $L^{-}\in X$ for some $L^{-}\in\nbody{r}$.
  Assume
  $L^{-}\in\Tind{k}{\Omega}{\emptyset}$
  for some minimal $k$.
  Then, we get
  $\body{\bl{2}{r}{L^{-}}}\subseteq\Tind{\max(i,k)}{\Omega}{\emptyset}$.
  This implies 
  $\blocked{\name{r}}\in \Tind{\max(i,k)+1}{\Omega}{\emptyset}$.
  That is,
  $\blocked{\name{r}}\in \Tind{j}{\Omega}{\emptyset}$ for some $j>i$.

\item [\ref{l:results:7}.]
  Assume 
  $\applied{\name{r}}\in \Tind{j}{\Omega}{\emptyset}$
  for some $j< i+2$.
  Since 
  $\applied{\name{r}}$ 
  can only be derived by means of rule 
  $\ap{2}{r}\in\mathcal{T}(\Pi)$, 
  we obtain that 
  $\ok{\name{r}}\in \Tind{j-1}{\Omega}{\emptyset}$.
  But
  $\Tind{j-1}{\Omega}{\emptyset}\subseteq\Tind{i}{\Omega}{\emptyset}$, 
  so
  $\ok{\name{r}}\in \Tind{i}{\Omega}{\emptyset}$.
  Similar arguments show that
  $\blocked{\name{r}}\in \Tind{k}{\Omega}{\emptyset}$
  for some $k< i+2$ 
  implies
  $\ok{\name{r}}\in \Tind{i}{\Omega}{\emptyset}$. $\mathproofbox$
\end{itemize}
\end{proof*}


\begin{proof}[Proof of Theorem~\ref{thm:order:implementing}]
Let $\Pi$ be an ordered logic program, 
$X$ a consistent answer set of $\mathcal{T}(\Pi)$,
and
\(
\langle r_i\rangle_{i\in I}
\)
a grounded enumeration of \GR{\mathcal{T}(\Pi)}{X}.

Consider $r_1,r_2\in\Pi$  with \PRECMi{X}{\name{r_1}}{\name{r_2}}.
Let $r_i=\cokt{1}{r_1}$ and $r_j=\cokt{1}{r_2}$.
We show that $j<i$.

Given that $r_i=\cokt{1}{r_1}$, we have
\(
\{\rdy{r_1}{r}\mid r\in\Pi\}
\subseteq
\{\head{r_k}\mid k<i\}
\).
In particular, we then have
\(
\rdy{r_1}{r_2}\in\{\head{r_k}\mid k<i\}
\).
Because of \PRECMi{X}{\name{r_1}}{\name{r_2}}, we have
\(
(\PREC{\name{r_1}}{\name{r_2}})\in X
\).
Therefore, $\cok{2}{r_1}{r_2}\not\in\GR{\mathcal{T}(\Pi)}{X}$.
Hence, we must have either
\[
\cok{3}{r_1}{r_2}\in\{r_k\mid k<i\}
\quad\text{ or }\quad
\cok{4}{r_1}{r_2}\in\{r_k\mid k<i\}
\]
and then furthermore that either
\[
\ap{2}{r_2}\in\{r_k\mid k<i\}
\text{ or }
\bl{1}{r_2}{L^{+}}\in\{r_k\mid k<i\}
\text{ or }                              
\bl{2}{r_2}{L^{-}}\in\{r_k\mid k<i\}
\]
for some $L^{+}\in\pbody{r_2}$ or some $L^{-}\in\nbody{r_2}$.
In either case, we must have
\(
\ok{r_2}\in\{\head{r_k}\mid k<i\}
\),
that is,
\(
\cokt{1}{r_2}\in\{r_k\mid k<i\}
\),
and thus $j<i$.
\end{proof}


\begin{proof}[Proof of Theorem~\ref{thm:order:preserving}]
Let $(\Pi,\PRECMo)$ be a statically ordered logic program
over $\mathcal{L}_{\mathcal{A}}$
and
$X$ a set of literals.

\paragraph{Only-if part.}

Let $X$ be a $\PRECMo$-preserving answer set of $\Pi$.

Define
\(
\overline{\Pi}
=
\TA{
(\Pi
\cup
\{(\PREC{\name{r}}{\name{r'}})\LPif{}\mid\PRECM{r}{r'}\})}
\)
and
\(
\overline{X}
=
X\cup\{{\neg(\PREC{\name{r'}}{\name{r}})}\mid\PRECM{r}{r'}\}
\).

We draw on the following propositions in the sequel.
\begin{lemma}
Given the above prerequisites, we have
\begin{enumerate}[(iii).]
\renewcommand{\theenumi}{(\roman{enumi})}
\item $X=\overline{X}\cap\mathcal{L}_{\mathcal{A}}$\ ;
\item $\PRECMo\;=\;\PRECMoi{X}\;=\;\PRECMoi{\overline{X}}$\ ;
\item $\overline{X}$ is a $\PRECMo$-preserving answer set of $\overline{\Pi}$.
\end{enumerate}
\end{lemma}

Hence there is an enumeration
\(
\langle r_i\rangle_{i\in I}
\)
of \GR{\overline{\Pi}}{\overline{X}} satisfying
Conditions~\ref{def:order:preserving:zero} to~\ref{def:order:preserving:two} in
Definition~\ref{def:order:preserving}.
Without loss of generality, assume that there is some $m\in I$ such that
\begin{equation}
  \label{eq:order:preserving:precs}
  \begin{array}{lrcl}
    &
    \{r_i\mid 0\leq i\leq m, i\in I\}
    &=&
    \{r\mid\head{r}\in\mathcal{L}_{\mathcal{A}_\prec}\}
    \\
    \text{ and }&
    \{\head{r_i}\mid i>m, i\in I\}
    \cap
    \mathcal{L}_{\mathcal{A}_\prec}
    &=&
    \emptyset
    \ \mbox{.}
  \end{array}
\end{equation}

From $\langle r_i\rangle_{i\in I}$,
we construct an enumeration
\(
\langle r_j\rangle_{j\in J}
\)
of $\overline{\Pi}$ with $I\subseteq J$ and $\ll_I\subseteq\ll_J$
(where $\ll_I,\ll_J$ are the total orders on
$\GR{\overline{\Pi}}{\overline{X}}$ and ${\overline{\Pi}}$ induced by 
$\langle r_i\rangle_{i\in I}$ and $\langle r_j\rangle_{j\in J}$, respectively)
such that for every $i,j\in J$ we have that:
\begin{itemize}
\item[\ref{def:order:preserving:x:one}.]
  If \PRECM{r_i}{r_j}, then $j<i$.

  This is obtainable by letting $\ll_J$ be a total extension of $\PRECMo$,
  that is, ${\PRECMo}\subseteq{\ll_J}$.
\item[\ref{def:order:preserving:x:tri}.]
  If
  \PRECM{r_i}{r_j},
  then
  there is some $r_k\in\GR{\overline{\Pi}}{\overline{X}}$
  such that
  \begin{enumerate}[2a.]
  \item[\ref{def:order:preserving:x:tri:a}.]
    $k<i$ and
  \item[\ref{def:order:preserving:x:tri:b}.]
    \(
    \head{r_k}= (\PREC{\name{r_i}}{\name{r_j}})
    \).
  \end{enumerate}
  
  This is trivially satisfiable because of~(\ref{eq:order:preserving:precs}).
\item[\ref{def:order:preserving:x:zero}.]
  If
  \(
  r_i\in\GR{\overline{\Pi}}{\overline{X}}
  \),
  then
  \(
  \pbody{r_i}
  \subseteq
  \{\head{r_k}\mid r_k\in\GR{\overline{\Pi}}{\overline{X}}, k<i\}
  \).

  This is a direct consequence of Condition~\ref{def:order:preserving:zero} in
  Definition~\ref{def:order:preserving}.
\item[\ref{def:order:preserving:x:two}.]
  If
  \(
  r_i\in{\overline{\Pi}\setminus\GR{\overline{\Pi}}{\overline{X}}},
  \)
  then
  \begin{enumerate}[4a.]
  \item[\ref{def:order:preserving:x:two:a}.]
    \(
    \pbody{r_i}\not\subseteq X
    \)
    or
  \item[\ref{def:order:preserving:x:two:b}.]
    \(
    \nbody{r_i}
    \cap
    \{\head{r_k}\mid r_k\in\GR{\overline{\Pi}}{\overline{X}}, k<i\}
    \neq
    \emptyset
    \).
  \end{enumerate}
  
  Suppose \PRECM{r_j}{r_i} for some ${r_j}\in\GR{\overline{\Pi}}{\overline{X}}$.
  Then, \ref{def:order:preserving:x:two:a} is a direct consequence of its
  counterpart in Definition~\ref{def:order:preserving}, expressed there as
  Condition~\ref{def:order:preserving:two:a}.

  Now, consider the $\ll_I$-smallest rule
  \(
  r_l\in\GR{\overline{\Pi}}{\overline{X}}
  \)
  with \PRECM{r_l}{r_i}.
  Then, the counterpart of \ref{def:order:preserving:x:two:b},
  Condition~\ref{def:order:preserving:two:b} in
  Definition~\ref{def:order:preserving}, implies that
  \(
  \nbody{r_i}
  \cap
  \{\head{r_k}\mid r_k\in\GR{\overline{\Pi}}{\overline{X}}, k<l\}
  \neq
  \emptyset
  \).
  Let $r_h\in\GR{\overline{\Pi}}{\overline{X}}$ be the predecessor of ${r_l}$
  in $\langle r_i\rangle_{i\in I}$.
  Then, we have
  \(
  \nbody{r_i}
  \cap
  \{\head{r_k}\mid r_k\in\GR{\overline{\Pi}}{\overline{X}}, k\leq h\}
  \neq
  \emptyset
  \).
  Without violating any other constraints,
  we can then position $r_i$ in $\langle r_j\rangle_{i\in J}$
  such that $h<i<l$.
  Consequently, we obtain
  \(
  \nbody{r_i}
  \cap
  \{\head{r_k}\mid r_k\in\GR{\overline{\Pi}}{\overline{X}}, k< i\}
  \neq
  \emptyset
  \).

  Otherwise, that is, whenever \notPRECM{r_j}{r_i} for every
  ${r_j}\in\GR{\overline{\Pi}}{\overline{X}}$,
  rule $r_i$ can be positioned after the $\ll_I$-maximal rule in
  $\langle r_j\rangle_{i\in J}$;
  this satisfies all constraints.
\end{itemize}
Given that $\PRECMo \ = \ \PRECMoi{\overline{X}}$, we get that ${\overline{X}}$ is
a $\PRECMoi{\overline{X}}$-preserving answer set of ${\overline{\Pi}}$.
According to Theorem~\ref{thm:order:preserving:x},
there is then some answer set $Y$ of\/ $\mathcal{T}(\Pi,\PRECMo)$ such
that ${\overline{X}}=Y\cap\mathcal{L}$.
Consequently, we obtain that
\(
{X}
=
{\overline{X}}\cap\mathcal{L}_{\mathcal{A}}
=
(Y\cap\mathcal{L})\cap\mathcal{L}_{\mathcal{A}}
=
Y\cap\mathcal{L}_{\mathcal{A}}
\).

\paragraph{If part.}
Let $Y$ be an answer set of $\mathcal{T}(\Pi,\PRECMo)$,
\iec of
\[
\mathcal{T}(\Pi
            \cup
            \{(\PREC{\name{r}}{\name{r'}})\LPif{}\mid\PRECM{r}{r'}\})\mbox{.}
\]
According to Theorem~\ref{thm:order:preserving:x},
there is then an enumeration
\(
\langle s_k\rangle_{k\in K}
\)
of
\[
\TA{(\Pi\cup\{(\PREC{\name{r}}{\name{r'}})\LPif\mid\PRECM{r}{r'}\})}
\]
satisfying Conditions~\ref{def:order:preserving:x:one},
\ref{def:order:preserving:x:tri},
\ref{def:order:preserving:x:zero}, and
\ref{def:order:preserving:x:two} in Definition~\ref{def:order:preserving:x}.

Define
\(
\langle r_i\rangle_{i\in I}
\)
as the enumeration obtained from $\langle s_k\rangle_{k\in K}$ by
\begin{enumerate}
\item deleting all rules apart from those of form
  \ap{2}{r} for $r\in\Pi$; and
\item replacing each rule of form
  \ap{2}{r} by $r$.
\end{enumerate}
Define $X=\{\head{r_i}\mid i\in I\}$.
Then, we have $X=Y\cap\mathcal{L}_{\mathcal{A}}$ and
$\GR{\Pi}{X}=\{\head{r_i}\mid i\in I\}$.
Hence $\langle r_i\rangle_{i\in I}$ is an enumeration of \GR{\Pi}{X}.
Moreover, $\langle r_i\rangle_{i\in I}$ satisfies
Conditions~\ref{def:order:preserving:zero},
\ref{def:order:preserving:one}, and
\ref{def:order:preserving:two} in Definition~\ref{def:order:preserving}
by virtue of $\langle s_k\rangle_{k\in K}$
satisfying Conditions~\ref{def:order:preserving:x:one},
\ref{def:order:preserving:x:tri},
\ref{def:order:preserving:x:zero}, and
\ref{def:order:preserving:x:two} in Definition~\ref{def:order:preserving:x}.
More specifically,
Condition~\ref{def:order:preserving:zero} in Definition~\ref{def:order:preserving}
is a direct consequence of
Condition~\ref{def:order:preserving:x:zero} in Definition~\ref{def:order:preserving:x}.
As well, Condition~\ref{def:order:preserving:one} in Definition~\ref{def:order:preserving}
follows immediately from
Condition~\ref{def:order:preserving:x:one} in Definition~\ref{def:order:preserving:x},
while
Condition~\ref{def:order:preserving:two} in Definition~\ref{def:order:preserving}
is a consequence of
Conditions~\ref{def:order:preserving:x:two} and \ref{def:order:preserving:x:one} 
in Definition~\ref{def:order:preserving:x}.
Hence, $X$ is a $\PRECMo$-preserving answer set of $\Pi$.
\end{proof}


\begin{proof*}[Proof of Theorem~\ref{thm:order:preserving:x}]
Let $\Pi$ be an ordered logic program over language $\mathcal{L}$.
%
\paragraph{If part.}
Let $Y$ be a consistent answer set of $\mathcal{T}(\Pi)$.
Define
\begin{eqnarray*}
X
& = & \hphantom{\cup\,}
\{\head{r}\mid\applied{\name{r}}\in Y\}
\cup
\{     \PREC{\name{r}}{\name{t}} \mid\PRECMi{Y}{r}{s},\PRECMi{Y}{s}{t}\} \\
&&
\cup \, \{\neg(\PREC{\name{s}}{\name{r}})\mid\PRECMi{Y}{r}{s}\}
\ \mbox{.}
\end{eqnarray*}
Observe that $X=Y\cap\mathcal{L}$.
Furthermore,
we note the following useful relationships for $X$ and $Y$.
\begin{lemma}\label{lem:language:Y:X}
  For any $L\in\Lit$,
  we have that $L\in Y$ iff $L\in X$.
\end{lemma}
Given that $Y$ is consistent, this implies that $X$ is consistent, too.
\begin{lemma}\label{lem:order:Y:X}
  We have that ${\PRECMoi{Y}}={\PRECMoi{X}}$.
\end{lemma}

\paragraph{If part, I.}
First, we show that $X$ is a (standard) answer set of \TA{\Pi}.
That is, $X=\Th{\reduct{(\TA{\Pi})}{X}}$.

Observe that
\(
{\reduct{(\TA{\Pi})}{X}}
=
{\reduct{\Pi}{X}}
\cup
\{{t(r,s,t)},{as(r,s)}\mid r,s,t \in \Pi\}
\).

\paragraph{``$\,\supseteq$'' part.}
We start by showing that $X$ is closed under $\reduct{(\TA{\Pi})}{X}$.

Consider $r^{+}\in\reduct{\Pi}{X}$ with $\pbody{r}\subseteq X$.
We then have $\nbody{r}\cap X=\emptyset$.
Hence, we also have $\nbody{r}\cap Y=\emptyset$
by Lemma~\ref{lem:language:Y:X}.
This implies that
\(
\ap{2}{r}^{+}\in\reduct{\mathcal{T}(\Pi)}{Y}
\).
Note that $\pbody{r}\subseteq X$ implies $\pbody{r}\subseteq Y$.
Also, we have $\ok{\name{r}}\in Y$ by
Condition~\ref{l:results:2} of Proposition~\ref{thm:results:i}.
The fact that $Y$ is closed under \reduct{\mathcal{T}(\Pi)}{Y} implies
\(
\applied{\name{r}}\in Y
\).
We thus get $\head{r}\in X$ by definition of $X$.

Consider ${t(r,s,t)}$ with
\(
\{\PREC{\name{r}}{\name{s}},\PREC{\name{s}}{\name{t}}\}\subseteq X
\).
By Lemma~\ref{lem:order:Y:X}, we get \PRECMi{Y}{r}{s} and \PRECMi{Y}{s}{t}.
We thus get $\head{t(r,s,t)}=\PREC{\name{r}}{\name{t}}\in X$ by definition of $X$.

Consider ${as(r,s)}$ with
\(
\PREC{\name{r}}{\name{s}}\in X
\).
By Lemma~\ref{lem:order:Y:X}, we get \PRECMi{Y}{r}{s}.
We thus get $\head{as(r,s)}=\neg(\PREC{\name{s}}{\name{r}})\in X$ by
definition of $X$.

Note that the closure of $X$ under $\reduct{(\TA{\Pi})}{X}$ shows that
$\Th{\reduct{(\TA{\Pi})}{X}}\subseteq X$.

\paragraph{``$\,\subseteq$'' part.}
We now show that $X$ is the smallest set being closed under $\reduct{(\TA{\Pi})}{X}$,
or equivalently that $X\subseteq\Th{\reduct{(\TA{\Pi})}{X}}$.
We observe that
\[
X
=
Y\cap\mathcal{L}
=
\left(
  \mbox{$\bigcup$}_{i\geq 0}\Tind{i}{\reduct{\mathcal{T}(\Pi)}{Y}}{\emptyset}
\right)
\cap\mathcal{L}
\ \mbox{.}
\]
We then show by induction that
\(
(\Tind{i}{\reduct{\mathcal{T}(\Pi)}{Y}}{\emptyset})\cap\mathcal{L}
\subseteq
\Th{\reduct{(\TA{\Pi})}{X}}
\)
for $i\geq 0$.
\begin{description}
\item[Base.] Trivial since
  \(
  \Tind{0}{\reduct{\mathcal{T}(\Pi)}{Y}}{\emptyset}={\emptyset}
  \). 
\item[Step.]
  Assume
  \(
  (\Tind{j}{\reduct{\mathcal{T}(\Pi)}{Y}}{\emptyset})\cap\mathcal{L}
  \subseteq
  \Th{\reduct{(\TA{\Pi})}{X}}
  \)
  for $0\leq j\leq i$.
  \begin{itemize}
  \item 
    If
    \(
    (\Tind{i+1}{\reduct{\mathcal{T}(\Pi)}{Y}}{\emptyset})\cap\mathcal{L}
    =
    \emptyset
    \),
    our claim obviously holds.
  \item
    If there is some $r\in\Pi$ such that
    \(
    \head{r}
    \in
    (\Tind{i+1}{\reduct{\mathcal{T}(\Pi)}{Y}}{\emptyset})\cap\mathcal{L}
    \),
    that is,
    \(
    \head{r}=\head{\reductr{\ap{1}{r}}}
    \),
    we have
    \[
    \left\{
      \begin{array}[c]{rrcl}
        \RULE{\reductr{\ap{1}{r}}}{\head{r}}{\applied{\name{r}} }
        \\
        \RULE{\reductr{\ap{2}{r}}}{\applied{\name{r}}}{\ok{\name{r}},\pbody{r}}
      \end{array}
    \right\}
    \subseteq
    \reduct{\mathcal{T}(\Pi)}{Y}
    \ \mbox{.}
    \]
    Consequently, we must have
    \(
    \{\applied{\name{r}},\ok{\name{r}}\}\cup\pbody{r}
    \subseteq
    \Tind{i}{\reduct{\mathcal{T}(\Pi)}{Y}}{\emptyset}
    \).

    According to the induction hypothesis,
    \(
    \pbody{r}\subseteq
    (\Tind{i}{\reduct{\mathcal{T}(\Pi)}{Y}}{\emptyset})\cap\mathcal{L}
    \)
    implies
    \(
    \pbody{r}\subseteq\Th{\reduct{(\TA{\Pi})}{X}}
    \).
    
    Also, $\reductr{\ap{2}{r}}\in\reduct{\mathcal{T}(\Pi)}{Y}$ implies that
    \(
    \nbody{r}\cap Y=\emptyset
    \).
    By Lemma~\ref{lem:language:Y:X}, we thus get
    \(
    \nbody{r}\cap X=\emptyset
    \),
    which implies
    \(
    \reductr{r}\in{\reduct{\Pi}{X}}
    \).
    
    Given that $\Th{\reduct{(\TA{\Pi})}{X}}$ is closed under \reduct{\Pi}{X},
    we get that
    \(
    \head{r}\in\Th{\reduct{(\TA{\Pi})}{X}}
    \).
  \item 
    If
    \(
    \head{t(r,s,t)}
    \in
    (\Tind{i+1}{\reduct{\mathcal{T}(\Pi)}{Y}}{\emptyset})\cap\mathcal{L}
    \),
    then we must have
    \(
    \{\PREC{\name{r}}{\name{s}},\PREC{\name{s}}{\name{t}}\}
    \subseteq
    \Tind{i}{\reduct{\mathcal{T}(\Pi)}{Y}}{\emptyset}
    \).
    And by the induction hypothesis, we further get
    \(
    \{\PREC{\name{r}}{\name{s}},\PREC{\name{s}}{\name{t}}\}
    \subseteq
    \Th{\reduct{(\TA{\Pi})}{X}}
    \).
    Given that $\Th{\reduct{(\TA{\Pi})}{X}}$ is closed under
    \(
    {t(r,s,t)}\in\reduct{(\TA{\Pi})}{X}
    \),
    we thus obtain that
    \(
    \head{t(r,s,t)}\in\Th{\reduct{(\TA{\Pi})}{X}}
    \).
  \item
    If
    \(
    \head{as(r,s)}
    \in
    (\Tind{i+1}{\reduct{\mathcal{T}(\Pi)}{Y}}{\emptyset})\cap\mathcal{L}
    \),
    then we must have
    \(
    \PREC{\name{r}}{\name{s}}\in\Tind{i}{\reduct{\mathcal{T}(\Pi)}{Y}}{\emptyset}
    \).
    And by the induction hypothesis, we further get
    \(
    \PREC{\name{r}}{\name{s}}\in\Th{\reduct{(\TA{\Pi})}{X}}
    \).
    Given that $\Th{\reduct{(\TA{\Pi})}{X}}$ is closed under
    \(
    {as(r,s)}\in\reduct{(\TA{\Pi})}{X}
    \),
    we thus obtain that
    \(
    \head{as(r,s)}\in\Th{\reduct{(\TA{\Pi})}{X}}
    \).
  \end{itemize}
\end{description}
In all, we have now shown that
\(
X
=
\Th{\reduct{(\TA{\Pi})}{X}}
\).
That is, $X$ is a (standard) answer set of \TA{\Pi}.

\paragraph{If part, II.}
Next, we show that $X$ is a \PRECMoi{X}-preserving answer set of \TA{\Pi}.
Since $Y$ is a (standard) answer set of ${\mathcal{T}(\Pi)}$,
according to Lemma~\ref{lem:two},
there is a grounded enumeration
\(
\langle s_k\rangle_{k\in K}
\)
of $\GR{\mathcal{T}(\Pi)}{Y}$.

Define
\(
\langle r_i\rangle_{i\in I}
\)
as the enumeration obtained from $\langle s_k\rangle_{k\in K}$ by
\begin{enumerate}
\item deleting all rules apart from those of form
  \ap{2}{r},
  \bl{1}{r}{L^{+}},
  \bl{2}{r}{L^{-}};
\item replacing each rule of form
  \ap{2}{r},
  \bl{1}{r}{L^{+}},
  \bl{2}{r}{L^{-}}
  by $r$; and
\item removing duplicates\footnote{Duplicates can only occur if a rule is blocked
    in multiple ways.} by increasing $i$
\end{enumerate}
for $r\in\TA{\Pi}$ and $L^{+}\in\pbody{r}$, $L^{-}\in\nbody{r}$.

First of all, we note that $\langle r_i\rangle_{i\in I}$ satisfies
Condition~\ref{def:order:preserving:x:zero}.
This is a direct consequence of the fact that $\langle s_k\rangle_{k\in K}$
enjoys groundedness.

For establishing Condition~\ref{def:order:preserving:x:two},
consider $r_i\not\in\GR{\TA{\Pi}}{X}$.
If we have
\(
\pbody{r_i}\not\subseteq X
\),
then this establishes Condition~\ref{def:order:preserving:x:two:a} 
and we are done.
Otherwise,
if we have
\(
\nbody{r}\cap X\neq\emptyset
\),
then there is some $L^{-}\in\nbody{r}$ such that $L^{-}\in X$.
This implies $L^{-}\in Y$ by Lemma~\ref{lem:language:Y:X}.
Given this and that $\ok{r}\in Y$ by Condition~\ref{l:results:2} 
of Proposition~\ref{thm:results:i},
we obtain $\bl{2}{r}{L^{-}}\in\GR{\mathcal{T}(\Pi)}{Y}$.
More precisely, let $\bl{2}{r}{L^{-}}=s_k$ for some $k\in K$.
Then, we have that
\(
L^{-}\in\{\head{s_l}\mid l\in K, l<k\}
\).
Given the way 
$\langle r_i \rangle_{i\in I}$
is obtained from 
$\langle s_k\rangle_{k\in K}$,
we thus obtain that
\(
L^{-}\in\{\head{r_l}\mid r_l\in\GR{\TA{\Pi}}{X}, l<i\}
\).
That is, we obtain Condition~\ref{def:order:preserving:x:two:b}.

Next, we show that $\langle r_i\rangle_{i\in I}$ satisfies
Condition~\ref{def:order:preserving:x:one}.
Consider $r_i,r_j\in\Pi$ for some $i,j\in I$.
Then, there are $k_i,k_j\in K$ such that
\[
\begin{array}{lll}
              & s_{k_i}=\ap{2}{r_i}      &                                     \\
  \text{ or } & s_{k_i}=\bl{1}{r_i}{L^{+}} & \text{ for some } L^{+}\in\pbody{r_i} \\
  \text{ or } & s_{k_i}=\bl{2}{r_i}{L^{-}} & \text{ for some } L^{-}\in\nbody{r_i}\ ,
  \qquad\qquad\text{ and}
  \\
              & s_{k_j}=\ap{2}{r_j}      &                                     \\
  \text{ or } & s_{k_j}=\bl{1}{r_j}{L^{+}} & \text{ for some } L^{+}\in\pbody{r_j} \\
  \text{ or } & s_{k_j}=\bl{2}{r_j}{L^{-}} & \text{ for some } L^{-}\in\nbody{r_j}\ .
  \qquad\qquad\phantom{\text{ and}}
\end{array}
\]
Assume \PRECMi{X}{r_i}{r_j}.
Then, we also have \PRECMi{Y}{r_i}{r_j} by Lemma~\ref{lem:order:Y:X}.
Given this, by the properties of transformation $\mathcal T$, 
it is easy to see that 
$k_j<k_i$ must hold.
Given the way 
$\langle r_i \rangle_{i\in I}$
is obtained from 
$\langle s_k\rangle_{k\in K}$,
the relation
$k_j<k_i$
implies that
\(
j<i
\).
This establishes Condition~\ref{def:order:preserving:x:one}.

For addressing  Condition~\ref{def:order:preserving:x:tri},
assume \PRECMi{X}{r_i}{r_j} for $i,j\in I$.
By definition,
\PRECMi{X}{r_i}{r_j} implies ${(\PREC{\name{r_i}}{\name{r_j}})}\in X$.
That is, there is some smallest $h\in I$ such that
\(
r_h\in\GR{\TA{\Pi}}{X}
\)
and
\(
\head{r_h}={(\PREC{\name{r_i}}{\name{r_j}})}
\).
Furthermore, we obtain
${(\PREC{\name{r_i}}{\name{r_j}})}\in Y$
by Lemma~\ref{lem:order:Y:X}.
We thus have
\(
\reductr{\cok{2}{r_i}{r_j}}\not\in\reduct{\mathcal{T}(\Pi)}{Y}
\),
while we have
\[
\left\{
  \begin{array}[c]{rrcl}
  \RULE{\reductr{\cok{3}{r_i}{r_j}}}
       {\rdy{\name{r_i}}{\name{r_j}}}
       {(\PREC{\name{r_i}}{\name{r_j}}),\applied{\name{r_j}}}
  \\
  \RULE{\reductr{\cok{4}{r_i}{r_j}}}
       {\rdy{\name{r_i}}{\name{r_j}}}
       {(\PREC{\name{r_i}}{\name{r_j}}),\blocked{\name{r_j}}}
  \end{array}
\right\}
\subseteq
\reduct{\mathcal{T}(\Pi)}{Y}
\ \mbox{.}
\]
Since ${\rdy{\name{r_i}}{\name{r_j}}}\subseteq Y$
(by Condition~\ref{l:results:2} in Proposition~\ref{thm:results:i}),
we have either ${\cok{3}{r_i}{r_j}}\in\GR{{\mathcal{T}(\Pi)}}{Y}$ or
${\cok{4}{r_i}{r_j}}\in\GR{{\mathcal{T}(\Pi)}}{Y}$.
Analogously, we have ${\cokt{1}{r_i}}\in\GR{{\mathcal{T}(\Pi)}}{Y}$.

Now, consider the enumeration $\langle s_k\rangle_{k\in K}$ of
\GR{{\mathcal{T}(\Pi)}}{Y}, and, in particular,
rules $s_{k_i}$ (as defined above) and $s_{k_h}=\ap{2}{r_h}$ for some ${k_h}\in K$.
Since $\langle s_k\rangle_{k\in K}$ is grounded,
the minimality of ${k_h}$ (due to that of $h$) implies that there are some
\(
l_1,l_2,l_3>0
\)
such that 
\begin{itemize}
\item either
  \(
  s_{{k_h}+l_1}={\cok{3}{r_i}{r_j}}
  \)
  or
  \(
  s_{{k_h}+l_1}={\cok{4}{r_i}{r_j}}
  \),
\item
  \(
  s_{{k_h}+l_1+l_2}={\cokt{1}{r_i}}
  \),
  and
\item
  \(
  s_{{k_h}+l_1+l_2+l_3}=s_{k_i}
  \).
\end{itemize}
Consequently, we have ${k_h}<{k_i}$.
The construction of $\langle r_i \rangle_{i\in I}$ from $\langle
s_k\rangle_{k\in K}$ implies that $h<i$,
which establishes Condition~\ref{def:order:preserving:x:tri}.

\paragraph{Only-if part.}
Let $X$ be a consistent \PRECMoi{X}-preserving answer set of \TA{\Pi}.

Define
\begin{eqnarray*}
  Y &=&\quad \{\head{r} \mid r\in\GR{\TA{\Pi}}{X}\}
  \\& &\cup\;\{\applied{\name{r}}\mid r    \in\GR{\TA{\Pi}}{X}\}
       \cup  \{\blocked{\name{r}}\mid r\not\in\GR{\TA{\Pi}}{X}\}
  \\& &\cup\;\{\ok{\name{r}}   \mid r\in\TA{\Pi}\}
       \cup  \{\rdy{\name{r}}{\name{r'}}\mid r,r'\in\TA{\Pi}\} \ \mbox{.}
\end{eqnarray*}
We note the following useful relationship for $X$ and $Y$.
\begin{lemma}\label{lem:language:X:Y}
  For any $L\in\Lit$,
  we have that $L\in X$ iff $L\in Y$.
\end{lemma}
Given that $X$ is consistent, this implies that $Y$ is consistent, too.
\begin{lemma}\label{lem:order:X:Y}
  We have that ${\PRECMoi{Y}}={\PRECMoi{X}}$.
\end{lemma}

We must show that $Y$ is an answer set of ${\mathcal{T}(\Pi)}$,
that is,
\(
Y=\Th{\reduct{\mathcal{T}(\Pi)}{Y}}
\).

\paragraph{``$\,\supseteq$'' part.}
We start by showing that $Y$ is closed under \reduct{\mathcal{T}(\Pi)}{Y}.
For this,
we show for $r\in\mathcal{T}(\Pi)$
that $\head{\reductr{r}}\in Y$
whenever
$\reductr{r}\in{\mathcal{T}(\Pi)^Y}$ with $\pbody{r}\subseteq Y$.

\begin{enumerate}
  \newcommand{\RULEtem}[3]{%
  \item Consider
    \(
    #1:{#2\LPif#3}\in\mathcal{T}(\Pi)
    \).}
  \newcommand{\mRULEtem}[3]{\RULEtem{#1}{#2}{#3}
    We have
    \(
    (#1)^{+}={#1}
    \).
    Clearly,
    \(
    {#1}\in\reduct{\mathcal{T}(\Pi)}{Y}
    \).}
  \newcommand{\nmRULEtem}[5]{\RULEtem{#1}{#2}{#3}
    Suppose
    \(
    (#1)^{+}={#2\LPif#4}\in\reduct{\mathcal{T}(\Pi)}{Y}
    \)
    with
    \(
    {#5}\cap Y=\emptyset
    \).}
  \mRULEtem{\ap{1}{r}}%
           {\head{r}}%
           {\applied{\name{r}}}
  By definition of $Y$,    
  we have $\head{r}\in Y$ whenever $\applied{\name{r}}\in Y$.

  \nmRULEtem{\ap{2}{r}}%
            {\applied{\name{r}}}%
            {\ok{\name{r}},\body{r}}%
            {\ok{\name{r}},\pbody{r}}%
            {\nbody{r}}
  Assume
  \(
  \{\ok{\name{r}}\}\cup\pbody{r}\subseteq Y
  \).
  From Lemma~\ref{lem:language:X:Y}, we get
  \(
  \pbody{r}\subseteq X
  \).
  Accordingly, ${\nbody{r}}\cap Y=\emptyset$ implies
  $\nbody{r}\cap X=\emptyset$.
  Hence, we have $r\in\GR{\TA{\Pi}}{X}$, which implies ${\applied{\name{r}}}\in Y$ by
  definition of $Y$.
  
  \item Consider
    \(
    \bl{1}{r}{L^{+}}: \blocked{\name{r}} \LPif\ok{\name{r}},\naf L^{+}\in\mathcal{T}(\Pi)
    \),
and assume 
    \(
   (\bl{1}{r}{L^{+}})^{+}={\blocked{\name{r}}
    \LPif\ok{\name{r}}}\in\reduct{\mathcal{T}(\Pi)}{Y}
    \)
    with
    \(
    {L^{+}}\cap Y=\emptyset
    \).
  The latter and Lemma~\ref{lem:language:X:Y} imply that
  \(
  L^{+}\not\in X
  \)
  for $L^{+}\in\pbody{r}$.
  Hence, $r\not\in\GR{\TA{\Pi}}{X}$, implying that ${\blocked{\name{r}}}\in Y$ by
  definition of $Y$.
  
  \mRULEtem{\bl{2}{r}{L^{-}}}%
           {\blocked{\name{r}}}%
           {\ok{\name{r}},L^{-}}
  Suppose
  \(
  \{\ok{\name{r}},L^{-}\}\subseteq Y
  \).
  Lemma~\ref{lem:language:X:Y} implies that
  \(
  L^{-}\in X
  \)
  for $L^{-}\in\nbody{r}$.
  Hence, $r\not\in\GR{\TA{\Pi}}{X}$, implying that ${\blocked{\name{r}}}\in Y$ by
  definition of $Y$.
  
  \mRULEtem{\cokt{1}{r}}%
           {\ok{\name{r}}}%
           {\rdy{\name{r}}{\name{r_1}},\dots,\rdy{\name{r}}{\name{r_k}}}
  We trivially have ${\ok{\name{r}}}\in Y$ by definition of $Y$.

  \item Consider 
    \(
    \{{\cok{2}{r}{r'}}, {\cok{3}{r}{r'}}, {\cok{4}{r}{r'}}\}
    \subseteq
    {\mathcal{T}(\Pi)}
    \).
    Clearly,
    \(
    \head{\cok{i}{r}{r'}}={\rdy{\name{r}}{\name{r'}}}
    \)
    for $i=2,3,4$.
    We trivially have ${\rdy{\name{r}}{\name{r'}}}\in Y$ by definition of~$Y$.
  
  \mRULEtem{t(r,r',r'')}%
           {\PREC{\name{r}}{\name{r''}}}%
           {\PREC{\name{r}}{\name{r'}},\PREC{\name{r'}}{\name{r''}}}
  Assume  $\{{\PREC{\name{r}}{\name{r'}}}$, ${\PREC{\name{r'}}{\name{r''}}}\}\subseteq Y$.
  From Lemma~\ref{lem:language:X:Y}, we get
  \(
  \{{\PREC{\name{r}}{\name{r'}},\PREC{\name{r'}}{\name{r''}}}\}\subseteq X
  \).
  That is,
  \(
  \pbody{t(r,r',r'')}\subseteq X
  \)
  for ${t(r,r',r'')}\in\TA{\Pi}$.
  In analogy to what we have shown above in~2 and~1, we then obtain
  \(
  {\applied{\name{t(r,r',r'')}}}\in Y
  \)
  and
  \(
  {\head{t(r,r',r'')}}\in Y
  \).
  We thus get ${\PREC{\name{r}}{\name{r''}}\in Y}$.
  
  \mRULEtem{as(r,r')}%
           {{\neg(\PREC{\name{r'}}{\name{r}})}}%
           {\PREC{\name{r}}{\name{r'}}}
  Assume $({\PREC{\name{r}}{\name{r'}}})\in Y$.
  Then, in analogy to 7, we get $\neg({\PREC{\name{r'}}{\name{r}})\in Y}$. 
  
\end{enumerate}
Note that the closure of $Y$ under \reduct{\mathcal{T}(\Pi)}{Y} shows that
$\Th{\reduct{\mathcal{T}(\Pi)}{Y}}\subseteq Y$.

\paragraph{``$\,\subseteq$'' part.}
We must now show that $Y$ is the smallest set being closed under
\reduct{\mathcal{T}(\Pi)}{Y},
or equivalently that $Y\subseteq\Th{\reduct{\mathcal{T}(\Pi)}{Y}}$.

Since $X$ is a \PRECMoi{X}-preserving answer set of $\TA{\Pi}$,
there is an enumeration
\(
\langle r_i\rangle_{i\in I}
\)
of \TA{\Pi} satisfying all conditions given in
Definition~\ref{def:order:preserving:x}.
We proceed by induction on $\langle r_i\rangle_{i\in I}$ and show that
\[
\begin{array}{clcl}
       & \{\head{r_i},\applied{\name{r_i}}\mid r_i    \in\GR{\TA{\Pi}}{X},i\in I\} &&
\\\cup & \{\blocked{\name{r_i}}           \mid r_i\not\in\GR{\TA{\Pi}}{X},i\in I\} &&
\\\cup & \{\ok{\name{r_i}}   \mid i\in I\}                          
  \cup  \{\rdy{\name{r_i}}{\name{r_j}}\mid i,j\in I\} &
\subseteq&\Th{\reduct{\mathcal{T}(\Pi)}{Y}}
\ \mbox{.}
\end{array}
\]
\begin{description}
\item[Base.]\label{BASECASE}
  Consider $r_0\in\TA{\Pi}$.
  Given that $X$ is consistent,
  we have
  \(
  (r_0,r)\not\in{\PRECMoi{X}}
  \)
  for all $r\in\TA{\Pi}$
  by Condition~\ref{def:order:preserving:x:one}.
  By Lemma~\ref{lem:order:X:Y}, we thus have
  \(
  (\PREC{\name{r_0}}{\name{r}})\not\in Y
  \)
  for all $r\in\TA{\Pi}$.
  Consequently,
  \[
  \reductr{\cok{2}{r_0}{r}}
  =
  {\rdy{\name{r_0}}{\name{r}}}\LPif\
  \in\reduct{\mathcal{T}(\Pi)}{Y}
  \text{ for all }
  r\in\TA{\Pi}\mbox{.}
  \]
  Since \Th{\reduct{\mathcal{T}(\Pi)}{Y}} is closed under
  {\reduct{\mathcal{T}(\Pi)}{Y}},
  we get
  \(
  {\rdy{\name{r_0}}{\name{r}}}\in\Th{\reduct{\mathcal{T}(\Pi)}{Y}}
  \)
  for all $r\in\TA{\Pi}$.
  Given that $\TA{\Pi} = \{r_1,\dots,r_k\}$ and that
\[
     {\cokt{1}{r_0}}  =  \reductr{\cokt{1}{r_0}} 
                     =  {\ok{\name{r_0}}} \
                          \LPif \ {\rdy{\name{r_0}}{\name{r_1}},\dots,
                          \rdy{\name{r_0}}{\name{r_k}}}
                          \in\reduct{\mathcal{T}(\Pi)}{Y},
\]
  the closure of \Th{\reduct{\mathcal{T}(\Pi)}{Y}} under
  {\reduct{\mathcal{T}(\Pi)}{Y}} implies furthermore that
  ${\ok{\name{r_0}}}\in\Th{\reduct{\mathcal{T}(\Pi)}{Y}}$.

  We distinguish two cases.
  \begin{itemize}
  \item
    Suppose $r_0\in\GR{\TA{\Pi}}{X}$. 
    Since $\langle r_i\rangle_{i\in I}$ satisfies
    Condition~\ref{def:order:preserving:x:zero}, we have
    \(
    \pbody{r_0}=\emptyset
    \).
    Also, $r_0\in\GR{\TA{\Pi}}{X}$ implies that $\nbody{r_0}\cap X=\emptyset$,
    from which we obtain by Lemma~\ref{lem:language:X:Y} that
    \(
    \nbody{r_0}\cap Y=\emptyset
    \).
    We thus obtain that
    \begin{equation}
      \label{eq:ap:ii}
                \ap{2}{r_0}
      =\reductr{\ap{2}{r_0}}
      =
      \applied{\name{r_0}}\LPif\ok{\name{r_0}}
      \in\reduct{\mathcal{T}(\Pi)}{Y}
      \ \mbox{.}
    \end{equation}
    We have shown above that
    ${\ok{\name{r_0}}}\in\Th{\reduct{\mathcal{T}(\Pi)}{Y}}$.
    Accordingly, 
    since \Th{\reduct{\mathcal{T}(\Pi)}{Y}} is closed under
    {\reduct{\mathcal{T}(\Pi)}{Y}},
    we obtain
    \(
    {\applied{\name{r_0}}}\in\Th{\reduct{\mathcal{T}(\Pi)}{Y}}
    \)
    because of~(\ref{eq:ap:ii}).

    This, the closure of \Th{\reduct{\mathcal{T}(\Pi)}{Y}} under
    {\reduct{\mathcal{T}(\Pi)}{Y}}, and the fact that
    \[
              {\ap{1}{r_0}}
    = \reductr{\ap{1}{r_0}}
    = {\head{r_0}}\LPif{\applied{\name{r_0}}}
    \in{\reduct{\mathcal{T}(\Pi)}{Y}}
    \]
    imply furthermore that
    \(
    {\head{r_0}}\in\Th{\reduct{\mathcal{T}(\Pi)}{Y}}
    \).
  \item
    Otherwise, we have $r_0\in\TA{\Pi}\setminus\GR{\TA{\Pi}}{X}$.
    Because $r_0$ cannot satisfy Condition~\ref{def:order:preserving:x:two:b},
    since
    \(
    \nbody{r}\cap\{\head{r_j}\mid j<0\}=\emptyset
    \),
    we must have $\pbody{r_0}\not\subseteq X$.
    Then, there is some
    \(
    L^{+}\in\pbody{r_0}
    \)
    with $L^{+}\not\in X$.
    By Lemma~\ref{lem:language:X:Y}, 
    we also have $L^{+}\not\in Y$.
    Therefore,
    \begin{equation}
      \label{eq:bl:base}
                  \bl{1}{r_0}{L^{+}}
        =\reductr{\bl{1}{r_0}{L^{+}}}
        =
        \blocked{\name{r_0}}\LPif\ok{\name{r_0}}
        \in\reduct{\mathcal{T}(\Pi)}{Y}
        \ \mbox{.}
    \end{equation}
    We have shown above that
    $\ok{\name{r_0}}\in\Th{\reduct{\mathcal{T}(\Pi)}{Y}}$.
    Given this along with~(\ref{eq:bl:base}) and the fact that
    \Th{\reduct{\mathcal{T}(\Pi)}{Y}} is closed under
    ${\reduct{\mathcal{T}(\Pi)}{Y}}$,
    we obtain
    \(
    \blocked{\name{r_0}}\in\Th{\reduct{\mathcal{T}(\Pi)}{Y}}
    \).      
  \end{itemize}

\item[Step.] 
  Consider $r_i\in\TA{\Pi}$ and assume that our claim holds for all
  $r_j\in\TA{\Pi}$ with $j<i$.
  
  We start by providing the following lemma.
  \begin{lemma}\label{lem:preserving:ok}
    Given the induction hypothesis, we have
    \begin{enumerate}[(iii).]
    \renewcommand{\theenumi}{(\roman{enumi})}
    \item 
      \(
      \ok{\name{r_i}}\in\Th{\mathcal{T}(\Pi)^Y},
      \)
      and
    \item 
      \(
      \{\rdy{\name{r_i}}{\name{r_j}}\mid j\in I\}\subseteq\Th{\mathcal{T}(\Pi)^Y} 
      \).
    \end{enumerate}
  \end{lemma}
  \begin{proof}[Proof of Lemma~\ref{lem:preserving:ok}]
    We first prove that
    $\rdy{\name{r_i}}{\name{r_j}}\in\Th{\reduct{\mathcal{T}(\Pi)}{Y}}$
    for all $j\in I$.
    
    Suppose \PRECMi{X}{r_i}{r_j}.
    Since $\langle r_i\rangle_{i\in I}$ satisfies
    Condition~\ref{def:order:preserving:x:one},
    we have that $j<i$.
    Then, the induction hypothesis implies that
    either
    \(
    \applied{\name{r_j}}\in\Th{\reduct{\mathcal{T}(\Pi)}{Y}}
    \)
    or
    \(
    \blocked{\name{r_j}}\in\Th{\reduct{\mathcal{T}(\Pi)}{Y}}
    \).
    According to Condition~\ref{def:order:preserving:x:tri},
    there is some $r_k\in\GR{\TA{\Pi}}{X}$ with $k\in I$
    such that
    \(
    \head{r_k}= (\PREC{\name{r_i}}{\name{r_j}})
    \) 
    and $k<i$.
    By the induction hypothesis, we get that
    \(
    {\head{r_k}}\in\Th{\reduct{\mathcal{T}(\Pi)}{Y}}
    \).
    Hence,
    we obtain
    \(
    (\PREC{\name{r_i}}{\name{r_j}})\in\Th{\reduct{\mathcal{T}(\Pi)}{Y}}
    \).
    Clearly, we have that
    \[
    \left\{
      \begin{array}[c]{rrcl}
        \RULE{\reductr{\cok{3}{r_i}{r_j}}}
        {\rdy{\name{r_i}}{\name{r_j}}}
        {(\PREC{\name{r_i}}{\name{r_j}}),\applied{\name{r_j}}}
        \\
        \RULE{\reductr{\cok{4}{r_i}{r_j}}}
        {\rdy{\name{r_i}}{\name{r_j}}}
        {(\PREC{\name{r_i}}{\name{r_j}}),\blocked{\name{r_j}}}
      \end{array}
    \right\}
    \subseteq
    \reduct{\mathcal{T}(\Pi)}{Y}
    \ \mbox{.}
    \]
    Since \Th{\reduct{\mathcal{T}(\Pi)}{Y}} is closed under
    {\reduct{\mathcal{T}(\Pi)}{Y}},
    we obtain
    \(
    \rdy{\name{r_i}}{\name{r_j}}\in\Th{\reduct{\mathcal{T}(\Pi)}{Y}}
    \).
    Therefore, we have shown that
    \(
    \rdy{\name{r_i}}{\name{r_j}}\in\Th{\reduct{\mathcal{T}(\Pi)}{Y}}
    \)
    whenever \PRECMi{X}{r_i}{r_j}.
    
    Contrariwise, 
    suppose ${(r_i,r_j)}\not\in{\PRECMoi{X}}$.
    We get by Lemma~\ref{lem:language:X:Y} that
    \(
    (\PREC{\name{r_i}}{\name{r_j}})\not\in Y
    \).
    This implies that 
    \(
    \reductr{\cok{2}{r_i}{r_j}}
    =
    {\rdy{\name{r_i}}{\name{r_j}}}\LPif\
    \in{\reduct{\mathcal{T}(\Pi)}{Y}}
    \).
    Since \Th{\reduct{\mathcal{T}(\Pi)}{Y}} is closed under
    {\reduct{\mathcal{T}(\Pi)}{Y}},
    we obtain
    \(
    \rdy{\name{r_i}}{\name{r_j}}\in\Th{\reduct{\mathcal{T}(\Pi)}{Y}}
    \).
    
    Hence, we have
    \(
    \rdy{\name{r_i}}{\name{r_j}}\in\Th{\reduct{\mathcal{T}(\Pi)}{Y}}
    \)
    for all $j\in I$.

    Since \Th{\reduct{\mathcal{T}(\Pi)}{Y}} is closed under
    \(
    \reductr{\cokt{1}{r_i}}\in{\reduct{\mathcal{T}(\Pi)}{Y}}
    \),
    we obtain
    \(
    \ok{\name{r_i}}\in\Th{\reduct{\mathcal{T}(\Pi)}{Y}}
    \).
  \end{proof}
  
  We now distinguish the following two cases.
  \begin{itemize}
  \item
    If ${r_i}\in\GR{\TA{\Pi}}{X}$, then
    \(
    \pbody{r_i}\subseteq\{\head{r_j}\mid r_j\in\GR{\TA{\Pi}}{X}, j<i\}
    \)
    according to Condition~\ref{def:order:preserving:x:zero}.
    By the induction hypothesis,
    we hence obtain
    \(
    \pbody{r_i}\subseteq\Th{\reduct{\mathcal{T}(\Pi)}{Y}}
    \).
    
    By Lemma~\ref{lem:preserving:ok}, 
    we have $\ok{\name{r_i}}\in\Th{\reduct{\mathcal{T}(\Pi)}{Y}}$.

    Also, ${r_i}\in\GR{\TA{\Pi}}{X}$ implies $\nbody{r_i}\cap X=\emptyset$.
    By Lemma~\ref{lem:language:X:Y}, 
    we thus have $\nbody{r_i}\cap Y=\emptyset$.
    This implies
    \begin{equation}
      \label{eq:ap:ii:ii}
                  \ap{2}{r_i}
        =\reductr{\ap{2}{r_i}}
        =
        \applied{\name{r_i}}\LPif\ok{\name{r_i}},\pbody{r_i}
        \in\reduct{\mathcal{T}(\Pi)}{Y}\mbox{.}
    \end{equation}
    As shown above, we have
    $\body{\reductr{\ap{2}{r_i}}}\subseteq\Th{\reduct{\mathcal{T}(\Pi)}{Y}}$.
    Given~(\ref{eq:ap:ii:ii}) and the fact that \Th{\reduct{\mathcal{T}(\Pi)}{Y}} is
    closed under ${\reduct{\mathcal{T}(\Pi)}{Y}}$,
    we therefore get
    \(
    \applied{\name{r_i}}\in\Th{\reduct{\mathcal{T}(\Pi)}{Y}}
    \).
    Accordingly, we obtain
    \(
    {\head{r_i}}\in\Th{\reduct{\mathcal{T}(\Pi)}{Y}}
    \)
    because of $\reductr{\ap{1}{r_i}}\in\reduct{\mathcal{T}(\Pi)}{Y}$.
    
  \item
    Otherwise, we have ${r_i}\in\TA{\Pi}\setminus\GR{\TA{\Pi}}{X}$.
    According to Condition~\ref{def:order:preserving:x:two}, we may distinguish
    the following two cases.
    \begin{itemize}
    \item If $\pbody{r_i}\not\subseteq X$,
      then there is some
      \(
      L^{+}\in\pbody{r_i}
      \)
      with $L^{+}\not\in X$.
      By Lemma~\ref{lem:language:X:Y}, 
      we also have $L^{+}\not\in Y$.
      Therefore,
      \begin{equation}
        \label{eq:bl:ii:ii}
                    \bl{1}{r_i}{L^{+}}
          =\reductr{\bl{1}{r_i}{L^{+}}}
          =
          \blocked{\name{r_i}}\LPif\ok{\name{r_i}}
          \in\reduct{\mathcal{T}(\Pi)}{Y}
          \ \mbox{.}
      \end{equation}
      By Lemma~\ref{lem:preserving:ok},
      we have $\ok{\name{r_i}}\in\Th{\reduct{\mathcal{T}(\Pi)}{Y}}$.
      Given this along with~(\ref{eq:bl:ii:ii}) and the fact that
      \Th{\reduct{\mathcal{T}(\Pi)}{Y}} is closed under
      ${\reduct{\mathcal{T}(\Pi)}{Y}}$,
      we obtain
      \(
      \blocked{\name{r_i}}\in\Th{\reduct{\mathcal{T}(\Pi)}{Y}}
      \).
      
    \item
      If
      \(
      \nbody{r}\cap\{\head{r_j}\mid r_j\in\GR{\TA{\Pi}}{X}, j<i\}\neq\emptyset
      \),
      then there is some
      \(
      L^{-}\in\nbody{r_i}
      \)
      with
      \(
      L^{-}\in\{\head{r_j}\mid r_j\in\GR{\TA{\Pi}}{X}, j<i\}
      \).
      That is, $L^{-}=\head{r_j}$ for some $r_j\in\GR{\TA{\Pi}}{X}$ with $j<i$.
      With the induction hypothesis, we then obtain
      \(
      L^{-}\in\Th{\reduct{\mathcal{T}(\Pi)}{Y}}
      \).
      By Lemma~\ref{lem:preserving:ok},
      we have $\ok{\name{r_i}}\in\Th{\reduct{\mathcal{T}(\Pi)}{Y}}$.
      Since \Th{\reduct{\mathcal{T}(\Pi)}{Y}} is closed under
      \[
                \bl{2}{r_i}{L^{-}}
      =\reductr{\bl{2}{r_i}{L^{-}}}
      =
      \blocked{\name{r_i}}\LPif\ok{\name{r_i}},L^{-}
      \in\reduct{\mathcal{T}(\Pi)}{Y}
      \ ,
      \]
      we obtain
      \(
      \blocked{\name{r_i}}\in\Th{\reduct{\mathcal{T}(\Pi)}{Y}}
      \). $\mathproofbox$
      
    \end{itemize}
  \end{itemize}
\end{description}
\end{proof*}


\begin{proof}[Proof of Theorem~\ref{thm:conservative}]
Suppose $\Pi$ contains no preference information. 
Obviously, $\Pi$ can then be regarded both as a dynamically ordered program 
as well as a statically ordered program $(\Pi,\PRECMo)$ with ${\PRECMo} =  \emptyset$.
Hence, Theorems~\ref{thm:order:preserving} and \ref{thm:order:preserving:x} 
imply that $X$ is \PRECMo-preserving iff $X$ is \PRECMoi{X}-preserving. 
This proves the equivalence of \ref{thm:conservative:one} and 
\ref{thm:conservative:two}.
Inspecting the conditions of Definition~\ref{def:order:preserving}, it is 
easily seen that any answer set is trivially \PRECMo-preserving for
${\PRECMo}=\emptyset$. 
It follows that \ref{thm:conservative:one} is equivalent to 
\ref{thm:conservative:three}.  
\end{proof}


\begin{proof}[Proof of Theorem~\ref{thm:principles}]

\medskip
\textit{Principle I.}
Let $(\Pi,\PRECMo)$ be a statically ordered logic program, let $X_1$ 
and $X_2$ be two answer sets of $\Pi$
generated by $\GR{\Pi}{X_1}=R\cup\{r_1\}$ and $\GR{\Pi}{X_2}=R\cup\{r_2\}$, 
respectively, where $r_1,r_2\not\in R$,  and assume $\PRECM{r_1}{r_2}$.

Suppose $X_1$ is a $\PRECMo$-preserving answer set of $\Pi$.
Clearly, both $X_1$ and $X_2$ must be consistent. 
Furthermore, we have that $r_2\notin\GR{\Pi}{X_1}$, \iec $r_2$ is defeated by $X_1$.
On the other hand, given that $\GR{\Pi}{X_2}=R\cup\{r_2\}$ and since 
$r_2\notin R$, it follows that $\pbody{r_2}\subseteq\{\head{r'}\mid r'\in R\}$,
which in turn implies that $\pbody{r_2}\subseteq X_1$. 
Since $\PRECM{r_1}{r_2}$ and $X_1$ is assumed to be $\PRECMo$-preserving, we
obtain that 
$\nbody{r_2}\cap\{\head{r'_k}\mid k<j_0\}\neq\emptyset$ for some enumeration 
\(
  \langle r'_i\rangle_{i\in I}
  \)
  of\/ $\GR{\Pi}{X_1}$
and some $j_0$ such that $r_1=r'_{j_0}$.
Hence, there is some $k_0<j_0$ such that $\head{r'_{k_0}}\in\nbody{r_2}$. 
But $r'_{k_0}\in R$, so $r_2$ is defeated by $X_2$, which is a contradiction.
It follows that $X_1$ is not $\PRECMo$-preserving.

\paragraph{Principle II-S.}
Let $(\Pi,\PRECMo)$ be a statically ordered logic program and let $X$ be a
$\PRECMo$-preserving answer set of $\Pi$. 
Furthermore, let $r$ be a rule where $\pbody{r}\not\subseteq X$, and let 
$(\Pi\cup\{r\},\PRECMo')$ be a statically ordered logic program where $<'$ 
is a strict partial order which agrees with $\PRECMo$ on rules from $\Pi$.
We show that $X$ is a $\PRECMo'$-preserving answer set of $\Pi\cup\{r\}$.

First of all, it is rather obvious that $X$ is an answer set of $\Pi\cup\{r\}$.
Now consider an enumeration 
\(
  \langle r_i\rangle_{i\in I}
  \)
of\/ $\GR{\Pi}{X}$ satisfying 
Items~\ref{def:order:preserving:zero}--\ref{def:order:preserving:two} of 
Definition~\ref{def:order:preserving} with respect to $\PRECMo$. 
Since $\GR{\Pi}{X}=\GR{\Pi}{X\cup\{r\}}$, 
\(
  \langle r_i\rangle_{i\in I}
  \)
is also an enumeration of\/ $\GR{\Pi}{X\cup\{r\}}$. 
It remains to show that 
\(
  \langle r_i\rangle_{i\in I}
  \)
satisfies Items~\ref{def:order:preserving:zero}--\ref{def:order:preserving:two} 
of Definition~\ref{def:order:preserving} with respect to $\PRECMo'$.
Clearly, Condition~\ref{def:order:preserving:zero} is unaffected by $\PRECMo'$ 
and is thus still satisfied. 
Moreover, $\PRECMo'$ agrees with $\PRECMo$ on the rules of $\Pi$, so 
Condition~\ref{def:order:preserving:one} holds for $\PRECMo'$ as well.
Finally, Condition~\ref{def:order:preserving:two} holds in virtue of  
$r\notin\GR{\Pi}{X\cup\{r\}}$ and $\pbody{r}\not\subseteq X$.

\paragraph{Principle II-D.}
Let $\Pi$ be a (dynamically) ordered logic program, let $X$ be a 
\PRECMoi{X}-preserving answer set of $\TA{\Pi}$, and consider a rule $r$ such that
$\pbody{r}\not\subseteq X$. 
We must show that $X$ is an \PRECMoi{X}-preserving answer set of $\TA{\Pi}\cup\{r\}$.

If $r\in\TA{\Pi}$, then Principle II-D holds trivially, so let us assume that $r\notin\TA{\Pi}$.
Similar to the above, $X$ is clearly an answer set of $\TA{\Pi}\cup\{r\}$.
Let 
\(
  \langle r_i\rangle_{i\in I}
  \)
be an enumeration of\/ $\TA{\Pi}$ satisfying 
Items~\ref{def:order:preserving:x:one}--\ref{def:order:preserving:x:two} of 
Definition~\ref{def:order:preserving:x}. Define  
\(
  \langle r'_j\rangle_{j\in J}
  \)
as the sequence starting with $r$ and continuing with the sequence 
\(
  \langle r_i\rangle_{i\in I}
\).
Clearly, 
\(
  \langle r'_j\rangle_{j\in J}
\)
is an enumeration of $\TA{\Pi}\cup\{r\}$.
Moreover, it satisfies Items~\ref{def:order:preserving:x:one}--\ref{def:order:preserving:x:two} of 
Definition~\ref{def:order:preserving:x}.
To see this, we first note that, according to the definition of an ordered program,  the term $\name{r}$
does not occur in the rules of $\TA{\Pi}$ because $r$ is assumed not to be a member of $\TA{\Pi}$. 
Hence, the relation $\PRECMi{X}{r'_i}{r'_j}$ holds at most for rules $r'_i,r'_j\in\TA{\Pi}$.
Consequently, 
\(
  \langle r'_j\rangle_{j\in J}
  \)
obeys Items~\ref{def:order:preserving:x:one} and \ref{def:order:preserving:x:tri}, because 
\(
  \langle r_i\rangle_{i\in I}
  \)
does.
Furthermore, Item~\ref{def:order:preserving:x:zero} holds, because 
$\GR{\TA{\Pi}}{X\cup\{r\}}=\GR{\TA{\Pi}}{X}$.
Concerning the final condition of Definition~\ref{def:order:preserving:x}, 
observe that 
$(\TA{\Pi}\cup\{r\})\setminus \GR{\TA{\Pi}}{X\cup\{r\}}=(\TA{\Pi}\setminus 
\GR{\TA{\Pi}}{X})\cup\{r\}$.
Now, each $r'\in\TA{\Pi}\setminus \GR{\TA{\Pi}}{X}$ satisfies 
Condition~\ref{def:order:preserving:x:two}, and $r$ satisfies it as well, 
because $\pbody{r}\not\subseteq X$.
Therefore, Condition~\ref{def:order:preserving:x:two} is met. It follows that 
$X$ is \PRECMoi{X}-preserving.
%
\end{proof}

\subsection{Proofs of Section~\ref{sec:brewka:eiter}}

\begin{proof}[Proof of Theorem~\ref{thm:brewka:eiter}]
Let $(\Pi,<)$ be a statically ordered logic program over language $\mathcal{L}$.

We abbreviate
\(
\mathcal{U}(\Pi)\cup\{(\PREC{n_1}{n_2})\LPif\,\mid (r_1, r_2) \in\; < \}
\)
by
\(
\mathcal{U}'(\Pi)
\).

\paragraph{Only-if part.}

Let $X$ be a consistent \BEpreferred{} answer set of $(\Pi,<)$.
By definition, $X$ is then a standard answer set of $\Pi$.
Define
\begin{eqnarray*}
  Y &=& \hphantom{\cup\;} \{\head{r} \mid r\in\GR{\Pi}{X}\}
       \cup  \{\head{r'}\mid r\in\GR{\Pi}{X}\}
  \\& &\cup\;\{\ok{\name{r}}   \mid r\in\Pi\}
       \cup  \{\rdy{\name{r}}{\name{r'}}\mid r,r'\in\Pi\}
  \\& &\cup\;\{\applied{\name{r}}\mid r    \in\GR{\Pi}{X}\}
       \cup  \{\blocked{\name{r}}\mid r\not\in\GR{\Pi}{X}\}
  \\& &\cup\;\{    (\PREC{\name{r}}{\name{r'}})\mid (r ,r') \in\; <\}
       \cup  \{\neg(\PREC{\name{r}}{\name{r'}})\mid (r',r ) \in\; <\} \mbox{.}
\end{eqnarray*}
We note the following useful relationships for $X$ and $Y$.
\begin{lemma}\label{lem:be:X:Y}
  For any $L\in\Lit$, we have
  \begin{enumerate}
  \item $L\in X$   iff $L\in Y$;
  \item $L'\in X'$ iff $L'\in Y$; and
  \item $L\in Y$   iff $L'\in Y$.
  \end{enumerate}
\end{lemma}
Given that $X$ is consistent, this implies that $Y$ is consistent, too.
\begin{lemma}\label{lem:be:prec:i}
  For $r,s\in\Pi$, we have
  $\PREC{\name{r}}{\name{s}}\in Y$ iff \PRECM{r}{s}.
\end{lemma}

We must show that $Y$ is an answer set of $\mathcal{U}'(\Pi)$,
that is,
\(
Y=\Th{\mathcal{U}'(\Pi)^Y}
\).

\paragraph{``$\,\supseteq$'' part.}
We start by showing that $Y$ is closed under $\mathcal{U}'(\Pi)^Y$.
For this,
we show for $r\in\mathcal{U}'(\Pi)$
that $\head{\reductr{r}}\in Y$
whenever
$\reductr{r}\in{\mathcal{U}'(\Pi)^Y}$ with $\pbody{r}\subseteq Y$.

\begin{enumerate}
  \newcommand{\RULEtem}[3]{%
  \item Consider
    \(
    #1:{#2 \;\LPif \; #3}\in\mathcal{U}'(\Pi)
    \).}
  \newcommand{\mRULEtem}[3]{\RULEtem{#1}{#2}{#3}
    We have
    \(
    (#1)^{+}={#1}
    \).
    Clearly,
    \(
    {#1}\in\reduct{\mathcal{U}'(\Pi)}{Y}
    \).}
  \newcommand{\nmRULEtem}[5]{\RULEtem{#1}{#2}{#3}
    Suppose
    \(
    (#1)^{+}={#2\;\LPif\;#4}\in\reduct{\mathcal{U}'(\Pi)}{Y}
    \)
    and
    \(
    {#5}\cap Y=\emptyset
    \).}
  \mRULEtem{\ap{1}{r}}%
           {\head{r'}}%
           {\applied{\name{r}}}
  By definition of $Y$,    
  we have $\head{r'}\in Y$ whenever $\applied{\name{r}}\in Y$.
  
  \item Consider
    \(
    \ap{2}{r}:{\applied{\name{r}} \;\LPif \; \ok{\name{r}},\body{r},\naf\nbody{r'}}\in\mathcal{U}'(\Pi)
    \),
    and suppose that
    \(
    (\ap{2}{r})^{+}
\in\reduct{\mathcal{U}'(\Pi)}{Y}
    \),
    \iec
    \(
    {(\nbody{r}\cup\nbody{r'})}\cap Y=\emptyset
    \).
  Assume
  \(
  \{\ok{\name{r}}\}\cup\pbody{r}\subseteq Y
  \).
  From Condition~1 of Lemma~\ref{lem:be:X:Y} we get
  \(
  \pbody{r}\subseteq X
  \).
  Accordingly, $(\nbody{r}\cup\nbody{r'})\cap Y=\emptyset$ implies
  $\nbody{r}\cap X=\emptyset$.
  Hence, we have $r\in\GR{\Pi}{X}$, which implies ${\applied{\name{r}}}\in Y$ by
  definition of $Y$.
  
  \item Consider
    \(
    \bl{1}{r}{L}:{\blocked{\name{r}}} \;\LPif \; \ok{\name{r}},\naf L,\naf L'\in\mathcal{U}'(\Pi)
    \),
    and suppose that
    \(
    (\bl{1}{r}{L})^{+}=\blocked{\name{r}}\;\LPif\; \ok{\name{r}}\in\reduct{\mathcal{U}'(\Pi)}{Y}
    \)
    with
    \(
    {\{L,L'\}}\cap Y=\emptyset
    \).
%
  The latter and Condition~1 of Lemma~\ref{lem:be:X:Y} imply that
  \(
  L\not\in X
  \)
  for $L\in\pbody{r}$.
  Hence, $r\not\in\GR{\Pi}{X}$, implying that ${\blocked{\name{r}}}\in Y$ by
  definition of $Y$.
  
  \mRULEtem{\bl{2}{r}{K}}%
           {\blocked{\name{r}}}%
           {\ok{\name{r}},K,K'}
  Suppose
  \(
  \{\ok{\name{r}},K,K'\}\subseteq Y
  \).
  Condition~1 of Lemma~\ref{lem:be:X:Y} implies that
  \(
  K\in X
  \)
  for $K\in\nbody{r}$.
  Hence, $r\not\in\GR{\Pi}{X}$, implying that ${\blocked{\name{r}}}\in Y$ by
  definition of $Y$.

  \mRULEtem{\cokt{1}{r}}%
           {\ok{\name{r}}}%
           {\rdy{\name{r}}{\name{r_1}},\dots,\rdy{\name{r}}{\name{r_k}}}
  We trivially have ${\ok{\name{r}}}\in Y$ by definition of $Y$.
  \item Consider the case that 
    \(
    \{{\cok{2}{r}{s}}, {\cok{3}{r}{s}}, {\cok{4}{r}{s}}, {\cok{5}{r}{s,J}}\}
    \subseteq
    \mathcal{U}'(\Pi)
    \).
    Clearly,
    \(
    \head{\cok{i}{r}{s}}={\rdy{\name{r}}{\name{s}}}
    \)
    for $i=2,\dots,5$.
    We trivially have ${\rdy{\name{r}}{\name{s}}}\in Y$ by definition of $Y$.
  \item Consider
    \(
    \de{r}:{\LPif \naf\ok{\name{r}}}\in\mathcal{U}'(\Pi)
    \).
    Suppose
    \(
    (\de{r})^{+}\in\reduct{\mathcal{U}'(\Pi)}{Y}
    \),
    \iec
    \(
    \ok{\name{r}}\cap Y=\emptyset
    \).
%
  This is impossible by definition of $Y$.
  Hence, $\reductr{\de{r}}\not\in\mathcal{U}'(\Pi)^Y$.
  
  \item Consider
    \(
    t(r,s,t):{\PREC{\name{r}}{\name{t}}} \;\LPif \; \PREC{\name{r}}{\name{s}},\PREC{\name{s}}{\name{t}}\in\mathcal{U}'(\Pi)
    \).
Then, we have that
\(
(t(r,s,t))^{+} = t(r,s,t)
\).
Clearly,
    \(
    {t(r,s,t)}\in\reduct{\mathcal{U}'(\Pi)}{Y}
    \).
%
  Assume $\{{\PREC{\name{r}}{\name{s}},\PREC{\name{s}}{\name{t}}}\}\subseteq Y$.
  By definition of $Y$, this implies \PRECM{r}{s} and \PRECM{s}{t}.
  By transitivity of $<$, we then obtain \PRECM{r}{t}.
  Again, by definition of $Y$, we get ${\PREC{\name{r}}{\name{t}}\in Y}$.

  \mRULEtem{as(r,s)}%
           {{\neg(\PREC{\name{s}}{\name{r}})}}%
           {\PREC{\name{r}}{\name{s}}}
  Assume ${\PREC{\name{r}}{\name{s}}}\in Y$.
  By definition of $Y$, this implies \PRECM{r}{s}.
  Again, by definition of $Y$, we then get
  $\neg({\PREC{\name{s}}{\name{r}})\in Y}$. 
  
\end{enumerate}
Note that the closure of $Y$ under $\mathcal{U}'(\Pi)^Y$ shows that
$Y\supseteq\Th{\mathcal{U}'(\Pi)^Y}$.

\paragraph{``$\,\subseteq$'' part.}
We must now show that $Y$ is the smallest set being closed under
$\mathcal{U}'(\Pi)^Y$,
or equivalently that $Y\subseteq\Th{\mathcal{U}'(\Pi)^Y}$.

First, we show that
\begin{equation}
  \label{eq:be:first}
  \{\head{r} \mid r\in\GR{\Pi}{X}\}
  \subseteq
  \Th{\mathcal{U}'(\Pi)^Y}
  \ \mbox{.}
\end{equation}
Observe that
\(
\{\head{r} \mid r\in\GR{\Pi}{X}\}
=
\mbox{$\bigcup$}_{i\geq 0}\Tind{i}{\reduct{\Pi}{X}}{\emptyset}
\).
Given this, we proceed by induction and show that
\(
\Tind{i}{\reduct{\Pi}{X}}{X}\subseteq\Th{\mathcal{U}'(\Pi)^Y}
\)
for $i\geq 0$.
\begin{description}
\item[Base.] Trivial, since $\Tind{0}{\reduct{\Pi}{X}}{\emptyset}={\emptyset}$.
\item[Step.] Assume
  \(
  \Tind{j}{\reduct{\Pi}{X}}{\emptyset}\subseteq\Th{\mathcal{U}'(\Pi)^Y}
  \)
  for $0 \leq j \leq i$,
  and consider
  \(
  \head{\reductr{r}}\in\Tind{i+1}{\reduct{\Pi}{X}}{\emptyset}
  \)
  for some ${\reductr{r}}\in\reduct{\Pi}{X}$.
  
  Given that $r\in{\mathcal{U}'(\Pi)}$ and that
  $X=Y\cap\mathcal{L}$ by Condition~1 of Lemma~\ref{lem:be:X:Y},
  we also have ${\reductr{r}}\in{\mathcal{U}'(\Pi)}^Y$.

  Now,
  \(
  \head{\reductr{r}}\in\Tind{i+1}{\reduct{\Pi}{X}}{\emptyset}
  \)
  implies that
  \(
  \body{\reductr{r}}\subseteq\Tind{i}{\reduct{\Pi}{X}}{\emptyset}
  \).
  Then, the induction hypothesis provides us with
  \(
  \body{\reductr{r}}\subseteq\Th{\mathcal{U}'(\Pi)^Y}
  \).
  The fact that \Th{\mathcal{U}'(\Pi)^Y} is closed under
  ${\mathcal{U}'(\Pi)^Y}$ implies that
  \(
  \head{\reductr{r}}\in\Th{\mathcal{U}'(\Pi)^Y}
  \).
\end{description}

Second, 
we show that
\begin{equation}
  \label{eq:be:second}
        \{    (\PREC{\name{r}}{\name{r'}})\mid (r,r')    \in\; <\}
  \cup  \{\neg(\PREC{\name{r}}{\name{r'}})\mid (r,r')\not\in\; <\}
  \subseteq
  \Th{\mathcal{U}'(\Pi)^Y}
   \mbox{.}
\end{equation}
By definition, we have
\(
\{(\PREC{\name{r}}{\name{r'}})\LPif\,\mid (r, r') \in\; < \}
\subseteq
{\mathcal{U}'(\Pi)}
\)
as well as
\(
\{(\PREC{\name{r}}{\name{r'}})\LPif\,\mid (r, r') \in\; < \}
\subseteq
{\mathcal{U}'(\Pi)^Y}
\).
From this, we obviously get
\(
\{(\PREC{\name{r}}{\name{r'}})\mid (r, r') \in\; < \}
\subseteq
\Th{\mathcal{U}'(\Pi)^Y}
\).
On the other hand,
we have $as(r,r')\in{\mathcal{U}'(\Pi)}$ and clearly also
$\reductr{as(r,r')}\in{\mathcal{U}'(\Pi)^Y}$.
The fact that \Th{\mathcal{U}'(\Pi)^Y} is closed under
${\mathcal{U}'(\Pi)^Y}$ implies that
\(
\head{\reductr{as(r,r')}}\in\Th{\mathcal{U}'(\Pi)^Y}
\)
whenever
\(
\body{\reductr{as(r,r')}}\in\Th{\mathcal{U}'(\Pi)^Y}
\).
That is,
\(
{\neg(\PREC{\name{r'}}{\name{r}})}\in\Th{\mathcal{U}'(\Pi)^Y}
\)
whenever
\(
{\PREC{\name{r}}{\name{r'}}}\in\Th{\mathcal{U}'(\Pi)^Y}
\).
Given that we have just shown that
\(
\{(\PREC{\name{r}}{\name{r'}})\mid (r, r') \in\; < \}
\subseteq
\Th{\mathcal{U}'(\Pi)^Y}
\)
holds, we get
\(
\{\neg(\PREC{\name{r'}}{\name{r}})\mid (r, r') \in\; < \}
\subseteq
\Th{\mathcal{U}'(\Pi)^Y}
\).

Third, 
we show that
\[
\begin{array}{clcl}
       & \{\head{r'},\applied{\name{r}}\mid r\in\GR{\Pi}{X}\} &&
\\\cup & \{\blocked{\name{r}}\mid r\not\in\GR{\Pi}{X}\}       &&
\\\cup & \{\ok{\name{r}}   \mid r\in\Pi\}              
  \cup   \{\rdy{\name{r}}{\name{r'}}\mid r,r'\in\Pi\}         
&\subseteq&
\Th{\mathcal{U}'(\Pi)^Y} 
\ \mbox{.}
\end{array}
\]
We do so by induction on ${\ll}$,
the total order on $\Pi$ inducing $X$ according to
Definition~\ref{def:be:total:order}.
\begin{description}
\item[Base.] 
  Let $r_0\in\Pi$ be the $\ll$-greatest rule.
  In analogy to the base case in the ``$\subseteq$''-part of the 
  proof of the only-if direction of
  Theorem~\ref{thm:order:preserving:x},
  we can prove that ${\ok{\name{r_0}}}\in\Th{\reduct{\mathcal{U}'(\Pi)}{Y}}$.
  Given this, we can show that $\applied{\name{r_0}}\in\Th{\mathcal{U}'(\Pi)^Y}$
  or $\blocked{\name{r_0}}\in\Th{\mathcal{U}'(\Pi)^Y}$ holds in analogy to the
  (more general) proof carried out in the induction step below.
\item[Step.] Consider $r\in\Pi$ and assume that our claim holds for all
  $s\in\Pi$ such that $r\ll s$.

  We start by providing the following lemma.
  \begin{lemma}\label{lem:be:ok:i}
    Given the induction hypothesis, we have
    \begin{enumerate}[(ii).]
    \renewcommand{\theenumi}{(\roman{enumi})}
    \item 
      \(
      \ok{\name{r}}\in\Th{\mathcal{U}'(\Pi)^Y}
      \)
      and
    \item 
      \(
      \{\rdy{\name{r}}{\name{s}}\mid s\in\Pi\}\subseteq\Th{\mathcal{U}'(\Pi)^Y} 
      \).
    \end{enumerate}
  \end{lemma}
  \begin{proof}[Proof of Lemma~\ref{lem:be:ok:i}]
    First, we prove 
    \(
    \{\rdy{\name{r}}{\name{s}}\mid s\in\Pi\}\subseteq\Th{\mathcal{U}'(\Pi)^Y} 
    \).

    Consider ${\rdy{\name{r}}{\name{s}}}\in Y$ for some $s\in\Pi$.
    We distinguish the following cases.
    
    \begin{itemize}
      
    \item Assume that $\head{s}\in X$ and $\nbody{s}\cap X\neq\emptyset$.
      That is, $J\in X$ for some $J\in\nbody{s}$.
      We have shown in~(\ref{eq:be:first}) that
      \(
      X\subseteq\Th{\mathcal{U}'(\Pi)^Y}
      \).
      Hence,
      \(
      \{\head{s},J\}\subseteq X
      \)
      implies
      \(
      \{\head{s},J\}\subseteq\Th{\mathcal{U}'(\Pi)^Y}
      \).
      That is,
      \(
      \body{\reductr{\cok{5}{r}{s,J}}}\subseteq\Th{\mathcal{U}'(\Pi)^Y}
      \).
      We clearly have
      \(
      \reductr{\cok{5}{r}{s,J}}\in{\mathcal{U}'(\Pi)^Y}
      \).
      Since \Th{\mathcal{U}'(\Pi)^Y} is closed under ${\mathcal{U}'(\Pi)^Y}$,
      we have $\head{\reductr{\cok{5}{r}{s,J}}}\in\Th{\mathcal{U}'(\Pi)^Y}$.
      That is, $\rdy{\name{r}}{\name{s}}\in\Th{\mathcal{U}'(\Pi)^Y}$.
      
    \item 
      Assume $r\not\ll s$.
      Since $Y$ is consistent, we thus get
      $(\PREC{\name{r}}{\name{s}})\not\in Y$.
      Hence, we have $\reductr{\cok{2}{r}{s}}\in{\mathcal{U}'(\Pi)^Y}$. 
      Trivially, we have
      \(
      \body{\reductr{\cok{2}{r}{s}}}
      =
      \emptyset
      \subseteq
      \Th{\mathcal{U}'(\Pi)^Y}
      \).
      Since \Th{\mathcal{U}'(\Pi)^Y} is closed under ${\mathcal{U}'(\Pi)^Y}$,
      we have $\head{\reductr{\cok{2}{r}{s}}}\in\Th{\mathcal{U}'(\Pi)^Y}$.
      That is, ${\rdy{\name{r}}{\name{s}}}\in\Th{\mathcal{U}'(\Pi)^Y}$.

    \item Otherwise, $r\ll s$ holds.
      By the induction hypothesis, we have
      \[
      \{\applied{\name{s}}\mid r\ll s,s    \in\GR{\Pi}{X}\}
      \cup
      \{\blocked{\name{s}}\mid r\ll s,s\not\in\GR{\Pi}{X}\}
      \subseteq
      \Th{\mathcal{U}'(\Pi)^Y} \ \mbox{.}
      \]
      Moreover, we get
      \(
      (\PREC{\name{r}}{\name{s}})\in\Th{\mathcal{U}'(\Pi)^Y}
      \)
      by what we have shown in~(\ref{eq:be:second}).
      We distinguish the following two cases.
        
      \begin{itemize}
        
      \item Assume $\nbody{s}\cap X=\emptyset$.
        If $\pbody{s}\subseteq X$, then $s\in\GR{\Pi}{X}$.
        By the induction hypothesis, we get
        $\applied{\name{s}}\in\Th{\mathcal{U}'(\Pi)^Y}$.
        Therefore, we have
        \(
        \body{\reductr{\cok{3}{r}{s}}}
        =
        \{{(\PREC{\name{r}}{\name{s}}),\applied{\name{s}}}\}
        \subseteq
        \Th{\mathcal{U}'(\Pi)^Y}
        \).
        Clearly, we have $\reductr{\cok{3}{r}{s}}\in{\mathcal{U}'(\Pi)^Y}$.
        Since \Th{\mathcal{U}'(\Pi)^Y} is closed under ${\mathcal{U}'(\Pi)^Y}$,
        we have $\head{\reductr{\cok{3}{r}{s}}}\in\Th{\mathcal{U}'(\Pi)^Y}$.
        That is, ${\rdy{\name{r}}{\name{s}}}\in\Th{\mathcal{U}'(\Pi)^Y}$.
        If $\pbody{s}\not\subseteq X$, then $s\not\in\GR{\Pi}{X}$,
        as dealt with next.
      \item Assume $\head{s}\not\in X$.
        That is, $s\not\in\GR{\Pi}{X}$.
        By the induction hypothesis, we get
        $\blocked{\name{s}}\in\Th{\mathcal{U}'(\Pi)^Y}$.
        Therefore, we have
        \(
        \body{\reductr{\cok{4}{r}{s}}}
        =
        \{{(\PREC{\name{r}}{\name{s}}),\blocked{\name{s}}}\}
        \subseteq
        \Th{\mathcal{U}'(\Pi)^Y}
        \).
        Clearly, we have $\reductr{\cok{4}{r}{s}}\in{\mathcal{U}'(\Pi)^Y}$.
        Since \Th{\mathcal{U}'(\Pi)^Y} is closed under ${\mathcal{U}'(\Pi)^Y}$,
        we have $\head{\reductr{\cok{4}{r}{s}}}\in\Th{\mathcal{U}'(\Pi)^Y}$.
        That is, ${\rdy{\name{r}}{\name{s}}}\in\Th{\mathcal{U}'(\Pi)^Y}$.
      \end{itemize}
    \end{itemize}
    We have shown that
    \(
    {\rdy{\name{r}}{\name{s}}}\in\Th{\mathcal{U}'(\Pi)^Y}
    \)
    for all $s\in\Pi$.
    This implies that
    \(
    \body{\reductr{\cokt{1}{r}}}\subseteq\Th{\mathcal{U}'(\Pi)^Y}
    \).
    Clearly, we have $\reductr{\cokt{1}{r}}\in{\mathcal{U}'(\Pi)^Y}$.
    Since \Th{\mathcal{U}'(\Pi)^Y} is closed under ${\mathcal{U}'(\Pi)^Y}$,
    we have $\head{\reductr{\cokt{1}{r}}}\in\Th{\mathcal{U}'(\Pi)^Y}$.
    That is, ${\ok{\name{r}}}\in\Th{\mathcal{U}'(\Pi)^Y}$.
  \end{proof}

  We distinguish the following three cases.
  \begin{enumerate}
  \item Assume $\pbody{r}\subseteq X$ and $\nbody{r}\cap X=\emptyset$.
    That is, $r\in\GR{\Pi}{X}$.

    From $\nbody{r}\cap X=\emptyset$, we deduce 
    $\nbody{r}\cap Y=\emptyset$, by Condition~1 of 
    Lemma~\ref{lem:be:X:Y}.
    Moreover, Condition~3 of Lemma~\ref{lem:be:X:Y}
    gives $\nbody{r'}\cap Y=\emptyset$.
    Therefore, $\reductr{\ap{2}{r}}\in{\mathcal{U}'(\Pi)^Y}$.
    We have just shown in~(\ref{eq:be:first}) that
    \(
    X\subseteq\Th{\mathcal{U}'(\Pi)^Y}
    \).
    This and $\pbody{r}\subseteq X$ imply that
    \(
    \pbody{r}\subseteq\Th{\mathcal{U}'(\Pi)^Y}
    \).
    From Lemma~\ref{lem:be:ok:i},
    we get $\ok{\name{r}}\in\Th{\mathcal{U}'(\Pi)^Y}$.
    We thus have $\body{\reductr{\ap{2}{r}}}\subseteq\Th{\mathcal{U}'(\Pi)^Y}$.
    Since \Th{\mathcal{U}'(\Pi)^Y} is closed under ${\mathcal{U}'(\Pi)^Y}$,
    we have $\head{\reductr{\ap{2}{r}}}\in\Th{\mathcal{U}'(\Pi)^Y}$.
    That is, $\applied{\name{r}}\in\Th{\mathcal{U}'(\Pi)^Y}$.

    Clearly, $\reductr{\ap{1}{r}}\in{\mathcal{U}'(\Pi)^Y}$.
    Given $\applied{\name{r}}\in\Th{\mathcal{U}'(\Pi)^Y}$ 
    and the fact that \Th{\mathcal{U}'(\Pi)^Y} is closed under
    ${\mathcal{U}'(\Pi)^Y}$,
    we have $\head{\reductr{\ap{1}{r}}}\in\Th{\mathcal{U}'(\Pi)^Y}$.
    That is, $\head{r'}\in\Th{\mathcal{U}'(\Pi)^Y}$. 

  \item Assume $\pbody{r}\subseteq X$ and $\nbody{r}\cap X\neq\emptyset$.

    By Theorem~\ref{thm:be:characterisation},
    there is then some rule $r^*\in\GR{\Pi}{X}$ such that
    $r\ll r^*$ and $\head{r^*}\in\nbody{r}$.
    Clearly, $r^*\in\GR{\Pi}{X}$ implies that $\head{r^*}\in X$.
    We have shown in~(\ref{eq:be:first}) that
    \(
    X\subseteq\Th{\mathcal{U}'(\Pi)^Y}
    \).
    Accordingly, $\head{r^*}\in\Th{\mathcal{U}'(\Pi)^Y}$.
    By the induction hypothesis,
    we also have $\head{r^*}'\in\Th{\mathcal{U}'(\Pi)^Y}$.
    From Lemma~\ref{lem:be:ok:i},
    we obtain $\ok{\name{r}}\in\Th{\mathcal{U}'(\Pi)^Y}$.
    We thus have
    \(
    \body{\reductr{\bl{2}{r}{K}}}
    =
    \{\ok{\name{r}},K,K'\}
    \subseteq
    \Th{\mathcal{U}'(\Pi)^Y}
    \) for
    $K=\head{r^*}$.
    Clearly, we have $\reductr{\bl{2}{r}{K}}\in{\mathcal{U}'(\Pi)^Y}$.
    Given that \Th{\mathcal{U}'(\Pi)^Y} is closed under
    ${\mathcal{U}'(\Pi)^Y}$,
    we obtain that
    $\head{\reductr{\bl{2}{r}{K}}}\in\Th{\mathcal{U}'(\Pi)^Y}$.
    That is, $\blocked{\name{r}}\in\Th{\mathcal{U}'(\Pi)^Y}$.

  \item Assume $\pbody{r}\not\subseteq X$.
    That is, there is some $L\in\pbody{r}$ such that $L\not\in X$.

    By Lemma~\ref{lem:be:X:Y}, we then also have $L\not\in Y$
    and $L'\not\in Y$.
    We then have $\reductr{\bl{1}{r}{L}}\in{\mathcal{U}'(\Pi)^Y}$.
    Given that $\ok{\name{r}}\in\Th{\mathcal{U}'(\Pi)^Y}$ by
    Lemma~\ref{lem:be:ok:i}, and since \Th{\mathcal{U}'(\Pi)^Y} is closed under
    ${\mathcal{U}'(\Pi)^Y}$,
    we obtain that $\head{\reductr{\bl{1}{r}{L}}}\in\Th{\mathcal{U}'(\Pi)^Y}$.
    That is, $\blocked{\name{r}}\in\Th{\mathcal{U}'(\Pi)^Y}$.
  \end{enumerate}
\end{description}
This completes our proof of $Y\subseteq\Th{\mathcal{U}'(\Pi)^Y}$.

In all, we have thus shown that $Y=\Th{\mathcal{U}'(\Pi)^Y}$.
That is, $Y$ is an answer set of ${\mathcal{U}'(\Pi)}$.

\paragraph{If part.}

Let $Y$ be a consistent answer set of $\mathcal{U}'(\Pi)$.
We show that
\(
X=Y\cap\mathcal{L}_\mathcal{A}
\)
is a \BEpreferred{} answer set of $(\Pi,<)$.

The definition of $\mathcal{U}'(\Pi)$ induces the following properties for $X$
and $Y$.
\begin{lemma}\label{lem:be:Y:X}
  For any $L\in\Lit$, we have that 
  $L\in X$ iff $L\in Y$.
\end{lemma}
Given that $Y$ is consistent, this implies that $X$ is consistent, too.
\begin{lemma}\label{lem:be:prec:ii}
  For $r,s\in\Pi$, we have
  $\PREC{\name{r}}{\name{s}}\in Y$ iff \PRECM{r}{s}.
\end{lemma}

Consider $r\in\Pi$ such that $\pbody{r}\subseteq X$ and $\head{r}\not\in X$.
To show that $X$ is \BEpreferred{}, we must prove, according to
Theorem~\ref{thm:be:characterisation}, that there is some rule
${r^*}\in\Pi$ such that
(i) ${r^*}\in\GR{\Pi}{X}$,
(ii) $\head{r^*}\in\nbody{r}$, and
(iii) $r\ll r^*$.

Since $X$ is a standard answer set of $\Pi$ 
(by Property~\ref{thm:translation:results:1} of 
Proposition~\ref{thm:translation:results}), 
our choice of $r$ implies that
\(
\nbody{r}\cap X\neq\emptyset
\).
By Lemma~\ref{lem:be:Y:X},
the latter condition
implies, in turn, $\nbody{r}\cap Y\neq\emptyset$.
Hence,
\(
\reductr{\ap{2}{r}}\not\in{\mathcal{U}'(\Pi)^Y}
\).
And so, $\applied{\name{r}}\not\in Y$.
Consequently, we obtain $\blocked{\name{r}}\in Y$
from Condition~\ref{l:be2:results:3} of
Proposition~\ref{thm:be:results}
and the fact that $Y$ is consistent.
Moreover, $\pbody{r}\subseteq X\subseteq Y$ implies
\(
\reductr{\bl{1}{r}{L}}\not\in{\mathcal{U}'(\Pi)^Y}
\)
for all $L\in\pbody{r}$.
Since $\blocked{\name{r}}\in Y$, we must therefore have
\(
{\bl{2}{r}{K}}\in\GR{\mathcal{U}'(\Pi)}{Y}
\)
for some $K\in\nbody{r}$.
That is,
\(
\pbody{\bl{2}{r}{K}}
=
\{\ok{\name{r}},K,K'\}
\subseteq
Y
\).
Hence, there is some $r^*\in\Pi$ with
\(
\head{r^*}=K
\),
establishing Condition~(ii).

More precisely,
we have
\(
\head{r^*}'=K'
\)
such that
\(
\applied{\name{r^*}}\in Y
\).
That is,
\(
\{{\ap{1}{r^*}},{\ap{2}{r^*}}\}\subseteq\GR{\mathcal{U}'(\Pi)}{Y}
\).
In fact, ${\ap{2}{r^*}}\in\GR{\mathcal{U}'(\Pi)}{Y}$ implies that
\(
\pbody{r^*}\subseteq Y
\)
and
\(
\nbody{r^*}\cap Y\neq\emptyset
\)
holds.
With Lemma~\ref{lem:be:Y:X},
we get furthermore that
\(
\pbody{r^*}\subseteq X
\)
and
\(
\nbody{r^*}\cap X\neq\emptyset
\).
That is, ${r^*}\in\GR{\Pi}{X}$,
establishing Condition~(i).

Because $Y$ is an answer set of ${\mathcal{U}'(\Pi)}$ there is some minimal
$k$ such that
\(
\{\ok{\name{r}},K,K'\}\subseteq\Tind{k}{\mathcal{U}'(\Pi)^Y}{\emptyset}
\).
By minimality of $k$, we have
\(
\blocked{\name{r}}\not\in\Tind{k}{\mathcal{U}'(\Pi)^Y}{\emptyset}
\).
Obviously,
we also have
\(
\applied{\name{r}}\not\in\Tind{k}{\mathcal{U}'(\Pi)^Y}{\emptyset}
\)
because
\(
\applied{\name{r}}\not\in Y
\).
On the other hand, for deriving $K'$, we must have
\(
\applied{\name{r^*}}\in\Tind{k}{\mathcal{U}'(\Pi)^Y}{\emptyset}
\).
This implies that
\(
\ok{\name{r^*}}\in\Tind{k}{\mathcal{U}'(\Pi)^Y}{\emptyset}
\)
and, in particular, that
\(
\rdy{\name{r^*}}{\name{r}}\in\Tind{k}{\mathcal{U}'(\Pi)^Y}{\emptyset}
\).
Now, observe that
\(
{\cok{5}{\name{r^*}}{\name{r},J}}\not\in\GR{\mathcal{U}'(\Pi)}{Y}
\)
because $\nbody{r}\cap Y\neq\emptyset$, as shown above.
Also, we have just shown that
\(
\applied{\name{r}}\not\in\Tind{k}{\mathcal{U}'(\Pi)^Y}{\emptyset}
\)
and
\(
\blocked{\name{r}}\not\in\Tind{k}{\mathcal{U}'(\Pi)^Y}{\emptyset}
\).
Therefore, we must have
\(
\rdy{\name{r^*}}{\name{r}}\in\Tind{k}{\mathcal{U}'(\Pi)^Y}{\emptyset}
\)
because of
\(
{\cok{2}{\name{r^*}}{\name{r}}}\in\GR{\mathcal{U}'(\Pi)}{Y}
\).
That is,
\(
\reductr{\cok{2}{\name{r^*}}{\name{r}}}\in\reduct{\mathcal{U}'(\Pi)}{Y}
\)
which implies that $(\PREC{\name{r^*}}{\name{r}})\not\in Y$.
This and Lemma~\ref{lem:be:prec:ii} imply that \notPRECM{r^*}{r}.
That is,
either
\PRECM{r}{r^*}
holds
or
neither \notPRECM{r^*}{r} nor \notPRECM{r}{r^*} is true.
In either case, there is some total order
$\ll$ extending $<$
with
\(
r\ll r^*
\).
This establishes Condition~(iii).

In all, we therefore obtain with Theorem~\ref{thm:be:characterisation} that $X$ is
a \BEpreferred{} answer set of $\Pi$.
\end{proof}


\begin{proof*}[Proof of Proposition~\ref{thm:translation:results}]
Let $\Pi$ be an ordered logic program over $\mathcal{L}$
and
let $X$ be a consistent answer set of~$\mathcal{U}(\Pi)$.

\begin{itemize}
\item [\ref{thm:translation:results:1}.]
The consistency of $X\cap\mathcal{L}$ follows from that of $X$.

Since $X$ is an answer set of $\mathcal{U}(\Pi)$,
we have
\(
X=\Th{\mathcal{U}(\Pi)^X}
\).
That is,
\[
X=\Th{\reduct{\Pi}{X}\cup\reduct{(\mathcal{U}(\Pi)\setminus\Pi)}{X}}
\ \mbox{.}
\]
Observe that
\(
\reduct{\Pi}{X}=\reduct{\Pi}{X\cap\mathcal{L}}
\)
and
\(
\head{\reduct{(\mathcal{U}(\Pi)\setminus\Pi)}{X}}
\cap
\mathcal{L}
=
\emptyset
\).
Consequently, we have
\(
X\cap\mathcal{L}
=
\Th{\reduct{\Pi}{X\cap\mathcal{L}}}
\).

\item [\ref{thm:translation:results:2}.]
Consider $L\in\mathcal{L}$. 
We show $L\in X$ iff $L'\in X$.

\paragraph{If part.}
Suppose $L'\in X$. 
Then, there is some $r\in\Pi$ such that 
$L=\head{r}$ and $\applied{\name{r}}\in X$. 
From the latter, we get that 
\(
\ap{2}{r}\in\GR{\mathcal{U}(\Pi)}{X}
\) 
must hold, which in turn implies 
\(\pbody{r}\subseteq X\)
and 
$\nbody{r}\cap X=\emptyset$. 
Hence, $L\in X$.

\paragraph{Only-if part.}
Assume $L\in X$. 
Since $L\in\mathcal{L}$, there is some $r\in \Pi$ 
such that 
$r\in\GR{\mathcal{U}(\Pi)}{X}$.
Suppose $L'\notin X$. 
Then, $\applied{\name{r}}\notin X$. 
Since $\ok{\name{r}}\in X$ 
(by Condition~\ref{l:be2:results:2} of Proposition~\ref{thm:be:results}) and both $\pbody{r}\subseteq X$ and 
$\nbody{r}\cap X=\emptyset$ holds (by condition  
$r\in\GR{\mathcal{U}(\Pi)}{X}$), it follows that there is some 
$K'\in\nbody{r'}\cap X$. 
This in turn implies that there exists some $s\in \Pi$ such that 
$K=\head{s}$ (and thus $K'=\head{s'}$) and 
$\ap{2}{s}\in\GR{\mathcal{U}(\Pi)}{X}$.
From the last condition we get $\pbody{s}\subseteq X$ and 
$\nbody{s}\cap X=\emptyset$, and therefore $K=\head{s}\in X$. 
Thus, $K\in X\cap\nbody{r}$, contradicting 
$r\in\GR{\mathcal{U}(\Pi)}{X}$.

\item [\ref{thm:translation:results:3}.]
This property is a simple consequence of the observation that 
$\body{\ap{2}{r}}=\body{r}\cup\nbody{r'}$, together with 
Item~\ref{thm:translation:results:2}. 
The details follow.

\paragraph{If part.} 
Since $\body{r}\subseteq\body{\ap{2}{r}}$, the conditions
$\pbody{\ap{2}{r}}\subseteq X$ and $\nbody{\ap{2}{r}}\cap X=\emptyset$ 
yield $\pbody{r}\subseteq X$ and $\nbody{r}\cap X=\emptyset$.
So, $\ap{2}{r}\in\GR{\mathcal{U}(\Pi)}{X}$
implies $r\in\Pi\cap\GR{\mathcal{U}(\Pi)}{X}$.

\paragraph{Only-if part.} 
From Item~\ref{thm:translation:results:2}, it holds that 
$\nbody{r}\cap X=\emptyset$ implies $\nbody{r'}\cap X=\emptyset$.
Thus, using the relation $\body{\ap{2}{r}}=\body{r}\cup\nbody{r'}$, 
we get that the two conditions $\pbody{r}\subseteq X$ and 
$\nbody{r}\cap X=\emptyset$ jointly imply  $\pbody{\ap{2}{r}}\subseteq X$ 
and $\nbody{\ap{2}{r}}\cap X=\emptyset$.
Therefore, $r\in\Pi\cap\GR{\mathcal{U}(\Pi)}{X}$
only if $\ap{2}{r}\in\GR{\mathcal{U}(\Pi)}{X}$.

\item [\ref{thm:translation:results:4}.]
Consider $r\in\Pi$.

\paragraph{If part.}
Assume $\bl{1}{r}{L}\in\GR{\mathcal{U}(\Pi)}{X}$ or
$\bl{2}{r}{L}\in\GR{\mathcal{U}(\Pi)}{X}$
for some $L\in\pbody{r}\cup\nbody{r}$.

\begin{itemize}
\item Suppose $\bl{1}{r}{L}\in\GR{\mathcal{U}(\Pi)}{X}$. 
Then, $L\notin X$, for $L\in \pbody{r}$. 
Hence, $\pbody{r}\not\subseteq X$ and therefore
$r\notin\GR{\mathcal{U}(\Pi)}{X}$. 

\item Suppose $\bl{2}{r}{L}\in\GR{\mathcal{U}(\Pi)}{X}$. 
Then, $L\in X$, for $L\in \nbody{r}$. 
Hence, $\nbody{r}\cap X\neq\emptyset$ and therefore 
$r\notin\GR{\mathcal{U}(\Pi)}{X}$. 
\end{itemize}

\paragraph{Only-if part.}
Suppose $r\notin \GR{\mathcal{U}(\Pi)}{X}$. 
There are two cases to distinguish.

\begin{itemize}
\item $\pbody{r}\not\subseteq X$: 
Then, there is some $L\in\pbody{r}$ such that $L\notin X$. 
According to Item~\ref{thm:translation:results:2}, we also have $L'\notin X$.
Furthermore, by Condition~\ref{l:be2:results:2} of 
Proposition~\ref{thm:be:results}, we have $\ok{\name{r}}\in X$. 
So, $\pbody{\bl{1}{r}{L}}\subseteq X$ and 
$\nbody{\bl{1}{r}{L}}\cap X=\emptyset$. 
This  means that $\bl{1}{r}{L}\in\GR{\mathcal{U}(\Pi)}{X}$ holds.

\item $\nbody{r}\cap X\neq \emptyset$: 
Then, there is some $L\in \nbody{r}$ such that $L\in X$. 
Invoking Item~\ref{thm:translation:results:2} again, we get $L'\in X$. 
Thus, $\bl{2}{r}{L}\in\GR{\mathcal{U}(\Pi)}{X}$.
\end{itemize}

\item [\ref{thm:translation:results:5}.]
Suppose $\cok{5}{r}{s,L}\in\GR{\mathcal{U}(\Pi)}{X}$.
Then, $L\in X$ for $L\in\nbody{s}$. 
By Item~\ref{thm:translation:results:2}, we also have $L'\in X$.
Furthermore, $\ok{\name{r}}\in X$, by Condition~\ref{l:be2:results:2} 
of Proposition~\ref{thm:be:results}.
It follows that $\bl{2}{s}{L}\in\GR{\mathcal{U}(\Pi)}{X}$ holds.
$\mathproofbox$
\end{itemize}
\end{proof*}


\begin{proof}[Proof of Proposition~\ref{thm:be:results}]
Analogous to the proof of Proposition~\ref{thm:results:i}.
\end{proof}


\begin{proof}[Proof of Theorem~\ref{thm:be:ordering}]
Let $(\Pi,\PRECMo)$ be a statically ordered logic program
and let $X$ be a consistent answer set of
\[
\mathcal{U}'(\Pi)
\; = \;
\mathcal{U}(\Pi)\cup\{(\PREC{n_1}{n_2})\LPif\,\mid (r_1, r_2) \in\;\PRECMo \}
\ \mbox{.}
\]
Furthermore, let
\[
\hat{\Pi}
\; = \;
\Pi\setminus\{r\in\Pi\mid\head{r}\in X,\nbody{r}\cap X\neq\emptyset\}
\]
and let
\(
\langle s_i\rangle_{i\in I}
\)
be some grounded enumeration of \GR{\mathcal{U}'(\Pi)}{X}.
\begin{lemma}\label{lem:be:ordering}
  Given the above prerequisites, we have for all
  \(
  r,r'\in\hat{\Pi}
  \)\/:
  
  If \PRECM{r}{r'},
  then $j<i$
  for
  \(
  s_i=\cokt{1}{r}
  \)
  and
  \(
  s_j=\cokt{1}{r'}
  \).
\end{lemma}
\begin{proof}[Proof of Lemma~\ref{lem:be:ordering}]
Analogous to the proof of Theorem~\ref{thm:order:implementing}.
\end{proof}
By Proposition~\ref{thm:be:results}, we have $\ok{r}\in X$ for all $r\in\Pi$.
Therefore, we also have $\cokt{1}{r}\in\GR{\mathcal{U}'(\Pi)}{X}$ for all $r\in\Pi$.
Now, define for all $r_1,r_2\in\hat{\Pi}$ that
\(
r_1\ll r_2
\)
if $j<i$
for
\(
s_i=\cokt{1}{r_1}
\)
and
\(
s_j=\cokt{1}{r_2}
\).
By definition, $\ll$ is a total ordering on $\hat{\Pi}$.
Furthermore, \PRECM{r_1}{r_2} implies $r_1\ll r_2$.
That is, 
\(
(\PRECMo\cap\;({\hat{\Pi}\times \hat{\Pi}}))\subseteq\;\ll
\).
\end{proof}

\bibliographystyle{tlp}
\bibliography{pref-tplp}

\begin{thebibliography}{}

\bibitem[\protect\citename{Baader \& Hollunder, }1993]{baahol93a}
Baader, F., \& Hollunder, B. (1993).
\newblock How to prefer more specific defaults in terminological default logic.
\newblock {\em Pages  669--674 of:} Bajcsy, R. (ed), {\em {Proceedings of the
  Thirteenth International Joint Conference on Artificial Intelligence {\rm
  (}IJCAI'93\/{\rm )}}}.
\newblock Morgan Kaufmann Publishers.

\bibitem[\protect\citename{Benferhat {\em et~al.}\relax, }1993]{bcdlp93a}
Benferhat, S., Cayrol, C., Dubois, D., Lang, J., \& Prade, H. (1993).
\newblock Inconsistency management and prioritized syntax-based entailment.
\newblock {\em Pages  640--647 of:} Bajcsy, R. (ed), {\em {Proceedings of the
  Thirteenth International Joint Conference on Artificial Intelligence {\rm
  (}IJCAI'93\/{\rm )}}}.
\newblock Morgan Kaufmann Publishers.

\bibitem[\protect\citename{Besnard {\em et~al.}\relax, }2002]{bemesc01a}
Besnard, Ph., Mercer, R., \& Schaub, T. (2002).
\newblock Optimality theory via default logic.
\newblock  Benferhat, S., \& Giunchiglia, E. (eds), {\em {Proceedings of the
  Ninth International Workshop on Non-Monotonic Reasoning}}.
\newblock To appear.

\bibitem[\protect\citename{Brewka, }1989]{brewka89e}
Brewka, G. (1989).
\newblock Preferred subtheories: An extended logical framework for default
  reasoning.
\newblock {\em Pages  1043--1048 of:} {\em {Proceedings of the Eleventh
  International Joint Conference on Artificial Intelligence {\rm
  (}IJCAI'89\/{\rm )}}}.
\newblock Morgan Kaufmann Publishers.

\bibitem[\protect\citename{Brewka, }1994]{brewka94a}
Brewka, G. (1994).
\newblock Adding priorities and specificity to default logic.
\newblock {\em Pages  247--260 of:} Pereira, L., \& Pearce, D. (eds), {\em
  {European Workshop on Logics in Artificial Intelligence {\rm (}JELIA'94\/{\rm
  )}}}.
\newblock Lecture Notes in Artificial Intelligence, vol. 838.
\newblock Springer-Verlag.

\bibitem[\protect\citename{Brewka, }1996]{brewka96a}
Brewka, G. (1996).
\newblock Well-founded semantics for extended logic programs with dynamic
  preferences.
\newblock {\em {Journal of Artificial Intelligence Research}}, {\bf 4}, 19--36.

\bibitem[\protect\citename{Brewka \& Eiter, }1998]{breeit97a}
Brewka, G., \& Eiter, T. (1998).
\newblock Preferred answer sets for extended logic programs.
\newblock {\em Pages  86--97 of:} Cohn, A., Schubert, L., \& Shapiro, S. (eds),
  {\em {Proceedings of the Sixth International Conference on the Principles of
  Knowledge Representation and Reasoning {\rm (}KR'98\/{\rm )}}}.
\newblock Morgan Kaufmann Publishers.

\bibitem[\protect\citename{Brewka \& Eiter, }1999]{breeit99a}
Brewka, G., \& Eiter, T. (1999).
\newblock Preferred answer sets for extended logic programs.
\newblock {\em {Artificial Intelligence}}, {\bf 109}(1-2), 297--356.

\bibitem[\protect\citename{Brewka \& Eiter, }2000]{breeit98a}
Brewka, G., \& Eiter, T. (2000).
\newblock Prioritizing default logic.
\newblock {\em Pages  27--45 of:} H{\"o}lldobler, St. (ed), {\em {Intellectics
  and Computational Logic --- Papers in Honour of {W}olfgang {B}ibel}}.
\newblock Kluwer Academic Publishers.

\bibitem[\protect\citename{Delgrande \& Schaub, }1997]{delsch97a}
Delgrande, J., \& Schaub, T. (1997).
\newblock Compiling reasoning with and about preferences into default logic.
\newblock {\em Pages  168--174 of:} Pollack, M. (ed), {\em {Proceedings of the
  Fifteenth International Joint Conference on Artificial Intelligence {\rm
  (}IJCAI'97\/{\rm )}}}.
\newblock Morgan Kaufmann Publishers.

\bibitem[\protect\citename{Delgrande \& Schaub, }2000a]{delsch00a}
Delgrande, J., \& Schaub, T. (2000a).
\newblock Expressing preferences in default logic.
\newblock {\em {Artificial Intelligence}}, {\bf 123}(1-2), 41--87.

\bibitem[\protect\citename{Delgrande \& Schaub, }2000b]{delsch00b}
Delgrande, J., \& Schaub, T. (2000b).
\newblock The role of default logic in knowledge representation.
\newblock {\em Pages  107--126 of:} Minker, J. (ed), {\em {Logic-Based
  Artificial Intelligence}}.
\newblock Dordrecht: Kluwer Academic Publishers.

\bibitem[\protect\citename{Delgrande {\em et~al.}\relax, }2000a]{descto00d}
Delgrande, J., Schaub, T., \& Tompits, H. (2000a).
\newblock A compilation of {B}rewka and {E}iter's approach to prioritization.
\newblock {\em Pages  376--390 of:} Ojeda-Aciego, M., Guzm{\'a}n, I., Brewka,
  G., \& Pereira, L. (eds), {\em {Proceedings of the Eighth European Workshop
  on Logics in Artificial Intelligence {\rm (}JELIA 2000\/{\rm )}}}.
\newblock Lecture Notes in Artificial Intelligence, vol. 1919.
\newblock Springer-Verlag.

\bibitem[\protect\citename{Delgrande {\em et~al.}\relax, }2000b]{descto00b}
Delgrande, J., Schaub, T., \& Tompits, H. (2000b).
\newblock A compiler for ordered logic programs.
\newblock  Baral, C., \& Truszczy\'{n}ski, M. (eds), {\em {Proceedings of the
  Eighth International Workshop on Non-Monotonic Reasoning}}.
\newblock \texttt{arXiv.org} e-Print archive.
\newblock System Abstract.

\bibitem[\protect\citename{Delgrande {\em et~al.}\relax, }2000c]{descto00a}
Delgrande, J., Schaub, T., \& Tompits, H. (2000c).
\newblock Logic programs with compiled preferences.
\newblock  Baral, C., \& Truszczy\'{n}ski, M. (eds), {\em {Proceedings of the
  Eighth International Workshop on Non-Monotonic Reasoning}}.
\newblock \texttt{arXiv.org} e-Print archive.

\bibitem[\protect\citename{Delgrande {\em et~al.}\relax, }2000d]{descto00c}
Delgrande, J., Schaub, T., \& Tompits, H. (2000d).
\newblock Logic programs with compiled preferences.
\newblock {\em Pages  392--398 of:} Horn, W. (ed), {\em {Proceedings of the
  Fourteenth European Conference on Artificial Intelligence {\rm (}ECAI
  2000\/{\rm )}}}.
\newblock IOS Press.

\bibitem[\protect\citename{Delgrande {\em et~al.}\relax, }2001]{descto01a}
Delgrande, J., Schaub, T., \& Tompits, H. (2001).
\newblock plp: A generic compiler for ordered logic programs.
\newblock {\em Pages  411--415 of:} Eiter, T., Faber, W., \& Truszczy\'{n}ski,
  M. (eds), {\em {Proceedings of the Sixth International Conference on Logic
  Programming and Nonmonotonic Reasoning {\rm (}LPNMR 2001\/{\rm )}}}.
\newblock Lecture Notes in Artificial Intelligence, vol. 2173.
\newblock Springer-Verlag.

\bibitem[\protect\citename{Doyle \& Wellman, }1991]{doywel91a}
Doyle, J., \& Wellman, M. (1991).
\newblock Impediments to universal preference-based default theories.
\newblock {\em {Artificial Intelligence}}, {\bf 49}(1-3), 97--128.

\bibitem[\protect\citename{Eiter \& Gottlob, }1995]{eitgot95}
Eiter, T., \& Gottlob, G. (1995).
\newblock The complexity of logic-based abduction.
\newblock {\em {Journal of the ACM}}, {\bf 42}, 3--42.

\bibitem[\protect\citename{Eiter {\em et~al.}\relax, }1997]{dlv97a}
Eiter, T., Leone, N., Mateis, C., Pfeifer, G., \& Scarcello, F. (1997).
\newblock A deductive system for nonmonotonic reasoning.
\newblock {\em Pages  363--374 of:} Dix, J., Furbach, U., \& Nerode, A. (eds),
  {\em {Proceedings of the Fourth International Conference on Logic Programming
  and Non-Monotonic Reasoning {\rm (}LPNMR'97\/{\rm )}}}.
\newblock Lecture Notes in Artificial Intelligence, vol. 1265.
\newblock Springer-Verlag.

\bibitem[\protect\citename{Eiter {\em et~al.}\relax, }2002]{eifisato01a}
Eiter, T., Fink, M., Sabbatini, G., \& Tompits, H. (2002).
\newblock A generic approach for knowledge-based information-site selection.
\newblock  {\em {Proceedings of the Eighth International Conference on the
  Principles of Knowledge Representation and Reasoning {\rm (}KR 2002\/{\rm
  )}}}.
\newblock Morgan Kaufmann Publishers.

\bibitem[\protect\citename{Geffner \& Pearl, }1992]{gefpea92}
Geffner, H., \& Pearl, J. (1992).
\newblock Conditional entailment: Bridging two approaches to default reasoning.
\newblock {\em {Artificial Intelligence}}, {\bf 53}(2-3), 209--244.

\bibitem[\protect\citename{Gelfond \& Lifschitz, }1988]{gellif88b}
Gelfond, M., \& Lifschitz, V. (1988).
\newblock The stable model semantics for logic programming.
\newblock {\em Pages  1070--1080 of:} {\em {Proceedings of the Fifth
  International Conference on Logic Programming {\rm (}ICLP'88\/{\rm )}}}.
\newblock The MIT Press.

\bibitem[\protect\citename{Gelfond \& Lifschitz, }1991]{gellif91a}
Gelfond, M., \& Lifschitz, V. (1991).
\newblock Classical negation in logic programs and deductive databases.
\newblock {\em {New Generation Computing}}, {\bf 9}, 365--385.

\bibitem[\protect\citename{Gelfond \& Son, }1997]{gelson97a}
Gelfond, M., \& Son, T. (1997).
\newblock Reasoning with prioritized defaults.
\newblock {\em Pages  164--223 of:} Dix, J., Pereira, L., \& Przymusinski, T.
  (eds), {\em {Third International Workshop on Logic Programming and Knowledge
  Representation}}.
\newblock Lecture Notes in Computer Science, vol. 1471.
\newblock Springer-Verlag.

\bibitem[\protect\citename{Gelfond {\em et~al.}\relax, }1989]{geprpr89}
Gelfond, M., Przymusinska, H., \& Przymusinski, T. (1989).
\newblock On the relationship between circumscription and negation as failure.
\newblock {\em {Artificial Intelligence}}, {\bf 38}(1), 75--94.

\bibitem[\protect\citename{Gordon, }1993]{gordon93a}
Gordon, T. (1993).
\newblock {\em The pleading game: An artificial intelligence model of
  procedural justice}.
\newblock Dissertation, Technische Hochschule Darmstadt, Alexanderstra{\ss}e
  10, D-64283 Darmstadt, Germany.

\bibitem[\protect\citename{Kager, }1999]{kager99}
Kager, R. (1999).
\newblock {\em Optimality theory: A textbook}.
\newblock Cambridge University Press.

\bibitem[\protect\citename{Konolige, }1988]{konolige88b}
Konolige, K. (1988).
\newblock Hierarchic autoepistemic theories for nonmonotonic reasoning.
\newblock {\em Pages  439--443 of:} {\em {Proceedings of the Seventh National
  Conference on Artificial Intelligence {\rm (}AAAI'88\/{\rm )}}}.
\newblock Morgan Kaufmann Publishers.

\bibitem[\protect\citename{Lifschitz, }1996]{lifschitz96a}
Lifschitz, V. (1996).
\newblock Foundations of logic programming.
\newblock {\em Pages  69--127 of:} Brewka, G. (ed), {\em {Principles of
  Knowledge Representation}}.
\newblock CSLI Publications.

\bibitem[\protect\citename{McCarthy, }1986]{mccarthy86}
McCarthy, J. (1986).
\newblock Applications of circumscription to formalizing common-sense
  knowledge.
\newblock {\em {Artificial Intelligence}}, {\bf 28}, 89--116.

\bibitem[\protect\citename{Nebel, }1998]{nebel98a}
Nebel, B. (1998).
\newblock How hard is it to revise a belief base?
\newblock {\em Pages  77--145 of:} Dubois, D., \& Prade, H. (eds), {\em
  {Handbook of Defeasible Reasoning and Uncertainty Management Systems, Volume
  3: Belief Change}}.
\newblock Dordrecht: Kluwer Academic Publishers.

\bibitem[\protect\citename{Niemel{\"a} \& Simons, }1997]{niesim97a}
Niemel{\"a}, I., \& Simons, P. (1997).
\newblock Smodels: An implementation of the stable model and well-founded
  semantics for normal logic programs.
\newblock {\em Pages  420--429 of:} Dix, J., Furbach, U., \& Nerode, A. (eds),
  {\em {Proceedings of the Fourth International Conference on Logic Programming
  and Nonmonotonic Reasoning {\rm (}LPNMR'97\/{\rm )}}}.
\newblock Lecture Notes in Artificial Intelligence, vol. 1265.
\newblock Springer-Verlag.

\bibitem[\protect\citename{Papadimitriou \& Sideri, }1994]{papsid94}
Papadimitriou, C., \& Sideri, M. (1994).
\newblock Default theories that always have extensions.
\newblock {\em {Artificial Intelligence}}, {\bf 69}, 347--357.

\bibitem[\protect\citename{Prince \& Smolensky, }1993]{prismo93}
Prince, A., \& Smolensky, P. (1993).
\newblock {\em Optimality theory: Constraint interaction in generative
  grammar}.
\newblock Tech. rept. University of Colorado, Boulder.

\bibitem[\protect\citename{Reiter, }1980]{reiter80}
Reiter, R. (1980).
\newblock A logic for default reasoning.
\newblock {\em {Artificial Intelligence}}, {\bf 13}(1-2), 81--132.

\bibitem[\protect\citename{Rintanen, }1994]{rintanen94a}
Rintanen, J. (1994).
\newblock Prioritized autoepistemic logic.
\newblock {\em Pages  232--246 of:} MacNish, C., Pearce, D., \& Pereira, L.~M.
  (eds), {\em {Proceedings of the European Workshop on Logics in Artificial
  Intelligence {\rm (}JELIA'94\/{\rm )}}}.
\newblock Lecture Notes in Artificial Intelligence, vol. 838.
\newblock Berlin: Springer-Verlag.

\bibitem[\protect\citename{Rintanen, }1998a]{rintanen98a}
Rintanen, J. (1998a).
\newblock Complexity of prioritized default logics.
\newblock {\em {Journal of Artificial Intelligence Research}}, {\bf 9},
  423--461.

\bibitem[\protect\citename{Rintanen, }1998b]{rintanen98b}
Rintanen, J. (1998b).
\newblock Lexicographic priorities in default logic.
\newblock {\em {Artificial Intelligence}}, {\bf 106}, 221--265.

\bibitem[\protect\citename{Sakama \& Inoue, }1996]{sakino96}
Sakama, C., \& Inoue, K. (1996).
\newblock Representing priorities in logic programs.
\newblock {\em Pages  82--96 of:} Maher, M. (ed), {\em {Proceedings of the 1996
  Joint International Conference and Symposium on Logic Programming}}.
\newblock Cambridge: The MIT Press.

\bibitem[\protect\citename{Sakama \& Inoue, }2000]{sakino00a}
Sakama, C., \& Inoue, K. (2000).
\newblock Prioritized logic programming and its application to commonsense
  reasoning.
\newblock {\em {Artificial Intelligence}}, {\bf 123}(1-2), 185--222.

\bibitem[\protect\citename{Schaub \& Wang, }2001a]{schwan01b}
Schaub, T., \& Wang, K. (2001a).
\newblock A comparative study of logic programs with preference.
\newblock {\em Pages  597--602 of:} Nebel, B. (ed), {\em {Proceedings of the
  Seventeenth International Joint Conference on Artificial Intelligence {\rm
  (}IJCAI 2001\/{\rm )}}}.
\newblock Morgan Kaufmann Publishers.

\bibitem[\protect\citename{Schaub \& Wang, }2001b]{schwan01c}
Schaub, T., \& Wang, K. (2001b).
\newblock {\em Towards a semantic framework for preference handling in logic
  programming}.
\newblock Submitted for publication; extended version of~\cite{schwan01b}.

\bibitem[\protect\citename{van~der Hoek \& Witteveen, }2000]{hoewit00a}
van~der Hoek, W., \& Witteveen, C. (2000).
\newblock Classical and general frameworks for recovery.
\newblock {\em Pages  33--37 of:} Horn, W. (ed), {\em {Proceedings of the
  Fourteenth European Conference on Artificial Intelligence {\rm (}ECAI
  2000\/{\rm )}}}.
\newblock Amsterdam: IOS Press.

\bibitem[\protect\citename{Wang {\em et~al.}\relax, }2000]{wazhli00}
Wang, K., Zhou, L., \& Lin, F. (2000).
\newblock Alternating fixpoint theory for logic programs with priority.
\newblock {\em Pages  164--178 of:} {\em {Proceedings of the First
  International Conference on Computational Logic {\rm (}CL 2000\/{\rm )}}}.
\newblock Lecture Notes in Computer Science, vol. 1861.
\newblock Springer-Verlag.

\bibitem[\protect\citename{Zhang, }2000]{zhangXX}
Zhang, Y. (2000).
\newblock {\em Logic program based updates}.
\newblock Draft available at \url{http://www.cit.uws.edu.au/~yan/}.

\bibitem[\protect\citename{Zhang \& Foo, }1997]{zhafoo97a}
Zhang, Y., \& Foo, N. (1997).
\newblock Answer sets for prioritized logic programs.
\newblock {\em Pages  69--84 of:} Maluszynski, J. (ed), {\em {Proceedings of
  the International Symposium on Logic Programming {\rm (}ILPS'97\/{\rm )}}}.
\newblock The MIT Press.

\end{thebibliography}

\end{document}